%% file: mve_arxiv.tex
\title{Generalized Multi-view Embedding for Visual Recognition and Cross-modal Retrieval}
\author{Guanqun Cao, Alexandros Iosifidis, \emph{Senior Member, IEEE}, Ke Chen and Moncef Gabbouj, \emph{Fellow, IEEE}\\[1mm]
}

\date{\today}
\documentclass[journal]{IEEEtran}
\usepackage{graphicx,amssymb,amstext,amsmath}
\usepackage{color}
\usepackage{grffile}
\usepackage{xcolor}
\usepackage{mathtools}
\usepackage{hyperref}
\usepackage{amsmath}
\usepackage{setspace}
\usepackage{fancyhdr}
\usepackage{tabularx}
\usepackage{mathdots}
\usepackage{bm}
\usepackage{epstopdf}
\usepackage{makecell}
\usepackage{multirow}
\usepackage{caption}
\usepackage{array}
\usepackage{subcaption}
\usepackage{dblfloatfix}
\usepackage{standalone}

\newif\ifarial
\hypersetup{colorlinks=true, linkcolor=blue!50!black}
\begin{document}
\setlength{\belowcaptionskip}{-8pt}
\newcommand{\Tr}{\operatorname{Tr}}
\newcommand{\X}{\mathbf{X}}
\newcommand{\Y}{\mathbf{Y}}
\newcommand{\W}{\mathbf{W}}
\newcommand{\I}{\mathbf{I}}
\newcommand{\e}{\mathbf{e}}
\newcommand{\x}{\mathbf{x}}
\newcommand{\y}{\mathbf{y}}
\newcommand{\w}{\mathbf{w}}
\newcommand{\A}{\mathbf{A}}
\newcommand{\B}{\mathbf{B}}
\newcommand{\Pb}{\mathbf{P}}
\newcommand{\Q}{\mathbf{Q}}
\newcommand{\K}{\mathbf{K}}
\newcommand{\Hb}{\mathbf{H}}
\newcommand{\R}{\mathbb{R}}
\newcommand{\Phib}{\mathbf{\Phi}}
\newcommand{\phib}{\bm{\phi}}
\newcommand{\rhob}{\bm{\rho}}
\newcommand{\J}{\mathcal{J}}

\newcommand{\onedot}{$\mathsurround0pt\ldotp$}
\newcommand{\eg}{\emph{e.g}\onedot} \newcommand{\Eg}{\emph{E.g}\onedot}
\newcommand{\ie}{\emph{i.e}\onedot} \newcommand{\Ie}{\emph{I.e}\onedot}
\newcommand{\cf}{\emph{c.f}\onedot} \newcommand{\Cf}{\emph{C.f}\onedot}
\newcommand{\etc}{\emph{etc}\onedot} \newcommand{\vs}{\emph{vs}\onedot}
\newcommand{\wrt}{w.r.t\onedot} \newcommand{\dof}{d.o.f\onedot}
\newcommand{\etal}{\emph{et al}\onedot }

\renewcommand{\arraystretch}{1.1}
\newcolumntype{L}[1]{>{\raggedright\let\newline\\\arraybackslash\hspace{0pt}}m{#1}}
\newcolumntype{C}[1]{>{\centering\let\newline\\\arraybackslash\hspace{0pt}}m{#1}}
\newcolumntype{R}[1]{>{\raggedleft\let\newline\\\arraybackslash\hspace{0pt}}m{#1}}
\graphicspath{{./figures/}}
\maketitle
\pagestyle{plain}
{\abstract
In this paper, the problem of multi-view embedding from different visual cues and modalities is considered. We propose a unified solution for subspace learning methods using the Rayleigh quotient, which is extensible for multiple views, supervised learning, and non-linear embeddings. Numerous methods including Canonical Correlation Analysis, Partial Least Square regression and Linear Discriminant Analysis are studied using specific intrinsic and penalty graphs within the same framework. Non-linear extensions based on kernels and (deep) neural networks are derived, achieving better performance than the linear ones. Moreover, a novel Multi-view Modular Discriminant Analysis (MvMDA) is proposed by taking the view difference into consideration. We demonstrate the effectiveness of the proposed multi-view embedding methods on visual object recognition and cross-modal image retrieval, and obtain superior results in both applications compared to related methods.}
\section{Introduction}
People see the world differently, and objects are described from various point of views and modalities. Identifying an object can not only benefit from visual cues including color, texture and shape, but textual annotations from different observations and languages. Thanks to data enrichment from sensor technologies, the accuracy in image retrieval and recognition has been significantly improved by taking advantage of multi-view and cross-domain learning \cite{CostaPereira2014, Xu2013}. Since matching the data samples across various feature spaces directly is infeasible, subspace learning approaches, which learn a common feature space from multi-view spaces, becomes an effective approach in solving the problem. \par
Numerous methods have been proposed in subspace learning. They can be grouped into three major categories based on the characteristics of machine learning: \emph{two-view learning} and \emph{multi-view learning}; \emph{unsupervised learning} and \emph{supervised learning}; and \emph{linear learning} and \emph{non-linear learning}. While traditional techniques in multivariate analysis take two inputs \cite{Mardia1980}, multi-view methods have been proposed to find an optimal representation from more than two views \cite{Nielsen2002, Gong2014}. Compared to learning the feature transformation in an unsupervised manner, discriminative methods, such as Linear Discriminant Analysis (LDA) have been extended to multi-view cases. Additionally, the transformation can also be kernel-based or learned by (deep) neural nets to exploit their non-linear properties.\par
\emph{Two-view learning} and \emph{multi-view learning}: One of the most popular methods in multivariate statistics is Canonical Correlation Analysis (CCA) \cite{Hotelling1936}. It seeks to maximize the correlation between two sets of variables. Alternatively, its multi-view counterpart aims to obtain a common space from $V>2$ views \cite{Nielsen2002, Gong2014, Luo2015}. This is achieved either by scaling the cross-covariance matrices to incorporate the covariances from more than two views, or by finding the best rank-1 approximation of the
data covariance tensor. A similar approach to find the common subspace is Partial Least Square Regressions \cite{Wold1984}. It maximizes the cross-covariance from two views by regressing the data samples to the common space. Besides transformation and regression, Multi-view Fisher Discriminant Analysis (MFDA) \cite{Diethe2008} learns the transformation minimizing the difference between data samples of predicted labels. The Dropout regularization was introduced for the multi-view linear discirminant analysis in \cite{Cao2016a}.\par
\emph{Unsupervised learning} and \emph{supervised learning}: In contrast to unsupervised transformations, including CCA and PLS, LDA \cite{Yan2007, Iosifidis2013} exploits the class labels effectively by maximizing the between-class scatter while minimizing the within-class scatter simultaneously. CCA has been successfully combined with LDA to find a discriminative subspace in \cite{Sun2008, Ma2007, Sun2016}. Coupled Spectral Regression (CSR) \cite{Lei2009a} projects two different inputs to the low-dimensional embedding of labels by PLS regressions. Consistent with the original LDA, a Multi-view Discriminant Analysis (MvDA) \cite{Kan2016} finds a discriminant representation over $V$ views. The between-class scatter is maximized regardless of the difference between inter-view and intra-view covariances, while the within-class scatter is minimized in the mean time. Generalized Multi-view Analysis (GMA) \cite{Sharma2012} was proposed to maximize the intra-view discriminant information. Recently, a semi-supervised alternative \cite{Liu2015} was also proposed for multi-view learning, which adopts a non-negative matrix factorization method for view mapping and a robust sparse regression model for clustering the labeled samples. Moreover, a multi-view information bottleneck method \cite{Xu2014} was proposed to retain its discrimination and robustness for multi-view learning.\par
\emph{Linear} and \emph{non-linear learning}: Many problems are not linearly separable and thereby kernel-based methods and learning representation by (deep) neural nets are introduced. By mapping the features to the high dimensional feature space using the kernel trick \cite{Schoelkopf1999}, kernel CCA \cite{Hardoon2004} adopts a pre-defined kernel and limits its application on small datasets. Many linear multi-view methods subsequently made their kernel extension \cite{Sun2007a, Sun2016, Iosifidis2016}. Kernel approximation \cite{Gong2014} was adopted later to work on big data. Deep CCA \cite{Andrew2013} was proposed using neural nets to learn adaptive non-linear representations from two views, and uses the weights in the last layers to find the maximum correlation. A similar idea has been exploited on LDA \cite{Dorfer2016}. PCANet \cite{Chan2015} was introduced to adopt a cascade of linear transformation, followed by binary hashing and block histograms. \par
We make several contributions in this paper:
First, we propose a unified multi-view subspace learning method for CCA, PLS and LDA techniques using the graph embedding framework \cite{Yan2007}. We design both intrinsic and penalty graphs to characterize the intra-view and inter-view information, respectively. The intra-view and inter-view covariance matrices are scaled up to incorporate more than two views for numerous techniques by exploiting their specific intrinsic and penalty graphs. In our proposed Multi-view Modular Discriminant Analysis (MvMDA), the two graphs also charaterize the within-class compactness and between-class separability. Based on the aforementioned characteristics of subspace learning algorithms, we propose a generalized objective function for multi-view subspace learning using Rayleigh quotient. This unified multi-view embedding approach can be solved as a generalized eigenvalue problem. \par
Second, we introduce a Multi-view Modular Discriminant Analysis (MvMDA) method by exploiting the distances between centers representing classes of different views. This is of particular interest since the resulting scatter encodes cross-view information, which empirically is shown to provide superior results.
Third, we also extend the unified framework to the non-linear cases with kernels and (deep) neural networks. Kernel-based multi-view learning method is derived with an implicit kernel mapping. For larger datasets, we use the explicit kernel mapping \cite{Rahimi2007} to approximate the kernel matrices. We also derive the formulation of stochastic gradient descent (SGD) for optimizing the objective function in the neural nets.\par
Last but not least, we demonstrate the effectiveness of the proposed embedding methods on visual object recognition and cross-modal image retrieval. Specifically, zero-shot recognition is evaluated by discovering novel object categories based on the underlying intermediate representation \cite{Farhadi2009, Lampert2009a, Lampert2014}. Its performance is heavily dependent on the representation in the latent space shared by visual and semantic cues. We integrate observations from \emph{attributes} as a middle-level semantic property for the joint learning. Superior recognition results are achieved by exploiting the latent feature space with non-linear solutions learned from the multi-view representations. We also employ the proposed multi-view subspace learning  methods for cross-modal image retrieval \cite{CostaPereira2014, Kang2015, Wei2016, He2015}. This type of methods differs from the co-training methods for image classification \cite{Yu2014a} and web image reranking \cite{Yu2014b, Yu2017}. In the experiments, we show promising retrieval results performed by embedding more modalities into the common feature space, and find that even conventional content-based image retrieval can be improved.\par
The rest of the paper is organized as follows. Section \ref{sec:rel} reviews the related work. In Section \ref{sec:mva}, we show the unified formulation to generalize the subspace learning methods. It is followed by the extension to multi-view techniques and derivation in kernels and neural nets. Then, in Section \ref{sec:exp}, we present the comparative results in zero-shot object recognition and cross-modal image retrieval on three popular multimedia datasets. Finally, Section \ref{sec:con} concludes the paper.
\begin{figure}
\includegraphics[width=.49\textwidth,height=.5\textwidth]{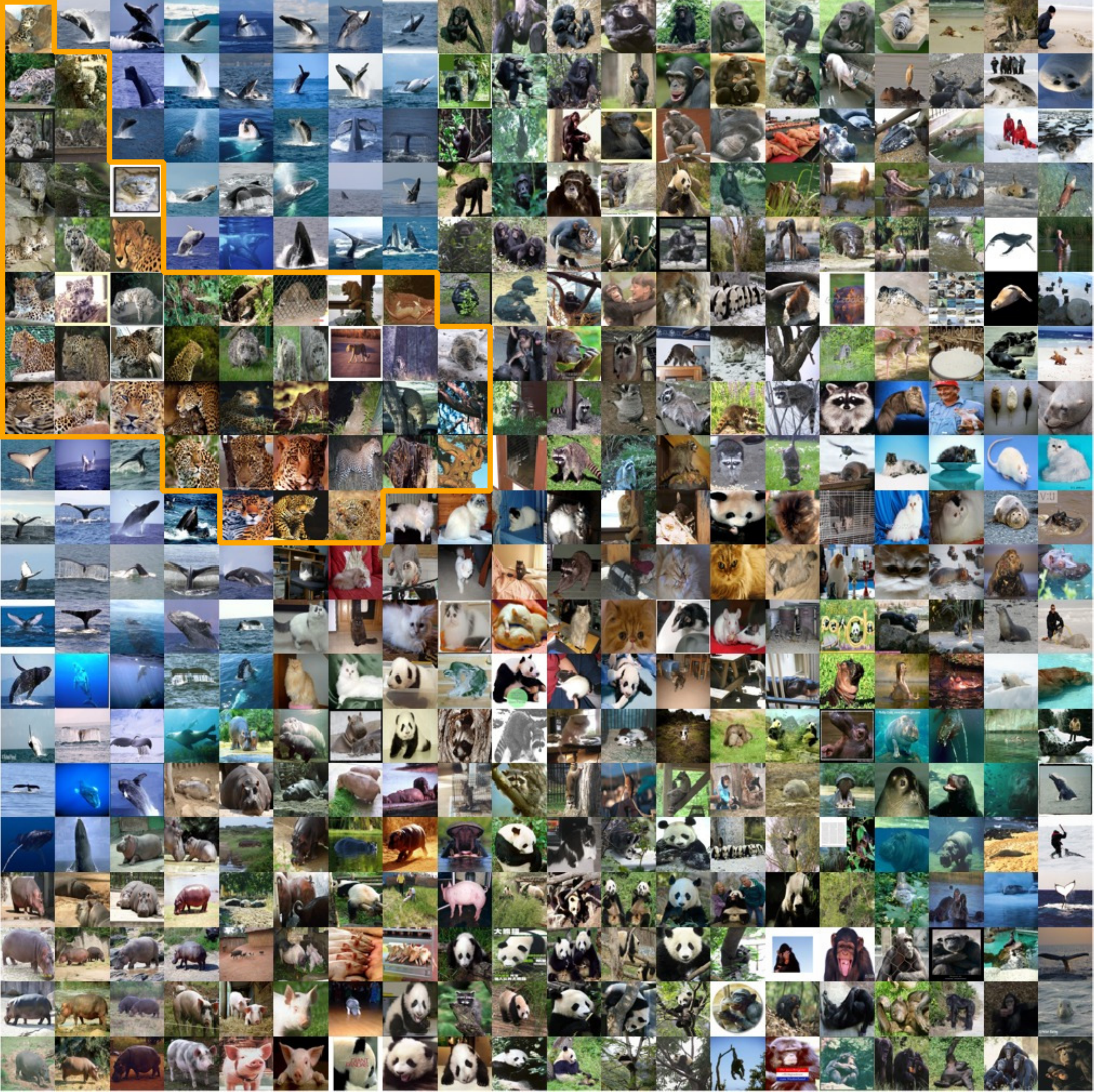}
\caption{Visualization of test images from the AwA dataset grouped by the features in the subspace. We highlight one of the representative  classes ``leopard'' bounded in orange to show images of the same animal categories are positioned in their neighborhoods after multi-view embedding. Note the 2-dimensional t-SNE map \cite{VanderMaaten2008} is generated from a near circular shape.}\label{fig:map}
\end{figure}
\section{Related work}\label{sec:rel}
In this section, we first define the common notations used throughout the paper. Then, we will briefly review the related methods for multi-view subspace learning. Moreover, recent work on non-linear methods concerning kernels and (deep) neural networks are discussed.
\subsection{Notations}
We define the data matrix $\X=[\x_1,\x_2,\dots,\x_N]$, $\x_i \in \mathbb{R}^{D}$, where $N$ is the number of samples and $D$ is the feature dimension. We also define $\X_v \in \mathbb{R}^{D_v \times N}, v=1,\dots,V$ for the feature vectors of the $v$th view, and discard the index in the single-view case for notation simplicity. Note that the dimensionality of the various feature spaces $D_v$ may vary across the views.
The covariance matrix is a statistics commonly used in CCA and PLS. We denote $\bar{\X}_v=\X_v-{1\over N}\X_v\,\e \, \e^\top$ as the centered data matrix. The cross-view covariance matrix between view $i$ and $j$ is then expressed as
$\mathbf{\Sigma}_{ij} = {1\over N}\bar{\X}_i{\bar{\X}_j}^\top
= {1\over N}\X_i \bigg(\I-{1\over N}\e\,\e^\top\bigg)\X_j^\top $, where
$\e\in \mathbb{R}^N$ is a vector of ones and $\I \in \mathbb{R}^{N \times N}$ is the identity matrix. For the supervised learning problems, the class label of the sample $\x_i$ is noted as $c_i \in \{1,2,\dots,C\}$, where $C$ is the number of classes. We define the class vector $\e^c\in \R^N$ with $e_c(i)=1$, if $c_i=c$, and $e_c(i)=0$, otherwise. $\W_v \in \R^{D_v \times d}, v=1,\dots,V$ is the projection matrix for each view, $d$ is the number of dimensions in the latent space.
The feature dimension $D_v$ in the original space of each view is usually high, which makes the distribution of the samples sparse, leading to several problems including the small sample size problem \cite{Hastie2013a}. Therefore we want to project the samples to the latent space. \par
The generic projection function is defined to project $\X \in \R^{D\times N}$ to $\mathbf{Y} \in \R^{d\times N}$.
We define the linear projection by $\mathbf{Y}=\W^\top \X$. In kernel methods, we map the data to a Hilbert space $\mathcal{F}$. Let us define $\phi(\cdot)$ as the non-linear function mapping $x_i\in \mathbb{R}^D$ to $\mathcal{F}$, and $\Phib=[\phib(\x_1),\dots,\phib(\x_N)]$ as the data matrix in $\mathcal{F}$. In multi-view cases,
${\Phib}=[
\Phib_1^\top, \hdots, \Phib_V^\top]^\top$.
Since the dimensionality of $\mathcal{F}$ is arbitrary, the kernel trick \cite{muller2001} is exploited in order to implicitly map the data to $\mathcal{F}$. The Gram matrix is given by
\begin{equation}
  \K_v=\kappa(\X_v,\X_v)={\Phib}_v^\top \cdot {\Phib}_v, \label{eq:kernel}
\end{equation}
where $\kappa(\cdot,\cdot)$ is the so-called kernel function.
The centered Gram matrix is
$\bar{\K}_v=\K_v - {1\over N}{\mathbf{1}}\,\K_v - {1\over N}\K_v\,{\mathbf{1}^\top}+
{1\over N^2}{\mathbf{1}}\K_v\,{\mathbf{1}}$,
where $\mathbf{1}\in \R^{N\times N}$ is an all-ones matrix.
In order to find the optimal projection, we can express $\W_v$ of each view as a linear combination of the training samples in the kernel space based on the Representer Theorem \cite{Schoelkopf1999, Schoelkopf2001}. This can be expressed by using a new weight matrix $\A_v$ as
\begin{equation}
\W_v = \Phib_v \A_v. \label{eq:weight}
\end{equation}
In the case where a neural network with $M$ layers is considered, $\bm{\beta}_j$ contains the weight parameters in the $j$th layer, $j=1,\dots,M$. The weights $\B=[\bm{\beta}_1,\dots,\bm{\beta}_M]$ are learned by applying stochastic gradient descent (SGD), and $h(\,\cdot\, ;\B)$ is a non-linear mapping function which maps $\X_v$ to the representation of the last hidden layer $\Hb_v$, i.e.
\begin{equation}
  \Hb_v = h(\X_v;\B_v),\label{eq:nnproj}
\end{equation}
where $\B_v$ is the weight matrix trained by applying backpropagation in the $v$th network.
\subsection{Canonical Correlation Analysis (CCA)}
Canonical Correlation Analysis (CCA) \cite{Hotelling1936,Borga2001} is a conventional statistical technique which finds the maximum correlation between two sets of data samples $\X_1\in \mathbb{R}^{D_1\times N}$ and $\X_2\in \mathbb{R}^{D_2 \times N}$ using the linear combination $\mathbf{Y}_1=\W_1^\top \X_1$ and $\mathbf{Y}_2=\W_2^\top \X_2$. $\W_1 $ and $\W_2$ are determined by optimizing:
\begin{align}
  \J &= \arg\max_{\W_1,\W_2}\text{corr}(\W_1^\top \X_1, \W_2^\top \X_2)\\
  & =\arg\max_{\W_1,\W_2}\frac{\W_1^\top\mathbf{\Sigma}_{12}\W_2}{\sqrt{\W_1^\top \mathbf{\Sigma}_{11}\W_1} \cdot \sqrt{\W_2^\top \mathbf{\Sigma}_{22}\W_2}},\label{eq:cca}
\end{align}
where
\begin{align}
    \mathbf{\mathbf{\Sigma}}&=\begin{bmatrix}
        \mathbf{\Sigma}_{11} & \mathbf{\Sigma}_{12}\\[1mm]
        \mathbf{\Sigma}_{21} & \mathbf{\Sigma}_{22}
  \end{bmatrix}
= \frac{1}{N} \begin{bmatrix}
     \bar{\X}_1 \bar{\X}_1^\top & \bar{\X}_1 \bar{\X}_2^\top \\[1mm]
     \bar{\X}_2 \bar{\X}_1^\top & \bar{\X}_2 \bar{\X}_2^\top
  \end{bmatrix}
\end{align}
\subsection{Kernel CCA}
Kernel CCA finds the maximum correlation between two views after mapping them to the kernel space \cite{Hardoon2004}. This is expressed by
\begin{align}
 \J = \arg\max_{\W_1,\W_2}\text{corr}(\W_1^\top \Phib_1, \W_2^\top \Phib_2)\label{eq:kcca}
\end{align}
We use the kernel trick \cite{muller2001} and the Representer Theorem in \eqref{eq:weight}, and derive the objective function for the kernel CCA as
\begin{equation}
\J =\arg\max_{\A_1,\A_2}\frac{\A_1^\top \K_1 \K_2 \; \A_2}{{\sqrt{\A_1^\top \K_1 \K_1 \; \A_1}} \cdot \sqrt{\A_2^\top \K_2 \K_2 \; \A_2}}.\label{eq:kcca}
\end{equation}
\subsection{Deep CCA}
Deep CCA maximizes the correlation between a pair of views by learning non-linear representations from the input data through multiple stacked layers of neurons \cite{Andrew2013, Wang2015b}.
A linear CCA layer is added on top of both networks, and the inputs to the CCA layer depend on the network outputs $\Hb_1$ and $\Hb_2$.
Similar to the non-linear case in \eqref{eq:kcca}, a modified objective function
$\min \limits_{{\W_1},{\W_2}}\, -{1\over N} \Tr\big({\W_1}^\top \Hb_1 \;\Hb_2^\top {\W_2}\big)$  is optimized, 
where $\W_1,\W_2$ are the projection matrices in the CCA layer, and the correlated outputs are $\mathbf{Y}_1=\W_1^\top \Hb_1$ and $\mathbf{Y}_2=\W_2^\top \Hb_2$.
A modified SGD method is developed with respect to the inputs $\Hb_1$ and $\Hb_2$ to the linear layer, which are also the outputs from the two networks.
The objective function is expressed as $\Tr\big(\W_1^\top \Hb_1 \;\Hb_2^\top {\W_2}\big)=\Tr(\mathbf{T}^\top \mathbf{T})^{1\over 2}$, which describes the correlation as the sum of the top $d$ singular vectors of $\mathbf{T}=\mathbf{\Sigma}_{11}^{-1/2}\mathbf{\Sigma}_{12}\mathbf{\Sigma}_{22}^{-1/2}$ whose definition can be found in \cite{Mardia1980}.
\subsection{Partial Least Squares (PLS) regression}
Partial Least Squares (PLS) regression \cite{Wold1984} is another dimensionality reduction technique derived from the linear combination of the input vectors ${\X_1}$ together with the target information which is considered as the second view ${\X_2}$. PLS maximizes the between-view covariance by solving
\begin{align}
  \J &= \arg\max_{\W_1,\W_2}[\Tr(\W^\top_1 \X_1\X_2^\top\,\W_2)],\\
  & \quad\, \text{subject to } \W_1^\top \W_1=\I, \W_2^\top \W_2=\I.
\end{align}
The non-linear extensions of PLS are obtained in the similar manner as the ones in CCA.
\subsection{Generalized Multi-view Analysis (GMA)}
\begin{singlespace}
GMA \cite{Sharma2012} is a generalized framework incorporating numerous dimensionality reduction methods. It maximizes the intra-view discriminant information, but ignores the inter-view information.
\end{singlespace}
{\footnotesize
\begin{align}
  \J &= \arg\max_{\W}\left[\Tr \left(\sum_i^V \sum_{i<j}^V 2 \lambda_{ij} \W_i^\top \X_i\X_j^\top\,\W_j + \sum_{i=1}^V\mu_i \W_i^\top \Pb_i \W_i \right)\right],\nonumber\\
& \quad\, \text{subject to } \sum_i^V \W_i^\top \Q_i \W_i=\I.
\end{align}
}%
\noindent
Here both $\Pb$ and $\Q$ are the intra-view covariance matrices. $\Pb$ is a square matrix and $\Q$ is a square symmetric definite matrix. We adopt Generalized Multiview Marginal Fisher Analysis (GMMFA) in this framework. The method is also kernelizable using the Representer Theorem and kernel trick.
\subsection{Linear Discriminant Analysis (LDA)}
Linear Discriminant Analysis (LDA) \cite{Yan2007, Martinez2001} finds the projection by maximizing the ratio of the between-class scatter to the within-class scatter.
Let us define by $\bm{\mu}_c$ the mean vector of the $c$'th class, formed by $N_c$ samples, and $\bm{\mu}$ the global mean. Then, LDA optimizes the following criterion:
\begin{equation}
    \J=\arg\max_\W  \frac{\Tr(\W ^\top \Pb \, \W )}{\Tr(\W ^\top \Q\,  \W) },\label{eq:lda}
\end{equation}
where
\footnotesize
\begin{align}
    \Pb  &= \sum_{c=1}^{C}N_c (\bm{\mu}_c-\bm{\mu})(\bm{\mu}_c-\bm{\mu})^\top = \X\bigg(\sum_{c=1}^{C} {1 \over N_c} \e_c{\e_c}^\top-{1\over N}\e\, \e^\top\bigg) \X^\top, \label{eq:sb}\\
  \Q &= \sum_{i=1}^N(\x_i-\bm{\mu}_{c})(\x_i-\bm{\mu}_{c})^\top
 = \X\bigg(\I-\sum_{c=1}^{C} {1 \over N_c} \e_c{\e_c}^\top\bigg)\X^\top.\label{eq:sw}
\end{align}
\normalsize
Non-linear extensions with kernels include KDA \cite{Baudat2000} and KRDA \cite{Iosifidis2014}.
\subsection{Multi-view Discriminant Analysis (MvDA)}
\begin{singlespace}
MvDA \cite{Kan2016} is the multi-view verison of LDA which maximizes the ratio of the determinant of the between-class scatter matrix to that of the within-class scatter matrix. Its objective function is
\end{singlespace}
\begin{equation}
    \J  =\arg\max_\W \frac{\Tr (\mathbf{S}^M_B) }{\Tr (\mathbf{S}^M_W)},
\end{equation}
where the between-class scatter matrix is
{ \small
\begin{equation}
\mathbf{S}^M_B = {\sum\limits_{i=1}^V \sum\limits_{j=1}^V \W_i^\top
\X_i\bigg(\sum_{c=1}^{C} {1 \over N_c} \e_c{\e_c}^\top-{1\over N}\e\, \e^\top\bigg) \X_j^\top \W_j}, \label{eq:mvdasb}
\end{equation}
}%
and the within-class scatter matrix is
{ \small
\begin{equation}
\mathbf{S}^M_W = {\sum\limits_{i=1}^V \sum\limits_{j=1}^V \W_i^\top
\X_i \bigg(\I-\sum_{c=1}^{C} {1 \over N_c} \e_c{\e_c}^\top\bigg)\X_j ^\top
\W_j}.
\end{equation}
}%
$\W$ contains the eigenvectors of the matrix
$\mathbf{S}={\mathbf{S}^M_W}^{-1}
\mathbf{S}_B^M$ corresponding to the leading $d$ eigenvalues $\lambda_i$.
\section{Generalized Multi-view Embedding}\label{sec:mva}
Here we propose a generalized expression of objective function for multi-view subspace learning. The generalized optimization problem is given by:
\begin{align}
  \J=\arg\max_\W \frac{\Tr(\W^\top \mathbf{P}\W)}
  {\Tr(\W ^\top \Q \W)}\label{eq:rq}
\end{align}
where $\mathbf{P}$ and $\mathbf{Q}$ are the matrices describing the inter-view and intra-view covariances, respectively.
The above equation has the form of the Rayleigh quotient. Therefore, all subspace learning methods that maximize the criterion can be reduced to a generalized eigenvalue problem:
\begin{equation}
  \mathbf{P}{\mathbf{W}}=\rhob \; \mathbf{Q}{\mathbf{W}},\label{eq:gep}
\end{equation}
and the solution is given in \eqref{eq:solution} below:
\begin{equation}
  {\mathbf{W}}=\begin{pmatrix}
    \W_1\\[1mm]
    \vdots\\[1mm]
    \W_V\\[1mm]
  \end{pmatrix}
  \text{ and }
  \rhob = \sum_{i=1}^{d} \lambda_i
  \label{eq:solution}
\end{equation}
are the generalized eigenvector and the sum of the top $d$ generalized eigenvalues $\lambda_i$, respectively. $\W$ contains the projection matrices of all views, and $\bm{\rho}$ is the value of Rayleigh quotient.
We address the Rayleigh quotient as the uniform objective function, reaching out to all subspace learning methods in the paper. The non-linear multi-view embeddings can be achieved by kernel mappings, or (deep) neural networks optimized by SGD. Suppose we have a linear projection $\mathbf{Y}=\W^\top \X$, $\mathbf{S}_{vij}$ is a similarity weight matrix which encodes the intra-view properties to be minimized, and $\mathbf{S}'_{vij}$ is a penalty weight expressing the inter-view properties to be maximized. Then based on \cite{Yan2007, Jia2009}, we can express the objective function as follows
\begin{align}
\J &= \underset{\W^\top \W=\I}{\arg\max}\,
\frac{\sum \limits_{v=0}^{V}\sum \limits_{i=0}^N\sum \limits_{j=0}^N \mathbf{S}'_{vij}\|\W_v^\top \X_{vi}- \W_v^\top \X_{vj}\|^2}
{\sum \limits_{v=0}^{V}\sum \limits_{i=0}^N\sum \limits_{j=0}^N \mathbf{S}_{vij}\|\W_v^\top \X_{vi}- \W_v^\top \X_{vj}\|^2 }\\[1mm]
&=\underset{\W^\top \W=\I}{\arg\max}\, \frac{\Tr(\W^\top \X \mathbf{L}'\X^\top \W)}
{\Tr(\W^\top \X \mathbf{L}\X^\top \W)}.
\end{align}
In the kernel case, we also have
\begin{equation}
  \J = \underset{\mathbf{\A^\top K \A= I}}{\arg\max} \frac{\Tr(\mathbf{\A}^\top \mathbf{K}  \, \mathbf{L}' \, \mathbf{K}\mathbf{\A})}
  {\Tr(\mathbf{\A}^\top \mathbf{K} \, \mathbf{L}\, \mathbf{K}\mathbf{\A})}.
\end{equation}
In the above, we define the diagonal matrix of each view pair as $\mathbf{D}_{uv}$  whose $i$-th element is $[\mathbf{D}_{uv}]_{ii}=\sum_j[\mathbf{S}_{uv}]_{ij}$, and the total graph Laplacian matrix as $\mathbf{L}=\mathbf{D}-\mathbf{S}$. Similarly, we have $\mathbf{D}',\mathbf{S}', \mathbf{L}'$ in the penalty graph. \par
\begin{figure}[h!]
\includegraphics[width=.49\textwidth]{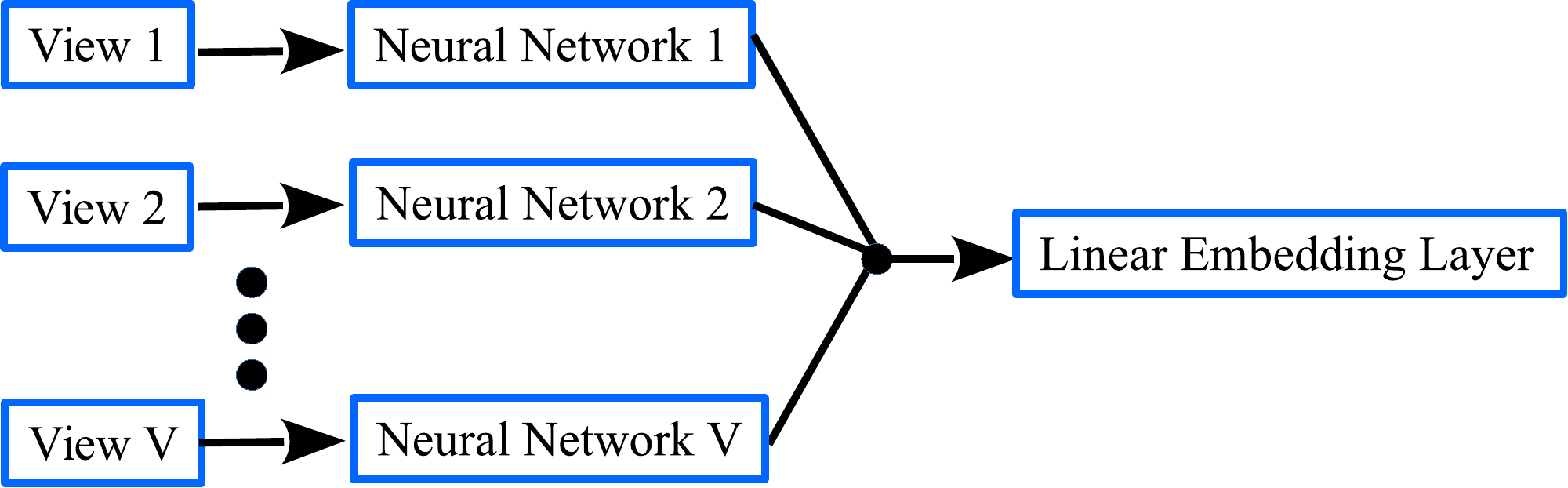}
\caption{An illustration of Multi-view (Deep) Embedding Neural Networks.}\label{fig:dccav}
\end{figure}

For the non-linear mapping by neural networks, we deploy a linear embedding layer on top of the networks. This scheme is illustrated in Fig. \ref{fig:dccav}. Since we have more than two input views, we train multiple neural networks whose outputs are connected to the linear layer and the objective is the same as in the linear case. By backpropagating the error of the weight matrix, we optimize the Rayleigh quotient criterion with respect to the non-linear feature representation from each view in the last hidden layer of the networks. The projection is found in the same way as in the linear case, and we will address the SGD formulation for the specific algortihms in the next section.\par
\begin{figure*}
\centering
\includegraphics[width=.6\textwidth]{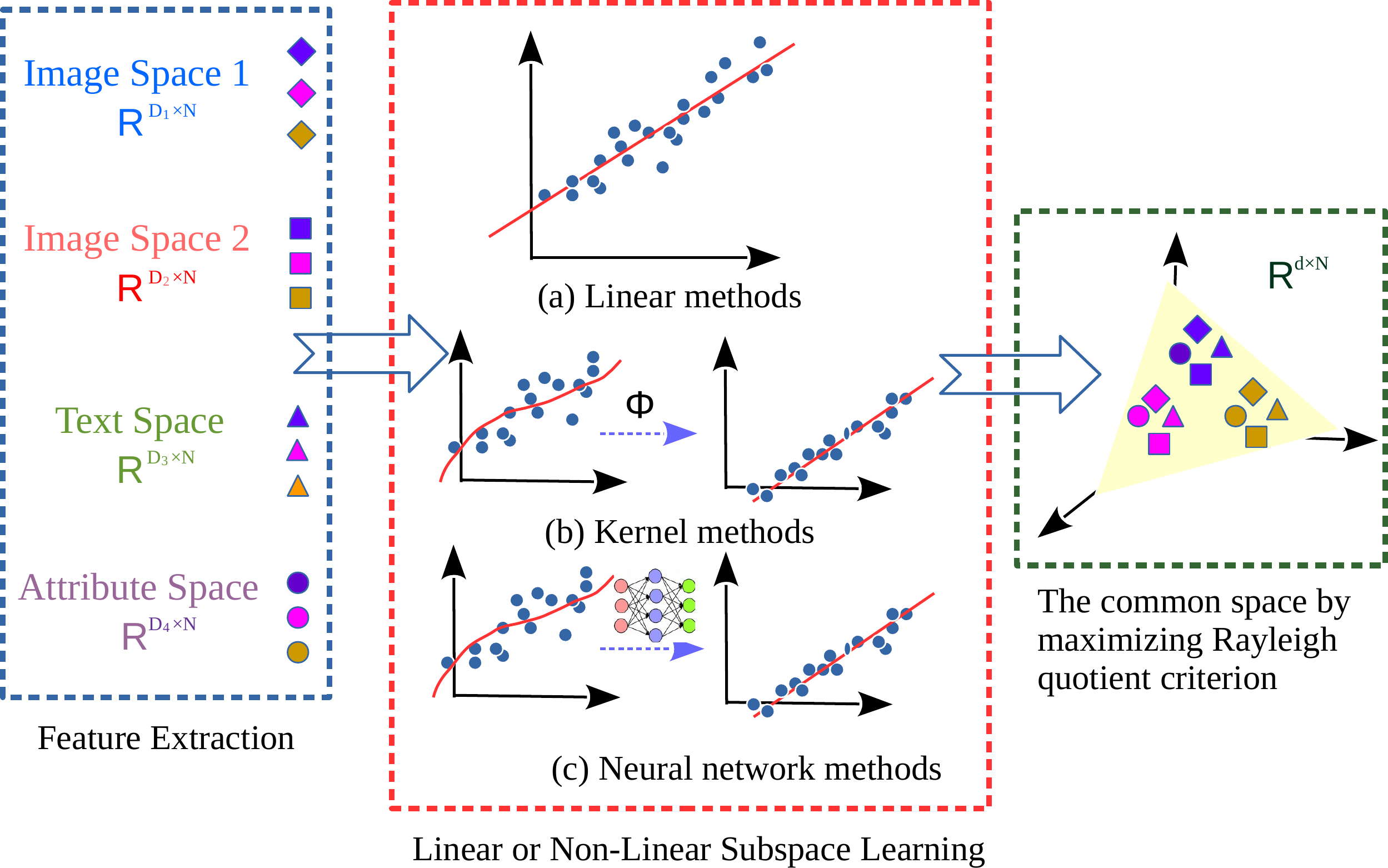}
\caption{Overview of the generalized multi-view embedding: Features from different modalities are extracted and either linearly or nonlinearly mapped into the common subspace by maximizing the Rayleigh quotient criterion.}\label{fig:framework}
\end{figure*}
Fig. \ref{fig:framework} illustrates the proposed framework graphically. We can extract different types of low-level features from images, texts, and intermediate representations. The multi-modal feature vectors are passed through linear or non-linear projections to the latent space. The projected features characterize the properties of the intra-view compactness and inter-view separability based on the proposed criterion.
We show the scaled inter-view and intra-view matrices for each multi-view algorithm in the next section. Then, the projection matrices are presented with respect to their own intrinsic and penalty graph matrices and the optimization methods.
\subsection{Scaling up the inter-view and intra-view covariance matrices}
The idea behind multi-view CCA (MvCCA) is to maximize the correlation between all pairs of views. Its objective can be rephrased as maximizing the inter-view covariance while minimizing the intra-view covariance in the latent space. Therefore, we consider inter-view covariance matrices between different view representations in $\Pb$ and the covariance matrices of each view in $\Q$. Multi-view PLS (MvPLS) maximizes the inter-view covariance directly. Since we also embed the target information for the subspace learning, the proposed MvPLS differs from MvCCA only in the intra-view minimization. Taking the class discrimination into consideration, the novel multi-view modular discriminant analysis (MvMDA) extends to separate the data of different classes between views while making the intra-class data compact. We illustrate the structure of $\Pb$ and $\Q$ for each method in Table \ref{tab:pq}.
\begin{center}
\begin{table}[h!]
\centering
\caption{The matrices $\Pb$ and $\Q$ for the proposed multi-view CCA, PLS and MvMDA.}\label{tab:pq}
\footnotesize
\begin{tabular}{ccc}
&$\mathbf{P}$ & $\mathbf{Q}$\\\hline
\noalign{\smallskip}
MvCCA&
$\begin{bmatrix}
\mathbf{0} &\mathbf{\Sigma}_{12}&\cdots & \mathbf{\Sigma}_{1V}\\[1mm]
\mathbf{\Sigma}_{21} &\mathbf{0}&\cdots & \mathbf{\Sigma}_{2V} \\[1mm]
\vdots&\vdots &\ddots & \vdots \\[1mm]
\mathbf{\Sigma}_{V1}&\mathbf{\Sigma}_{V2}&\cdots &  \mathbf{0}
\end{bmatrix}$
&
$\begin{bmatrix}
\mathbf{\Sigma}_{11}& \mathbf{0}&\cdots & \mathbf{0} \\[1mm]
\mathbf{0} &\mathbf{\Sigma}_{22} &\cdots & \mathbf{0} \\[1mm]
\vdots&\vdots &\ddots & \vdots \\[1mm]
\mathbf{0}&\mathbf{0}&\cdots &  \mathbf{\Sigma}_{VV}
\end{bmatrix}$\\[1mm]
\noalign{\smallskip}
MvPLS&$
\begin{bmatrix}
    \mathbf{0} &\mathbf{\Sigma}_{12}&\cdots & \mathbf{\Sigma}_{1V}\\[1mm]
    \mathbf{\Sigma}_{21} &\mathbf{0}&\cdots & \mathbf{\Sigma}_{2V} \\[1mm]
    \vdots&\vdots &\ddots & \vdots \\[1mm]
    \mathbf{\Sigma}_{V1}&\mathbf{\Sigma}_{V2}&\cdots &  \mathbf{0}
  \end{bmatrix}$
  &
$\begin{bmatrix} \I& \mathbf{0}&\cdots & \mathbf{0} \\[1mm]
\mathbf{0} &\I &\cdots & \mathbf{0} \\[1mm]
\vdots&\vdots &\ddots & \vdots \\[1mm]
\mathbf{0}&\mathbf{0}&\cdots & \I
\end{bmatrix}$\\[1mm]
\noalign{\smallskip}
MvMDA
& $\begin{bmatrix}
\Pb_{11}&\Pb_{12}&\cdots & \Pb_{1V}\\[1mm]
\Pb_{21} &\Pb_{22}&\cdots & \Pb_{21} \\[1mm]
\vdots&\vdots &\ddots & \vdots \\[1mm]
\Pb_{V1}&\Pb_{V2}&\cdots & \Pb_{VV}
\end{bmatrix}$
&
$  \begin{bmatrix}
  \Q_{11}& \mathbf{0}&\cdots & \mathbf{0} \\[1mm]
  \mathbf{0} &\Q_{22} &\cdots & \mathbf{0} \\[1mm]
    \vdots&\vdots &\ddots & \vdots \\[1mm]
    \mathbf{0}&\mathbf{0}&\cdots &  \Q_{VV}
  \end{bmatrix}$
\end{tabular}
\end{table}
\end{center}
\subsection{Linear subspace learning}
\normalsize
\begin{singlespace}
When the subspace projection is linear, we can obtain the latent feature vectors from each view as
\begin{equation}
  \mathbf{Y}_v=\W_v^\top \X_v,\label{eq:lproj}
\end{equation}
and the projection matrix is derived directly by solving the generalized eigenvalue problem in \eqref{eq:gep}.
As shown in Table \ref{tab:pq}, multi-view CCA has the total covariance matrix
$\mathbf{\Sigma}=\mathbf{P}+\mathbf{Q}$, and
we derive its projection matrix by fulfilling the criterion below
\end{singlespace}
\begin{small}
\begin{align}
  \J&=\underset{\W_v, v=1,\dots,V}{\arg\max} \frac{\Tr\bigg({\sum\limits_{i=1}^V \sum\limits_{\substack{j\neq i\\j=1}}^V} \W_i^\top \X_i\, \mathbf{L}\,\X_j^\top \W_j\bigg) }
  {\Tr\big(\sum \limits_{i=1}^V  \W_i^\top \X_i\, \mathbf{L}\,\X_i^\top \W_i\big) }, \label{eq:lincca}
\end{align}
\end{small}
\begin{singlespace}
\noindent where the Laplacian matrix $\mathbf{L}=\I-\frac{1}{N}{\e}\,{\e^\top}$. \par
Multi-view PLS has the same Laplacian matrix as the one in Multi-view CCA.
We only optimize the Rayleigh quotient by maximizing the cross-covariance matrices between different views as
\begin{align}
  \J &=\underset{\W^\top \W = \I}{\arg\max}
 \operatorname{Tr} \, \bigg({\sum\limits_{i=1}^V \sum\limits_{\substack{j\neq i\\j=1}}^V} \W_i^\top \X_i\, \mathbf{L}\,\X_j^\top \W_j \bigg),\label{eq:linpls}
\end{align}
whose solution is the projection matrix.\par
We propose two ways to determine the projection matrix in multi-view LDA. The first appoach is the multi-view extension of the standard LDA, and its
between-class scatter $\mathbf{S}_B$ maximizes the distance between the class means from all views:
\end{singlespace}
\begin{small}
\begin{align}
\mathbf{S}_B&= \sum_{i=1}^V\sum_{j=1}^V\sum_{p=1}^{C}\sum_{\substack{q=1\\p\neq q}}^{C}(\bm{m}_p^i-\bm{m}_q^j)(\bm{m}_p^i-\bm{m}_q^j)^\top \nonumber\\
&=\sum_{i=1}^V\sum_{j=1}^V \W_i^\top{\X}_i \mathbf{L}_B \X_j^\top \W_j,\label{eq:sbm}
\end{align}
\end{small}%
where the between-class Laplacian matrix is
{\small
\begin{equation}
\mathbf{L}_B=
\begin{dcases}
    2 \, \sum_{p=1}^{C}\sum_{\substack{q=1\\p\neq q}}^{C} \left({V \over N_p^2} \e_p \, \e_p^\top - {1 \over N_p N_q}\e_p\, \e_q^\top \right) &\text{ if } i = j,\\
- 2 \, \sum_{p=1}^{C}\sum_{\substack{q=1\\p\neq q}}^{C}{1 \over N_p N_q}\e_p \,\e_q^\top &\text{ if } i \neq j.
\end{dcases}\label{eq:lb1}
\end{equation}}%
$\bm{m}_p^i$ denotes the mean from the $i$th view of the $p$th class in the latent space, and $\e_p$ is the $N$-dimensional class vector, with $N_p$ as the number of samples in the $p$th class. The class $q$ is different from the class $p$. \par
\begin{singlespace}
Alternatively, we propose the between-class scatter matrix which maximizes the distance between different class centers across different views. Since it considers the samples from the class of the specific view origin, we call it Multi-view Modular Discriminant Analysis (MvMDA), and its forumulation is
\end{singlespace}
\begin{small}
\begin{align}
  \mathbf{S}'_B&=  \sum_{i=1}^V\sum_{j=1}^V\sum_{p=1}^{C}\sum_{\substack{q=1\\p\neq q}}^{C}(\bm{m}_p^i-\bm{m}_q^i)(\bm{m}_p^j-\bm{m}_q^j)^\top\nonumber\\
&=\sum_{i=1}^V\sum_{j=1}^V \W_i^\top{\X}_i \mathbf{L}'_B \X_j^\top \W_j,\label{eq:sb2}
\end{align}
\end{small}
and the Laplacian matrix is
\begin{equation}
\mathbf{L}'_B = 2 \, \sum_{p=1}^{C}\sum_{q=1}^{C}({1 \over N_p^2} \e_p \, \e_p^\top - {1 \over N_p N_q} \e_p\,\e_q^\top).\label{eq:lb2}
\end{equation}
\begin{singlespace}
The difference between the two approaches is that $\mathbf{S}_B$ has ${1 \over N_c^2}(V-1)\sum \limits_{i=1}^V\sum \limits_{c=1}^C \W_i^\top \X_i \e_c \e_c^\top \X_i^\top \W_i$,
while $\mathbf{S}'_B$ has the term ${1 \over N_c^2} \sum \limits_{i=1}^V\sum \limits_{\substack{j=1\\j\neq i}}^V\sum \limits_{c=1}^C \W_i^\top \X_i \e_c \e_c^\top \X_j^\top \W_j$
which suggests that  the first proposal only considers the maximum of the intra-view distances,
while the second proposal can maximize the distance between different views. We also validate experimentally that the second proposal achieves better results. Detailed derivation of the two approaches of \eqref{eq:sbm} and \eqref{eq:sb2} are included in the supplementary material.
\end{singlespace}
We extend the same formulation of within-class Laplacian matrix in the latent space as the single-view LDA, i.e.
\begin{small}
\begin{align}
\mathbf{S}_W & =  \sum_{i=1}^V \W_i^\top \X_i\bigg(\I-\sum_{c=1}^{C} {1 \over N_c} \e_c{\e_c}^\top\bigg)\X_i^\top \W_i \nonumber\\
& =  \sum_{i=1}^V \W_i^\top \mathbf{Q}_{ii} \; \W_i,\label{eq:swm}
\end{align}
\end{small}
\begin{singlespace}
\noindent
where $\Q_{ii}=\X_i\mathbf{L}_W  \X_i^\top$, and $\mathbf{L}_W=\I-\sum \limits_{c=1}^{C} {1 \over N_c} \e_c{\e_c}^\top$.
From \eqref{eq:sbm} and \eqref{eq:swm}, it is shown that the between-class and within-class scatters are equivalent to the projected inter-view and intra-view covariance, respectively.
The projection matrix of the multi-view LDA is found by optimizing the following objective function
\end{singlespace}
\footnotesize
\begin{equation}
\J= \underset{\W_v,v=1,\dots,V}{\arg\max}
\frac{\Tr \big(\sum\limits_{i=1}^V\sum \limits_{j=1}^V \W_i^\top{\X}_i \mathbf{L}^*_B \X_j^\top \W_j
\big)}
{\Tr\big(\sum \limits_{i=1}^V \W_i^\top \X_i \mathbf{L}_W  \X_i^\top \W_i\big)},
\end{equation}
\normalsize
where $\mathbf{L}_B^*$ is denoted as the Laplacian matrix of either $\mathbf{L}_B$ or $\mathbf{L}_B'$.
\subsection{Kernel-based non-linear subspace learning}
Exploiting the kernel trick in \eqref{eq:kernel} and the Representer theorem in \eqref{eq:weight} and \eqref{eq:lproj} can be expressed as follows
\begin{equation}
\mathbf{Y}_v=\A_v^\top \bm{\Phi}_v^\top\bm{\Phi}_v=\A_v^\top \K_v.
\end{equation}
The criterion of kernel multi-view CCA is then,
\begin{small}
\begin{equation}
  \J=\underset{\K_v,v=1,\dots,V}{\arg\max}\,\frac{\Tr\big({\sum\limits_{i=1}^V \sum\limits_{\substack{j\neq i\\j=1}}^V} \A_i^\top \K_i\, \mathbf{L}\,\K_j \A_j\big) }
 {\Tr\big(\sum \limits_{i=1}^V  \A_i^\top \K_i\, \mathbf{L}\,\K_i \A_i \big)}.\label{eq:mvkcca}
\end{equation}
\end{small}
\begin{singlespace}
It can be easily shown that the solution for $\A_v$ is the same as \eqref{eq:gep}.\par
Kernel multi-view PLS maximizes the covariance between pairs of feature vectors in the kernel space and therefore the objective function is
\end{singlespace}
\begin{small}
\begin{align}
\J = \underset{K_v,v=1,\dots,V}{\arg\max}
\operatorname{Tr}\Bigg(\,{{\sum\limits_{i=1}^V
\sum\limits_{\substack{j\neq i\\j=1}}^V} \A_i^\top \K_i\, \mathbf{L}\,\K_j \A_j }\Bigg).\label{eq:mvkpls}
\end{align}
\end{small}
\begin{singlespace}
\noindent
The criterion for kernel multi-view discriminant analysis is
\end{singlespace}
\small
\begin{equation}
\J = \underset{K_v,v=1,\dots,V}{\arg\max} \, \frac{\Tr \big(\sum \limits_{i=1}^V\sum \limits_{j=1}^V \A_i^\top{\K}_i \mathbf{L}^*_B \K_j\A_j
\big)}
{\Tr\big(\sum \limits_{i=1}^V \A_i^\top \K_i\mathbf{L}_W  \K_i \A_i\big)}\label{eq:mvkda}
\end{equation}
\normalsize
\subsection{Non-linear subspace learning using (deep) neural networks}
Exploiting the non-linear mapping using neural networks by \eqref{eq:nnproj}, \eqref{eq:lproj} can expressed as
\begin{equation}
\mathbf{Y}_v = \W_v^\top h(\X_v;\B_v)=\W_v^\top \Hb_v.
\end{equation}
Since the network outputs are combined by a linear layer as shown in Fig. \ref{fig:dccav}, the parameters $\B_v$ of each network are jointly trained to reach the optimal criterion value. After the transformation by neural networks, the projection becomes the same as the multi-view linear subspace learning with respect to $\Hb_v$. Therefore, we need an additional optimization solved by SGD.
We experimented with SGD without variance constraints, and found that we could obtain much better results with the projections constrained to have the unit variance, i.e. in Deep Multi-view CCA (DMvCCA), we have
\begin{equation}
\sum \limits_{i=1}^V  \W_i^\top \Hb_i\, \mathbf{L}\,\Hb_i^\top \W_i=\I.
\end{equation}
Without intra-view minimization, the optimization of Deep Multi-view PLS (DMvPLS) is constrained to have unit variance
$\sum \limits_{i=1}^V  \W_i^\top  \W_i=\I$,
while in Deep Multi-view Modular Discriminant Analysis (DMvMDA), we project the within-class scatter into unit, i.e.
\begin{equation}
\sum \limits_{i=1}^V \W_i^\top \Hb_i\mathbf{L}_W  \Hb_i^\top \W_i=\I
\end{equation}
With the variance constraint, the expressions of the gradients in DMvCCA and DMvPLS are the same as
\begin{small}
\begin{align}
\frac{\partial \J}{\partial \Hb_i}
&=\frac{\partial}{\partial \Hb_i} \operatorname{Tr}\Bigg(\,{\sum\limits_{i=1}^V \sum\limits_{\substack{j\neq i\\j=1}}^V} \W_i^\top \Hb_i\, {\mathbf{L}}\,\Hb_j^\top \W_j \Bigg)\nonumber \\
& ={\sum\limits_{i=1}^V \sum\limits_{\substack{j\neq i\\j=1}}^V}
\W_i \,\W_j^\top \Hb_j \,\mathbf{L},\label{eq:ccagrad}
\end{align}
\end{small}
and the gradient of DMvMDA is computed as
\begin{small}
\begin{align}
\frac{\partial \J} {\partial \Hb_i}
&=\frac{\partial}{\partial \Hb_i} \operatorname{Tr}\Bigg(\,{\sum\limits_{i=1}^V \sum\limits_{j=1}^V} \W_i^\top \Hb_i\, {\mathbf{L}^*_B}\,\Hb_j^\top \W_j \Bigg) \nonumber\\
 &={\sum\limits_{i=1}^V \sum\limits_{j=1}^V}
 \W_i \,\W_j^\top \Hb_j \,{\mathbf{L}^*_B},\label{eq:ldagrad}
\end{align}
\end{small}%
Detailed derivation of \eqref{eq:ccagrad} and \eqref{eq:ldagrad} can be found in the supplementary material.
\section{Experiments}\label{sec:exp}
In this section, we evaluate the multi-view methods on two important multimedia applications: zero-shot recognition on the Animal with Attribute (AwA) dataset, and cross-modal image retrieval on the Wikipedia and Microsoft-COCO datasets.
\subsection{Experimental Setup}
We conduct the experiments on three popular multimedia datasets. One common property in these datasets is that multi-modal feature representations can be generated. The Animal with Attribute (AwA) dataset consists of $50$ animal classes with $30,475$ images in total, and $85$ class-level attributes. We follow the same setup as in \cite{Lampert2014} by splitting 40 classes ($24,295$ images) to train the categorical model while the rest $10$ classes with $6,180$ images for testing. Sample images from the test set are shown in Fig. \ref{fig:map}. Each animal class contains more than one positive attribute, and the attributes are shared across classes which enables zero-shot recognition. The detailed class labels and attributes are provided in \cite{Lampert2014}.\par
Wikipedia is a cross-modal dataset collected from the ``Wikipedia featured articles'' \cite{CostaPereira2014}. The dataset is organized in 10 categories and consists of $2,866$ documents. Each document is a short paragraph with a median text length of $200$ words, and is associated with a single image. We follow the train/test split in \cite{CostaPereira2014} who use $2,173$ training and $693$ test pairs of images and documents. \par
The third dataset we use is the Microsoft COCO 2014 Dataset \cite{Lin2014} (abbreviated as COCO in latter paragraphs). We collect the images belonging to at least one fine-grained category, which amounts to $82,081$ training images, and $40,137$ validation images. More than 5 human-annotated different captions are associated to each image. We follow the same definition in \cite{Lin2014} to use 12 super classes as the class labels, and 91 fine-grained categories as the attributes. The class names and attributes are presented in Table \ref{tab:cococls}. The classes that the images belong to are highly semantic, and the same image can have multiple class labels. Meanwhile, similar images may belong to several different classes.\par
\begin{table}
[h!]
\centering
\caption{The class labels and attributes on the COCO dataset.}\label{tab:cococls}
\begin{tabular}{|p{.45\textwidth}|}
\hline
Classes\\
\hline
outdoor, food, indoor, appliance, sports, person, animal, vehicle, furniture, accessory, electronic, kitchen\\
\hline \hline
Attributes\\
\hline
person, bicycle, car, motorcycle, airplane, bus, train, truck, boat, traffic, light, fire, hydrant, stop, sign, parking, meter, bench, bird, cat, dog, horse, sheep, cow, elephant, bear, zebra, giraffe, backpack, umbrella, handbag, tie, suitcase, frisbee, skis, snowboard, sports, ball, kite, bat, baseball, glove,
skateboard, surfboard, tennis, racket, bottle, wine, glass, cup, fork, knife, spoon, bowl, banana, apple, sandwich, orange, broccoli, carrot, hot dog, pizza, donut, cake, chair, couch, potted, plant, bed, dining, table, toilet, tv, laptop, mouse, remote, keyboard, cell phone, microwave, oven, toaster, sink, refrigerator, book, clock, vase, scissors, teddy, bear, hair, drier, toothbrush\\
\hline
\end{tabular}
\end{table}
We use the following feature representations in the experiments:
\begin{itemize}
\item \textbf{Image feature by CNN models}: We employ the off-the-shelf CNN models as stated in \cite{Razavian2014} and \cite{Wei2016} on all image datasets --- Visual features are extracted by adopting two powerful pre-trained models. We rescale the size of the input images to $224 \times 224$, and generate the features from the outputs of the $fc8$ layer in a VGGNet with 16 weight layers \cite{Simonyan2015} (denoted as \emph{VGG-16} in latter sections), and the $loss3/classifier$ layer from a GoogleNet \cite{Szegedy2015}. Both models produce $1000$-dimension feature vectors.
\item \textbf{Class label encoding}: Since each image corresponds to one class label on the AwA and Wikipedia dataset, we can describe the image category using the textual feature mapped from the image feature. Specifically, we firstly train a 100-dimension skip-gram model \cite{Mikolov2013} on the entire English Wikipedia articles composed of 2.9 billion words. Then we can extract a separate set of word vectors from class labels of our datasets. In order to correlate the labels with the image contents, we train a ridge regressor with 10-fold cross-validation to map the \emph{VGG-16} image features to each dimension of the word vectors respectively. The regressor outputs are used as the class label features.
\item \textbf{Attribute encoding}: We also adopt another important modality from visual attributes on the AwA and COCO datasets. On the AwA dataset, we use the $50\times 85$ class-attribute matrix in \cite{Kemp2006, Osherson1991} which specifies attribute probabilities of each class, while on the COCO dataset, we develop a 91-bin feature vector as attributes for each image of which $1$'s denote the image has the fine-grained tag and $0$'s otherwise. Then, we train a ridge regressor between the \emph{VGG-16} image feature and formulated attribute probabilities. The predicted probabilities associated with each image are used as the attribute feature.
  \item \textbf{Sentence encoding}: A vital feature of cross-modal retreival system is that we make use of textual features directly. We can find a paragraph of text describing each image on the Wikipedia dataset, while on the COCO dataset, a similar paragraph can be developed by concantenating all captions from the annotators which are associated to each image. We generated the sentence vectors from the paragraphs by the pre-trained skip-thoughts model \cite{Kiros2015}. The model was trained over the MovieBook and BookCorpus dataset \cite{zhu2015}. On the Wikipedia, we employ the \emph{combined-skip} vector of $4800$ dimensions, while due to the large size of COCO dataset, we only use the \emph{uni-skip} vector of $2400$ dimensions.
\end{itemize}
The Experiment protocol and performance metrics are described below:
\begin{itemize}
\item \textbf{Zero-shot recognition on the AwA dataset}:
We follow a similar experiment pipeline as in \cite{Fu2015}, and the comparative results show the performance of the proposed multi-view embedding methods. We project the multi-view representations to the latent space. Zero-shot recognition is achieved by semi-supervised label propagation on a transductive hypergraph in the latent space. Specifically, the cross-domain knowledge learned from the common semantic space is tranferred to the target space of 10 test animal classes via attributes. The prediction of target classes is undertaken on a hypergraph to better integrate different views. We replace the multi-view linear CCA for joint embedding in \cite{Fu2015} with the generalized embedding methods. Since the same hypergraph is used, the recognition results indicate the different performance by the multi-view methods in this paper.
For the evaluation metric, we use the average classification accuracy which is also employed in \cite{Lampert2014, Fu2015}.
\item \textbf{Cross-modal retrieval on the Wikipedia and COCO datasets}:
We perform two tasks in cross-modal retrieval, i.e. text query for image retrieval and image query for text retrieval. Moreover, a conventional content-based image retrieval system is evaluated in Section \ref{sec:cbir}.
We first extract the test features in their own domains. A latent space is joinly learned from the image features, intermediate feature and sentence feature in the training set. Test features are then projected to the latent space by the trained model. The semantic matching from \cite{CostaPereira2014} is performed by training a logistic regressor over the embedded features from all of the ground truth samples which maps the projected features of both queries and to-be-retrieved images/texts towards the class labels.
The feature vectors generated from the ground truth class labels are essentially the class vectors, whose dimensionality is the number of classes. We use the class probabilities from the regressor outputs for matching between modalities.\par
We present the results using 11-point interpolated precision-recall (PR) curves. The {Mean Average Precision} (MAP) score, which is the average precision at the ranks where recall changes, can be computed based on the Precision Recall curves. The Average Precision (AP) measures the relevance between a query and retrieved items \cite{Manning2008a}, and the MAP score calculates the mean AP by querying all items in the test set. 
\end{itemize}
\subsection{Parameter Settings}
The dimensionality $d$ in the latent space is a pre-defined parameter. We will evaluate the effects of different $d$ values in the following section. In the experiment, we use $d=50$ for linear projections on all datasets. On the Wikipedia and AwA dataset, we choose $d=150$ for kernel mappings, and $d=200$ for the COCO dataset. For computational efficiency on the AwA and COCO dataset, an approximated RBF kernel mapping is adopted for the non-linear mappings. We set $\sigma$ in the RBF kernel as the average distance between samples from different views/modalities, which is the natural scaling factor for each dataset. In all of the experiments, the original training set is further partitioned into a $80\%$ training split and a $20\%$ validation split.\par
The topology of neural networks has more variabilities, and we chose the optimal one according to the held-out validation set. We refer to \cite{Yao1999, kiranyaz2009} for a detailed discussion on topologies.
On the AwA dataset, we took $3$ hidden layer, each with $1,024$ neurons with the \emph{relu} activation before the $50$-dimensional linear embedding layer. We only adopted the linear and kernel-based embeddings on the Wikipedia dataset in view of its small size. On the COCO dataset, we chose a single hidden layer with $1500$ \emph{relu} neurons, and the dimensionality of the final linear layer is also $1500$. We experimented both with the whole batch and multiple mini-batches for SGD, and adopted a batch size of $200$ which achieves a superior performance. The number of epoches is set to $50$ empirically.
\begin{figure}
\centering
  \begin{subfigure}[b]{.32\linewidth}
\includegraphics[width=\textwidth]{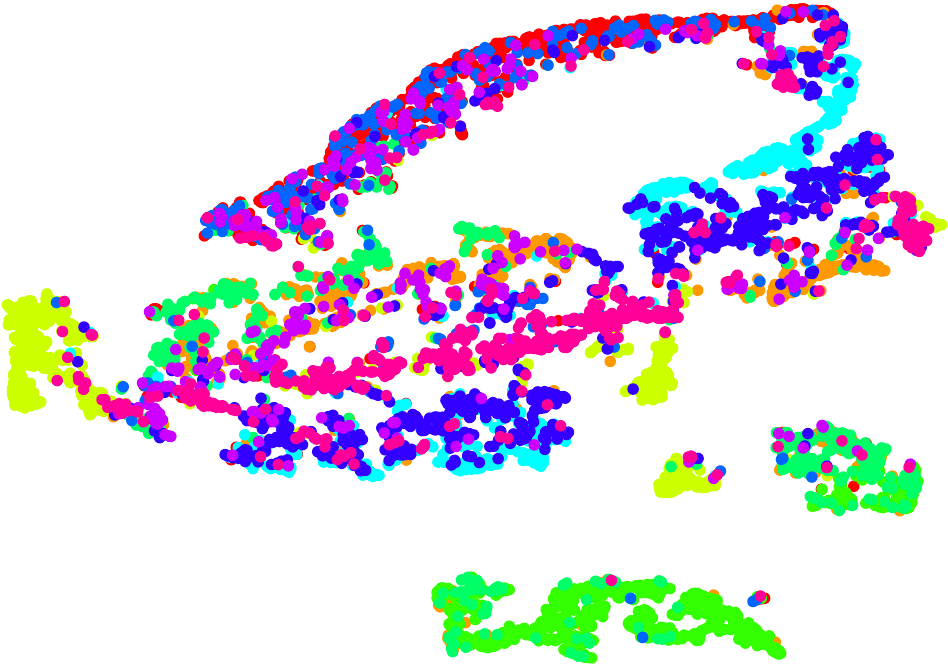}
\caption{\footnotesize 2-view LMvCCA}
\end{subfigure}
\begin{subfigure}[b]{.32\linewidth}
\includegraphics[width=\textwidth]{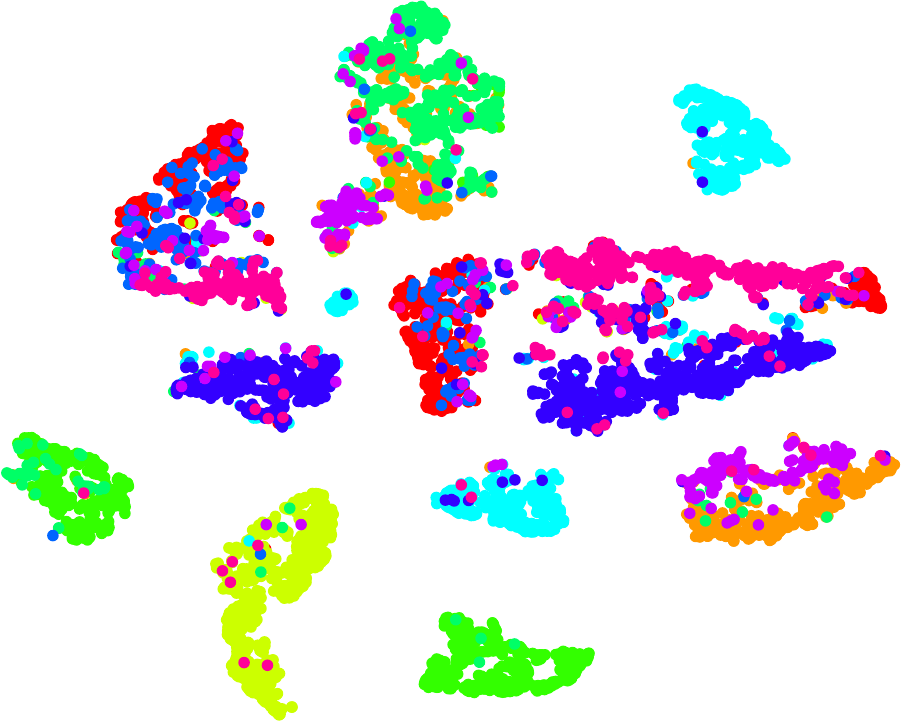}
\caption{\footnotesize 3-view LMvCCA}
\end{subfigure}
\begin{subfigure}[b]{.32\linewidth}
\includegraphics[width=\textwidth]{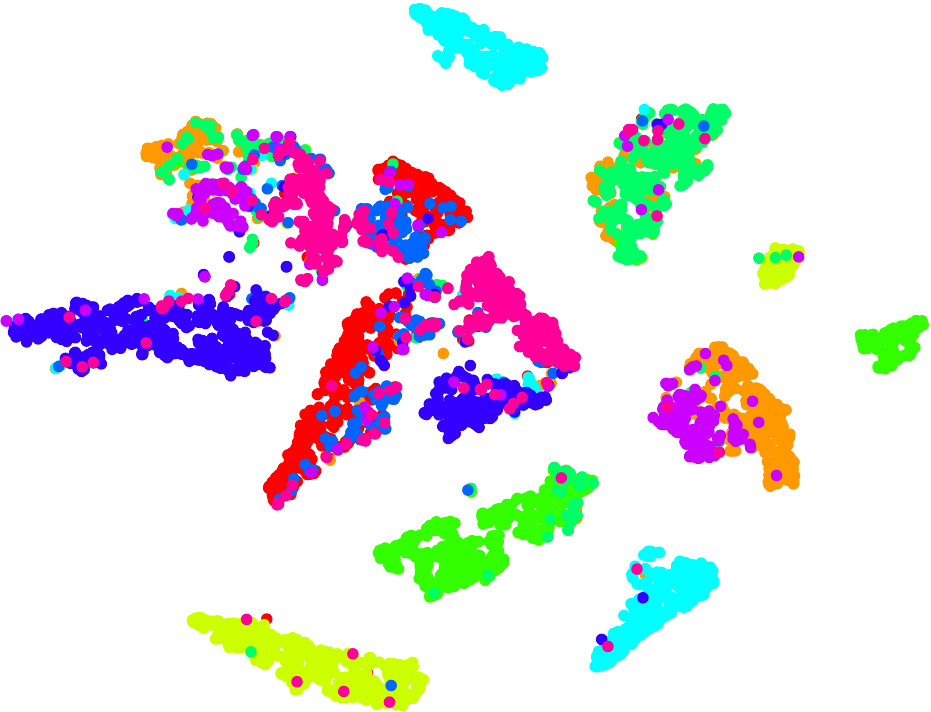}
\caption{\footnotesize 4-view LMvCCA}
\end{subfigure}\\[5mm]
\begin{subfigure}[b]{.32\linewidth}
\includegraphics[width=\textwidth]{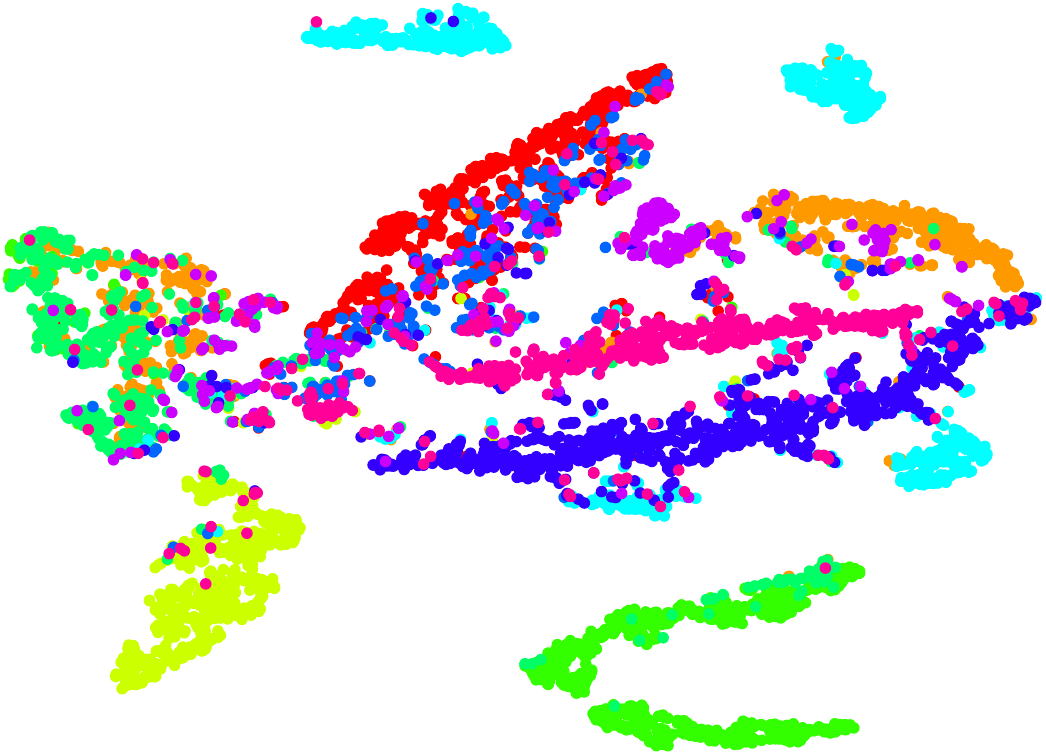}
\caption{\footnotesize 4-view MvDA \cite{Kan2016}}
\end{subfigure}
\begin{subfigure}[b]{.32\linewidth}
\includegraphics[width=\textwidth]{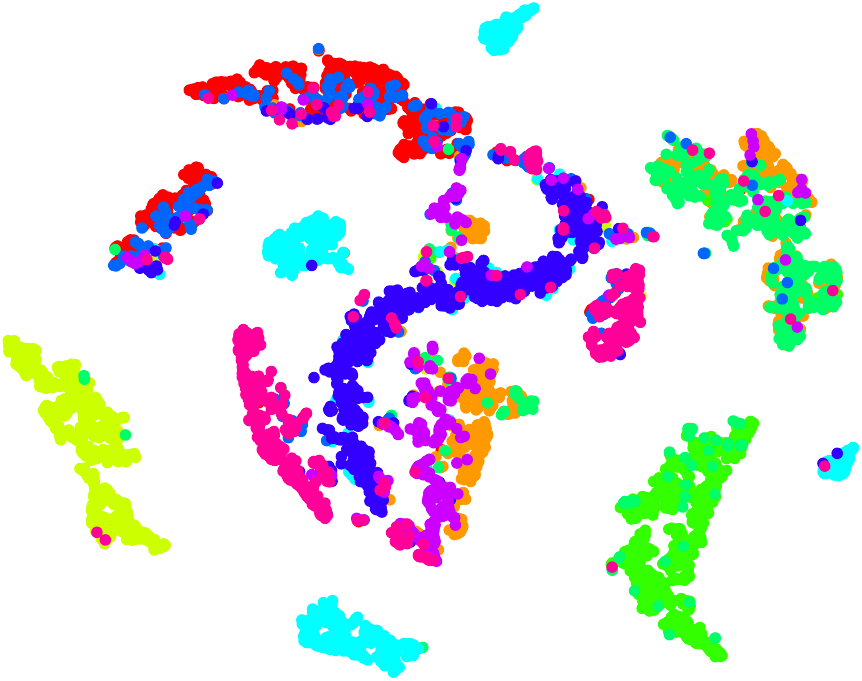}
\caption{\footnotesize 4-view GMA \cite{Sharma2012}}
\end{subfigure}
\begin{subfigure}[b]{.32\linewidth}
\includegraphics[width=\textwidth]{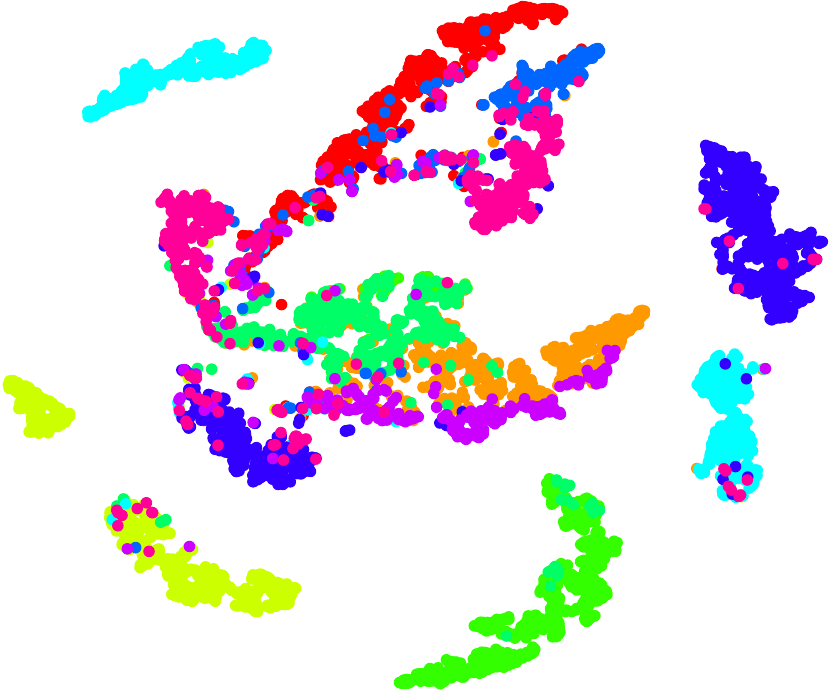}
\caption{\footnotesize 4-view DMvMDA}
\end{subfigure}\\[1mm]
\caption{The first row shows the 2-D visualization of embeddings by LMvCCA with an increasing number of views on the AwA dataset. The second row presents the embedding maps by different methods all with 4 views on the same dataset. The samples from different classes are denoted in different colors.}\label{fig:views}
\end{figure}
\subsection{Experimental Results}
\begin{table*}
[h!]
\begin{center}
\caption{List of Abbreviations}\label{tab:abb}
\resizebox{.9\linewidth}{!}{
\begin{tabular}{|l | p{15cm} | }
    \hline
LMvCCA / KMvCCA / KapMvCCA / DMvCCA & Linear / Kernel / Approximate Kernel / Deep Multi-view Canonical Correlation Analysis \\ \hline
LMvPLS / KMvPLS / KapMvPLS / DMvPLS & Linear / Kernel / Approximate Kernel / Deep Multi-view Partial Least Square Regression \\ \hline
SLMvDA/ SKMvDA & Standard Linear / Kernel Multi-view Discriminant Analysis using \eqref{eq:lb1}\\ \hline
LMvMDA / KMvMDA / KapMvMDA / DMvMDA & Linear / Kernel / Approximate Kernel / Deep Multi-view Modular Discriminant Analysis using \eqref{eq:lb2}\\ \hline
MULDA / KMUDA \cite{Sun2016} & Multi-view Uncorrelated Linear / Kernel Discrimiant Analysis \\ \hline
MvDA \cite{Kan2016} & Multi-view Discrimiant Analysis \\ \hline
GMA \cite{Sharma2012} & Generalized Mult-view Analysis\\ \hline
DCCA2 \cite{Andrew2013} & Deep Canonical Correlation Analysis\\ \hline
\end{tabular}
}
\end{center}
\end{table*}
The abbreviations of the numerous methods are shown in Table \ref{tab:abb}.
\subsubsection{Results on zero-shot recogntion}
\begin{table}[h!]
\centering
\caption{RECOGNITION ACCURACY (\%) on the AwA DATASET}\label{tab:awa}
\resizebox{0.8\linewidth}{!}{
\begin{tabular}{|l| c| c| c| }
\hline
Method & 2 views & 3 views & 4 views\\ \hline\hline
Proposed LMvCCA &55.86 & 75.88 & {82.01}\\
Proposed LMvPLS & \textbf{58.52} & 73.59 & 77.09 \\
Proposed LMvMDA &55.85  & \textbf{77.64} & \textbf{82.88}\\
Proposed SLMvDA &54.58&69.02& 70.56\\ \hline
Proposed KapMvCCA &56.41 &73.40  & 74.76\\
Proposed KapMvPLS &55.58 &\textbf{74.40} &  75.05\\
Proposed KapMvMDA &\textbf{57.19}  &71.64&  \textbf{75.63}\\\hline
Proposed DMvCCA &51.25 &71.12 &82.27\\
Proposed DMvPLS &43.28 &68.81 &74.63\\
Proposed DMvMDA & \textbf{53.87}&\textbf{75.61} & \textbf{83.66} \\ \hline
MvDA \cite{Kan2016}&49.95 &68.55 &70.00 \\
GMA \cite{Sharma2012}&52.12&73.49&78.46\\
MULDA \cite{Sun2016}&55.46 &74.13 &74.88\\
TMV-HLP \cite{Fu2015} & - &73.50& 80.50\\
DCCA2 \cite{Andrew2013}&50.47&-&-\\\hline
\end{tabular}}
\end{table}
We visualize the embedded space in Fig. \ref{fig:views}. We use the VGG-16 feature
and class label encoding for two views, and augment attribute
and GoogleNet encodings as the additional views. In the first row, it is shown with the increasing number of views in MvCCA, the latent feature vector progresses from being distributed incoherently to showing more distinct groups. In the second row, we compare different methods with 4 views. It is clearly shown we obtain a set of more compact and separable features by the proposed DMvMDA.\par
Recognition accuracy of different methods is compared quantitatively in Table \ref{tab:awa}. The first group contains the linear projection results, the second uses the kernel methods, the third are the results by deep neural nets, and the last category includes several comparative results in the literature. The linear methods perform favorably in general while the leading recognition rates can be found in the non-linear methods using neural nets with 4 views. The kernel approximation does not provide superior results compared to linear methods due to the information loss in sampling \cite{Rahimi2007}. Above all, the 4-view DMvMDA is reported to be the best method for zero-shot recognition. The results are also organized by the number of views in columns, and it is shown for all methods that we consistently obtain a better accuracy with more views. Specifically, the proposed LMvPLS achieves the highest accuracy with two input views. while the novel LMvMDA has a more discriminant representation in the latent space leading to a better recognition when more views are presented.
\begin{table*}
   \begin{center}
   \caption{MAP Scores (\%) on the Wikipedia}\label{tab:wiki}
  \resizebox{0.8\linewidth}{!}{
\begin{tabular}{|l|c|c| c|c|c| c|c|c|c| }\hline
 & \multicolumn{3}{ |c| }{2 views} & \multicolumn{3}{ |c| }{3 views} & \multicolumn{3}{ |c| }{4 views}\\ \hline\hline
& img. query & txt. query & avg. & img. query & txt. query & avg.& img. query & txt. query & avg.\\ \hline
MvDA \cite{Kan2016}& 39.73  & 37.14  & 38.43 & 39.34  & 35.04  & 37.19  & 41.07  & 39.21  & 40.14\\
GMA \cite{Sharma2012}& 41.91&   38.55&  40.23&  42.26&  38.66 & 40.46 & 42.26&  38.67 &  40.47\\
MULDA \cite{Sun2016} & 43.04 &39.87 & 41.46 & 43.45 & 40.68& 42.07 & 43.79 & 40.32 &  42.06 \\
Proposed LMvCCA & 41.37  & 39.07 & 40.22  & 42.10 & 39.64 & 40.87 & 42.53 & 39.98 & 41.26\\
Proposed LMvPLS & 42.49 & 40.42  & 41.46  & 41.29  & 39.34  & 40.31  & 41.86 & 39.74 & 40.80 \\
Proposed SLMvDA & 43.20  & 40.07 & 41.64 & 43.14 &  39.86& 41.50 & 43.77 & 40.24&  41.80\\
Proposed LMvMDA & 43.38 & 40.32 & \textbf{41.85}& 43.74 & 40.46& \textbf{42.10} & 43.90  & 40.23 &  \textbf{42.07}\\\hline
KMUDA \cite{Sun2016} & 44.38  & 39.52  &41.95 & 45.40  & 39.96  & 42.68 & 44.29 & 38.12  & 41.20\\
Proposed KMvCCA & 44.78 &  41.83  &43.30 &  44.06  & 41.41  & 42.73 &45.13  & 41.66 & 43.40\\
Proposed KMvPLS    & 42.94  &40.46   & 41.70 & 42.03  & 39.40  & 40.71  & 41.94   & 38.84  & 40.39\\
Proposed SKMvDA & 45.52   & 38.39   & 41.96 & 44.66   & 38.47  & 41.57 & 42.94  & 39.32  & 41.13 \\
Proposed KMvMDA & 46.01  & 40.96  &\textbf{43.49} & 45.40  & 40.16  & \textbf{42.78} & 46.48 & 40.73  & \textbf{43.61}\\ \hline
\end{tabular}}
\end{center}
\end{table*}
\subsubsection{Cross-modal retrieval results on the Wikipedia Dataset}
Due to the limited number of samples, we use PCA before performing the subspace learning. We use the \emph{VGG-16} and sentence features for two views, and augment attribute and GoogleNet encodings as the additional modalities. It is shown that a better MAP score is obtained when enriching the latent feature with more modalities as shown in Table \ref{tab:wiki}. We also observe that the supervised methods perform better than the unsupervised counterparts, and non-linear projections by kernel methods are superior. KMvMDA achieves the best retrieval results with supervision and non-linearity.\par
\begin{figure*}
\begin{subfigure}[b]{.3\linewidth}
\includegraphics[width=1\textwidth]{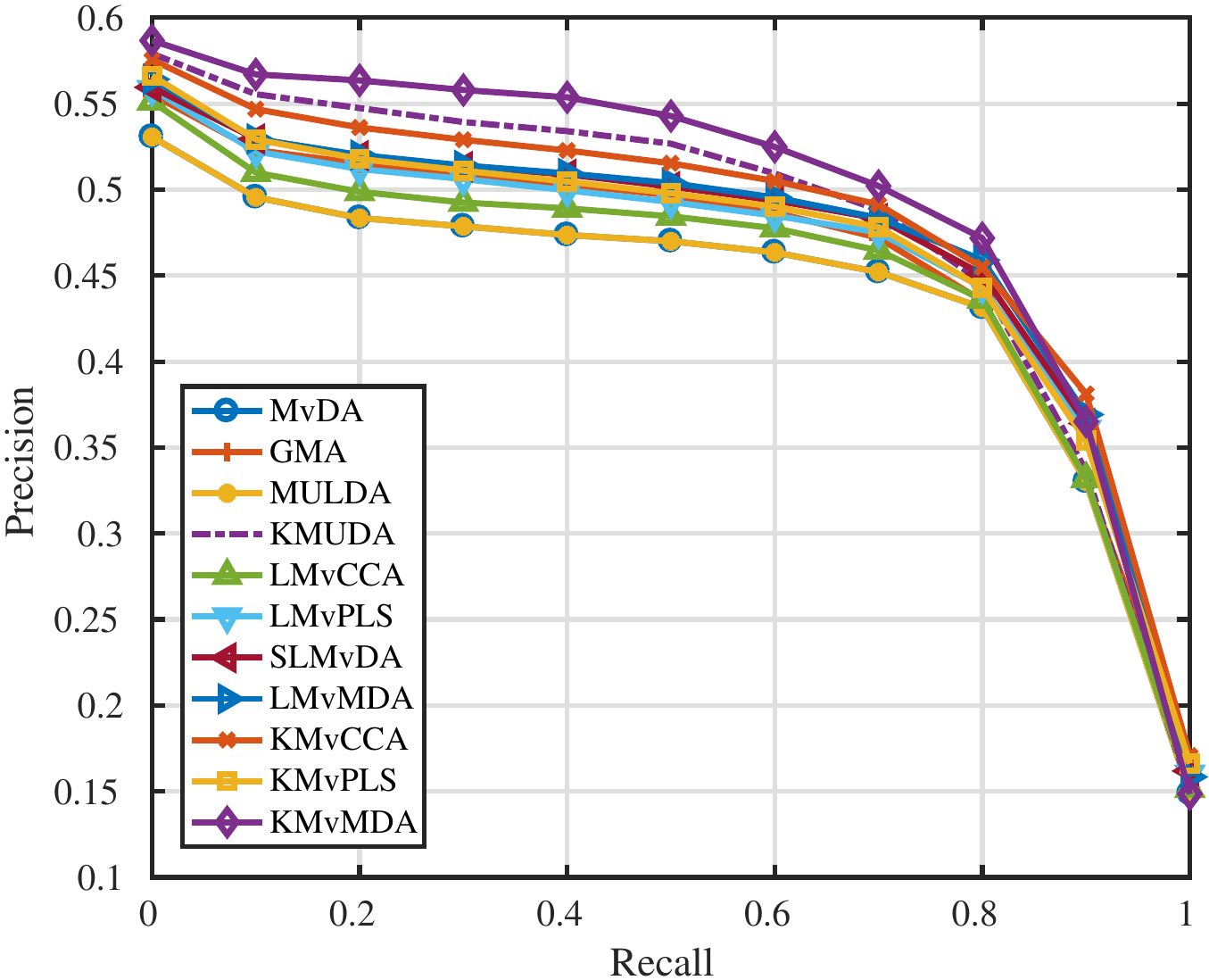}
\caption{PR curves for image queries with 2 views}
\end{subfigure}
\begin{subfigure}[b]{.3\linewidth}
\includegraphics[width=1\textwidth]{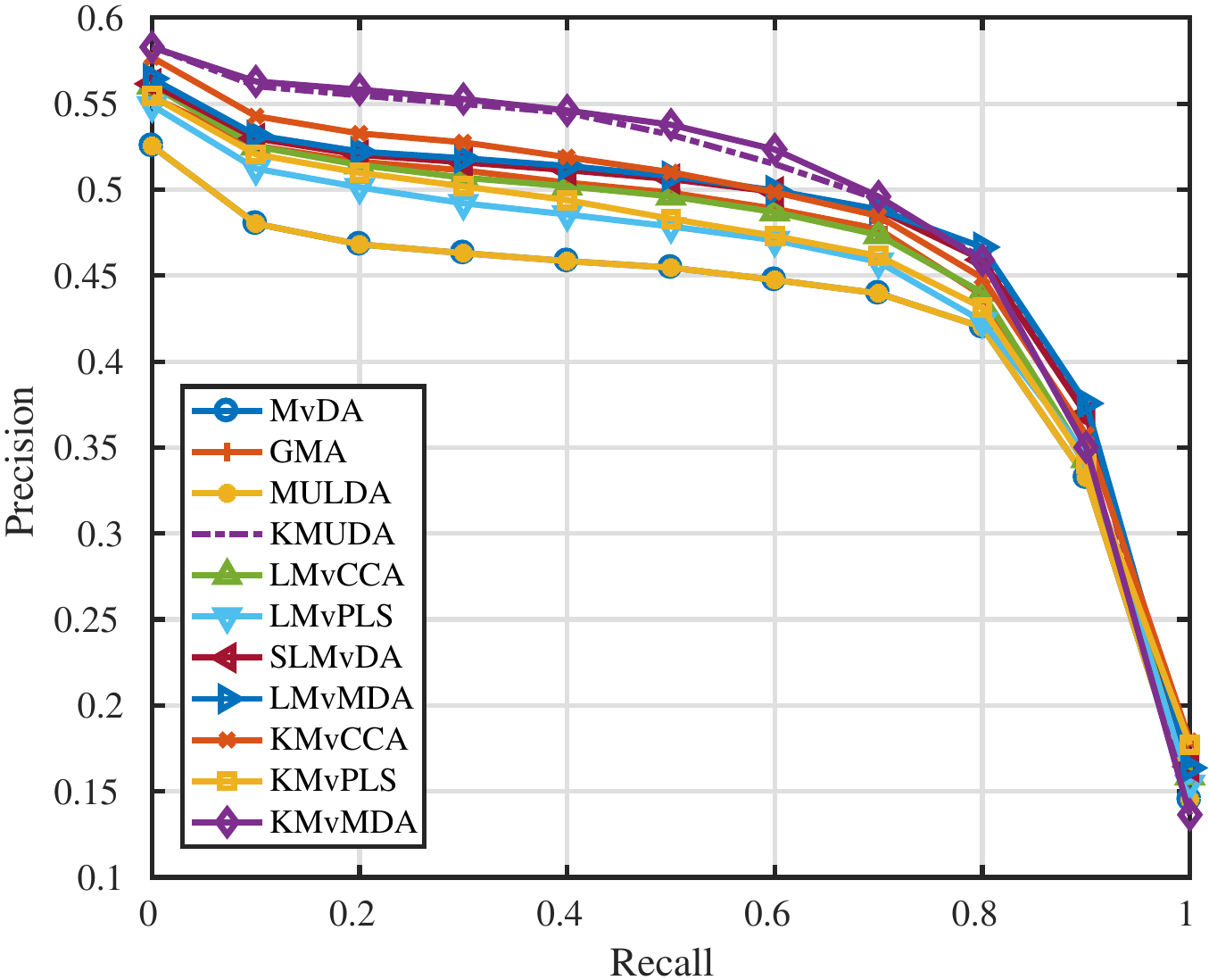}
\caption{PR curves for image queries with 3 views}
\end{subfigure}
\begin{subfigure}[b]{.3\linewidth}
\includegraphics[width=1\textwidth]{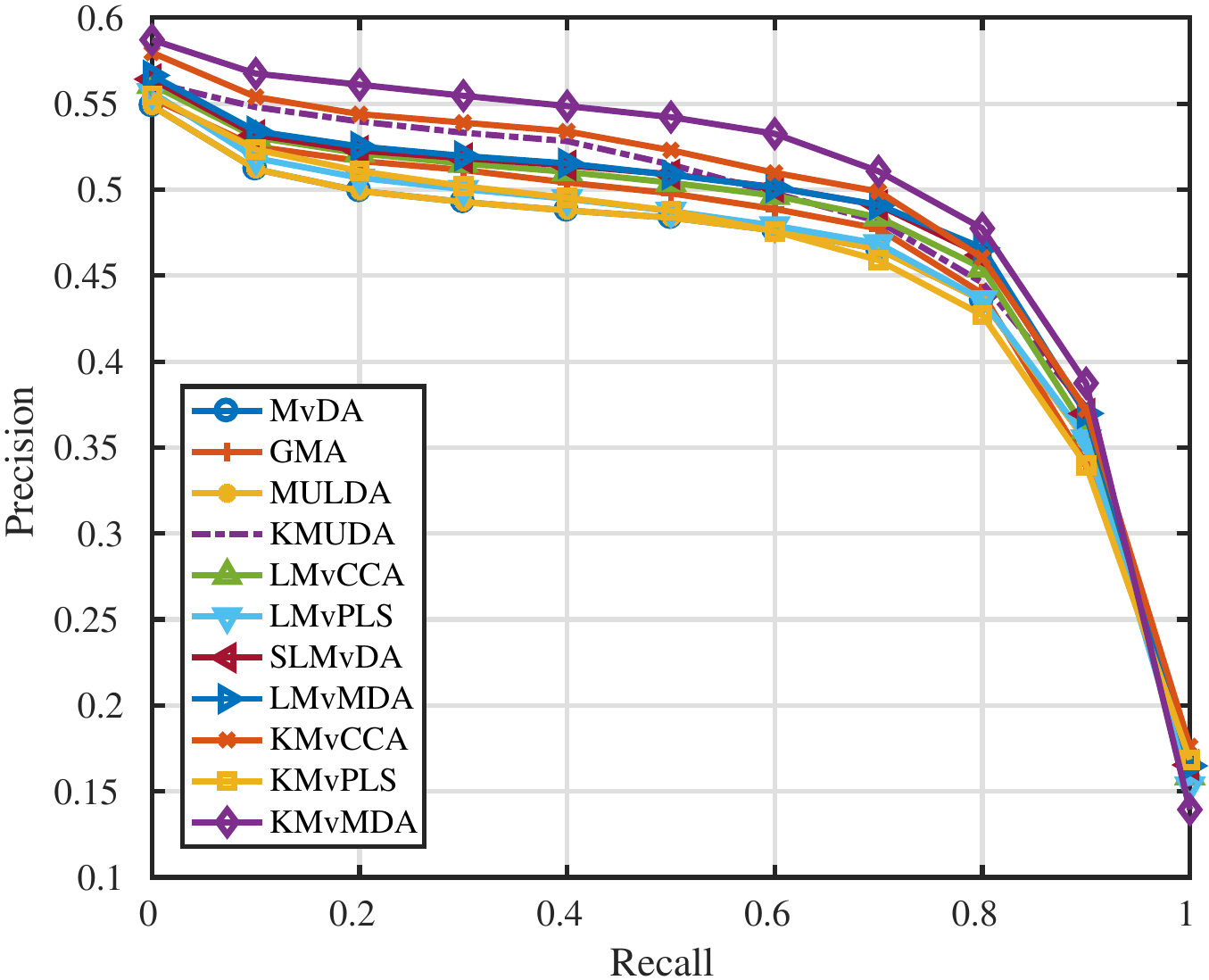}
\caption{PR curves for image queries with 4 views}
\end{subfigure}\\[6mm]
\begin{subfigure}[b]{.3\linewidth}
\includegraphics[width=1\textwidth]{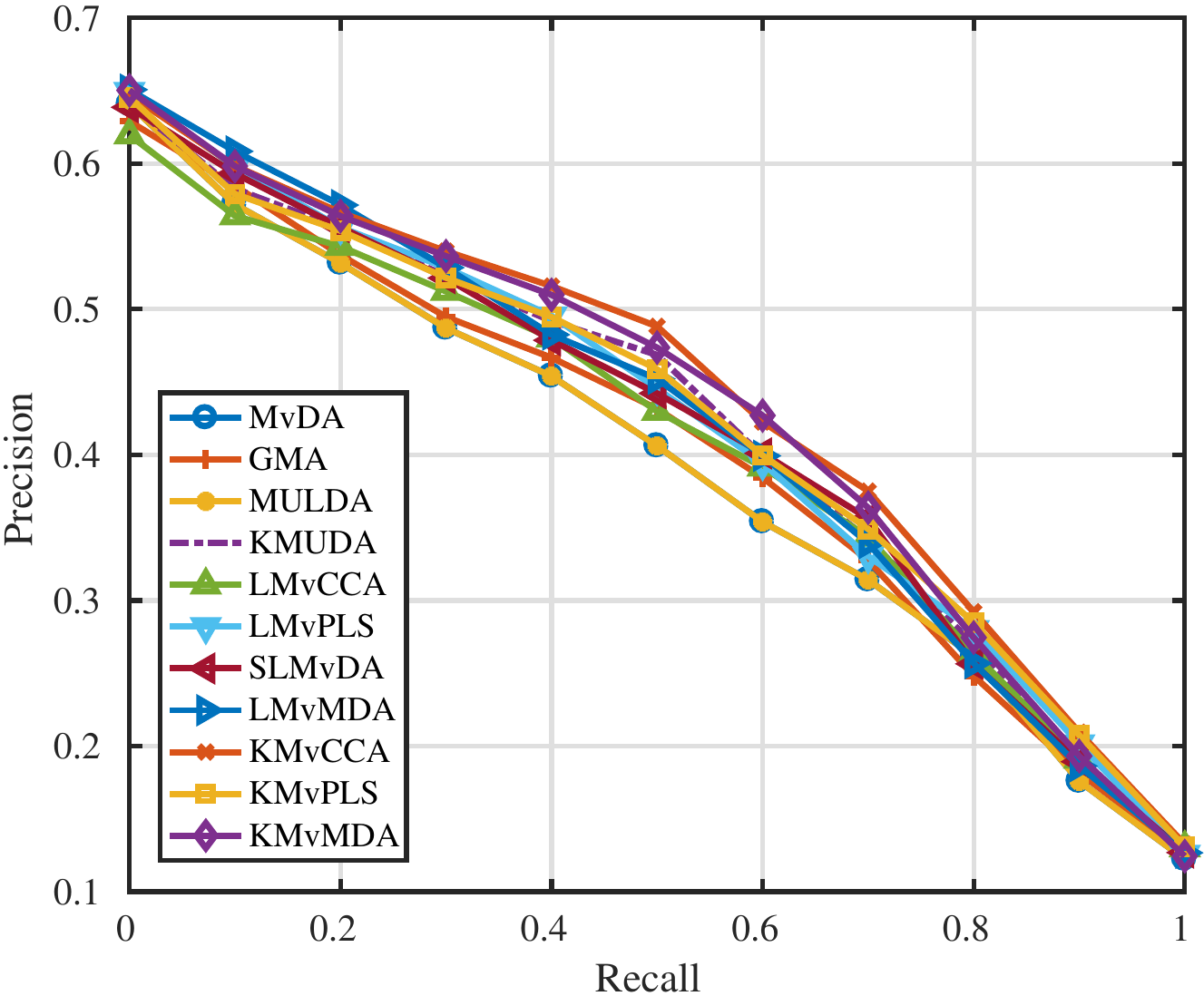}
\caption{PR curves for text queries with 2 views}
\end{subfigure}
\begin{subfigure}[b]{.3\linewidth}
\includegraphics[width=1\textwidth]{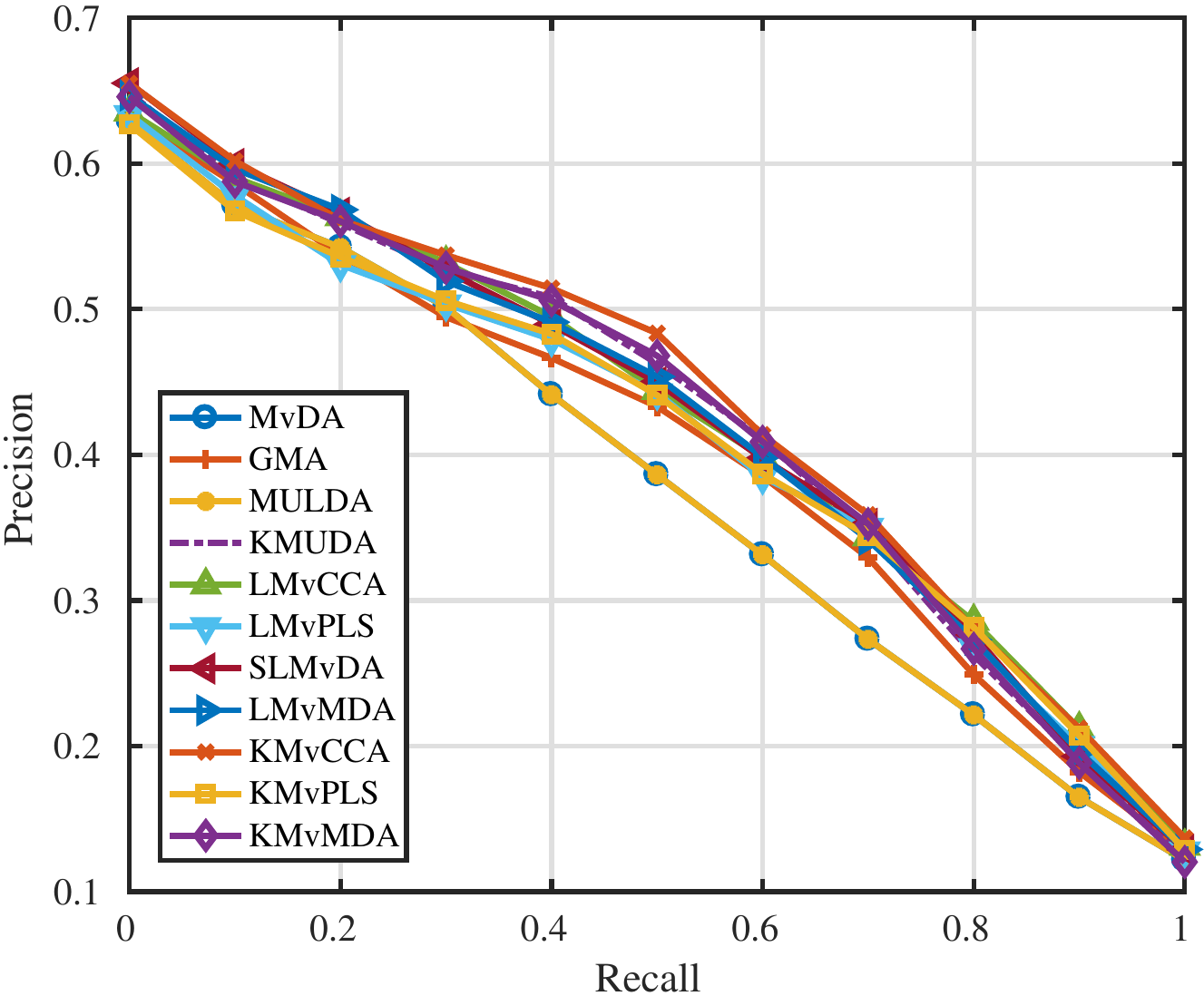}
\caption{PR curves for text queries with 3 views}
\end{subfigure}
\begin{subfigure}[b]{.3\linewidth}
\includegraphics[width=1\textwidth]{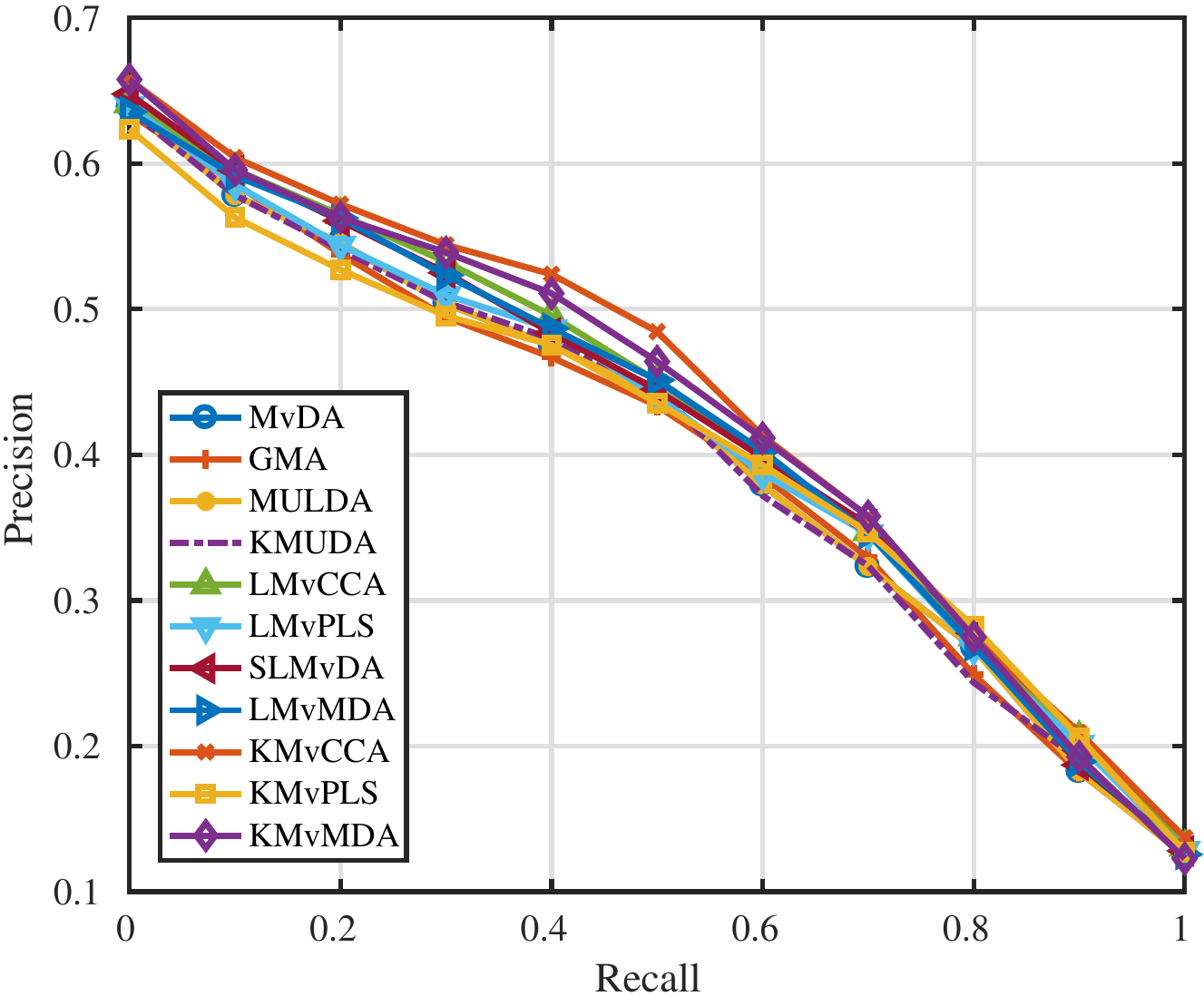}
\caption{PR curves for text queries with 4 views}
\end{subfigure}\\[1mm]
\caption{PR curves across different number of views on the Wikipedia dataset for the Image-to-Text retrieval and the Text-to-Image retrieval. }\label{fig:wiki_pr}
\end{figure*}

We present more detailed results in the form of PR curves in Fig. \ref{fig:wiki_pr}. For image queries, KMvMDA consistently outperforms the other methods across all views, which can be explained by its utilization of class labels and kernel-based representations. For text queries, the supervised and non-linear methods also outperform their linear counterparts. KMvCCA and KMvMDA are the leading methods in this category, which shows the strength of cross-modal retrieval by making use of view difference.
\subsubsection{Cross-modal retrieval results on the COCO Dataset}
 \begin{table*}
   \begin{center}
   \caption{MAP Scores (\%) on the COCO dataset}\label{tab:coco}
  \resizebox{0.8\linewidth}{!}{
\begin{tabular}{|l|c|c| c|c|c| c|c|c|c| }\hline
 & \multicolumn{3}{ |c| }{2 views} & \multicolumn{3}{ |c| }{3 views} & \multicolumn{3}{ |c| }{4 views}\\ \hline\hline
& img. query & txt. query & avg. & img. query & txt. query & avg.& img. query & txt. query & avg.\\ \hline
Proposed LMvCCA & 87.18  & 86.92  & 87.05  & 87.20   & 87.01  & 87.11 & 87.31  & 87.22  & 87.27\\
Proposed LMvPLS & 84.76  & 85.05  & 84.91 & 84.83  & 85.07  & 84.95  & 84.82  & 85.05  & 84.94\\\hline
Proposed KapMvCCA & 88.42  & 87.58  & 88.00    & 88.35  & 87.52  & 87.94 & 88.45  & 87.60   & 88.03\\
Proposed KapMvPLS & 87.16  & 86.58  & 86.87  & 87.14  & 86.56  & 86.85  & 87.14  & 86.56  & 86.85\\\hline
Proposed DMvCCA &88.14&88.10&88.12&88.20&88.26&\textbf{88.23}&88.49&88.40&\textbf{88.45} \\
Proposed DMvPLS &88.01&88.03&88.02&88.06&88.03&88.05&88.45&88.34& 88.40\\\hline
DCCA2 \cite{Andrew2013} &88.30&88.27&\textbf{88.29}&-&-&-&-&-&- \\ \hline
\end{tabular}}
\end{center}
\end{table*}
\begin{figure*}
[h!]
\begin{subfigure}[b]{.3\linewidth}
\includegraphics[width=1\textwidth]{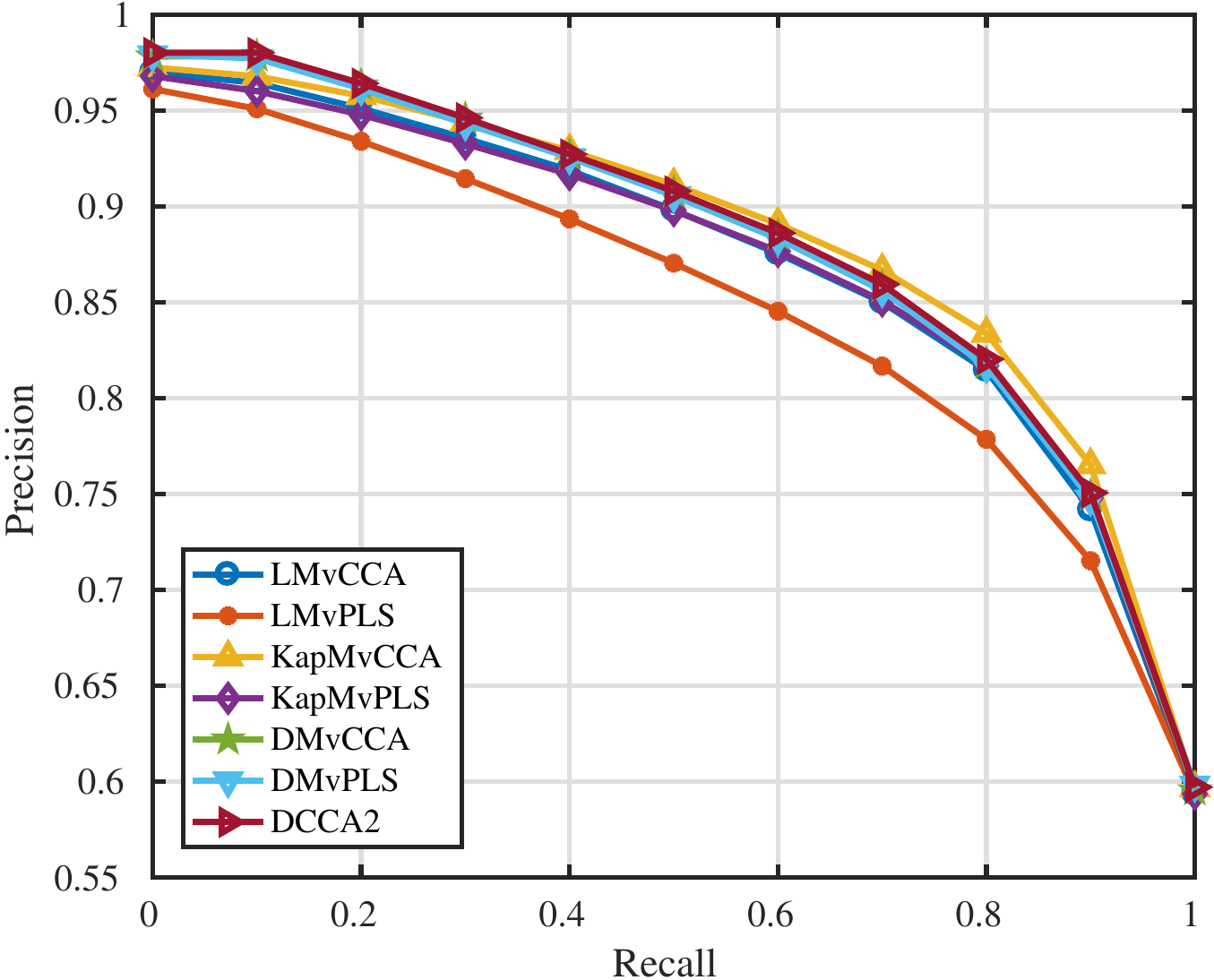}
\caption{PR curves for image queries with 2 views}
\end{subfigure}
\begin{subfigure}[b]{.3\linewidth}
\includegraphics[width=1\textwidth]{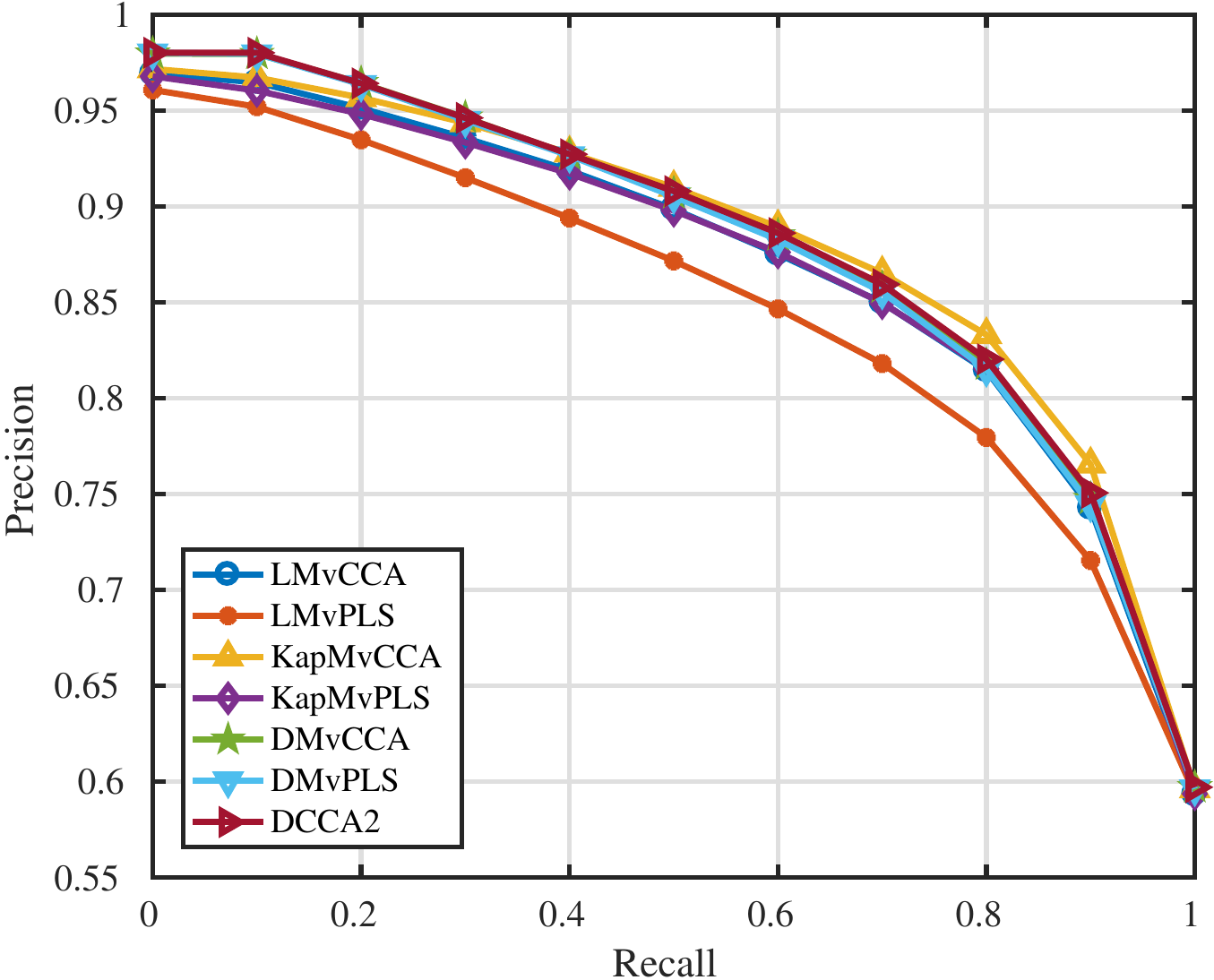}
\caption{PR curves for image queries with 3 views}
\end{subfigure}
\begin{subfigure}[b]{.3\linewidth}
\includegraphics[width=1\textwidth]{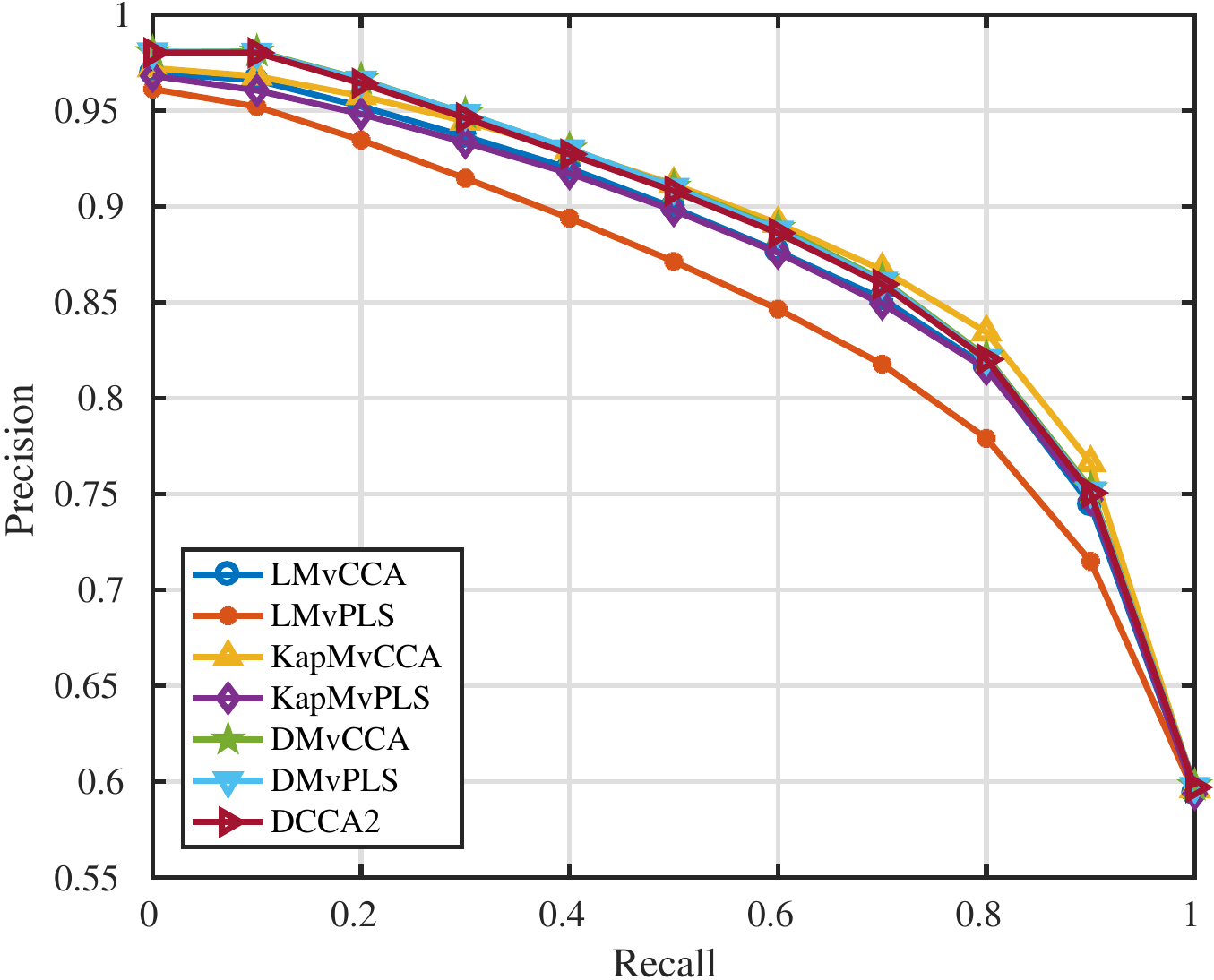}
\caption{PR curves for image queries with 4 views}
\end{subfigure}\\[6mm]
\begin{subfigure}[b]{.3\linewidth}
\includegraphics[width=1\textwidth]{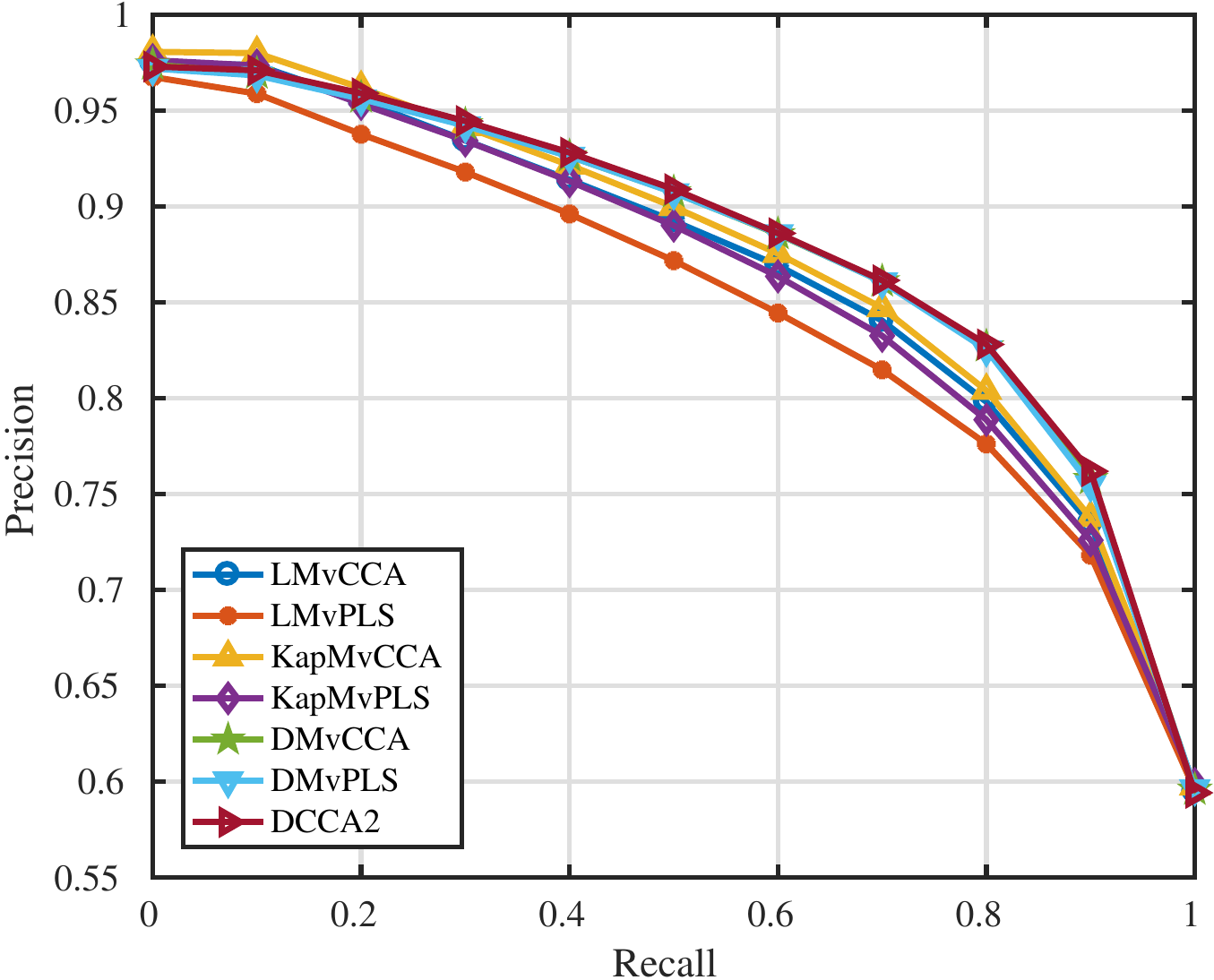}
\caption{PR curves for text queries with 2 views}
\end{subfigure}
\begin{subfigure}[b]{.3\linewidth}
\includegraphics[width=1\textwidth]{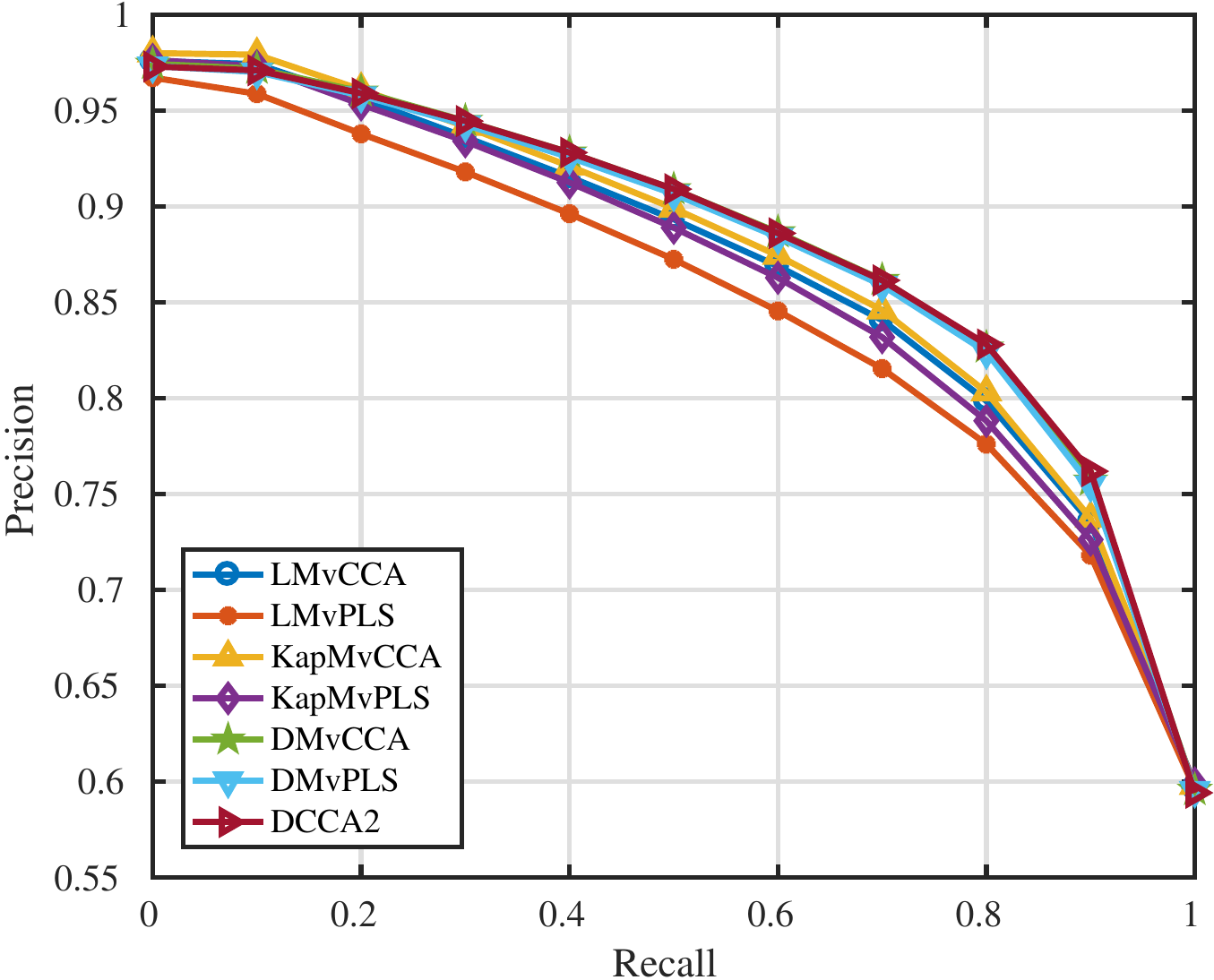}
\caption{PR curves for text queries with 3 views}
\end{subfigure}
\begin{subfigure}[b]{.3\linewidth}
\includegraphics[width=1\textwidth]{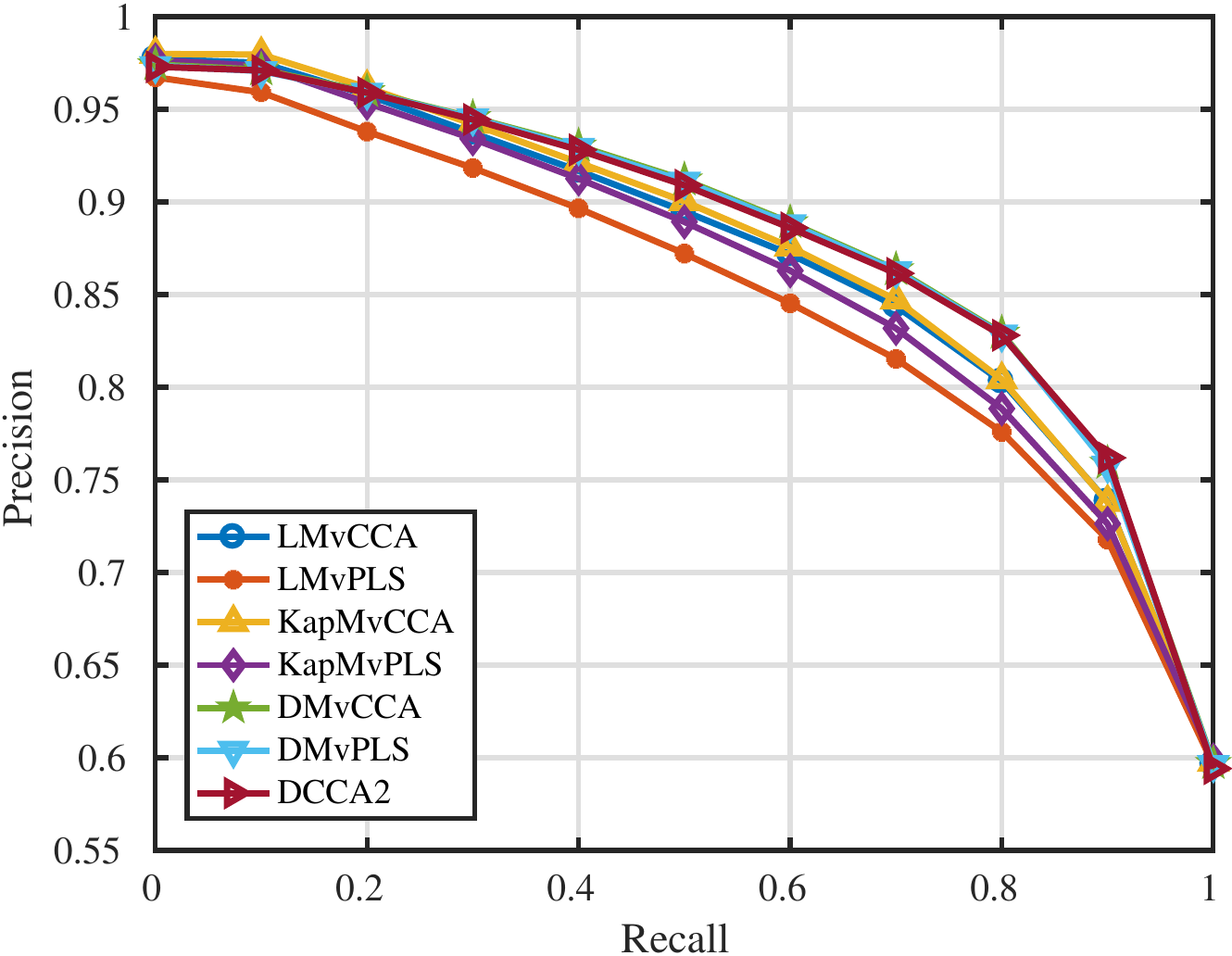}
\caption{PR curves for text queries with 4 views}
\end{subfigure} \\
\caption{PR curves across different number of views on the COCO dataset for the Image-to-Text retrieval and the Text-to-Image retrieval. Note the curve by DCCA2 \cite{Andrew2013} is presented across all numbers of views.}\label{fig:coco_pr}
\end{figure*}
The COCO dataset is much larger than the Wikipedia dataset, and we pay more attention to the non-linear methods especially the ones using neural networks. Many images have more than one class labels, and therefore we focus on the unsupervised learning algorithms. Similar to the experiments above, the MAP scores in Table \ref{tab:coco} show that a gain of retrieval accuracy can be obtained by embedding additional modalities into the latent space. DCCA2 \cite{Andrew2013} achieves a superior performance with 2 views thanks to its non-linear projection which makes the latent feature more discriminant for retrieval. However, its formulation limits the algorithm to 2 views, and DMvCCA and DMvPLS based on the proposed framework can improve the state-of-the-art method by increasing the number of modalities. From the PR curves in Fig. \ref{fig:coco_pr}, we compare the methods using the proposed objective function with DCCA2 which contains two views. For image queries, KapMvCCA obtains the best retrieval result with 2 views, but it is further improved by the methods using neural networks benefitted by attributes and GoogleNet features. For text queries, it also suggests more modalities and neural network-based representations contribute to the retrieval performance. The cross-modal retrieval by the 4-view DMvCCA achieves the overall highest precision score on this dataset.
\begin{figure*}
  \begin{center}
  \resizebox{0.9\linewidth}{!}{
    \begin{tabular}{|c c c|}\hline
        Image Query & \multicolumn{2}{|c|}{Text Query} \\\hline
\includegraphics[width=.1\linewidth,height=.1\linewidth]{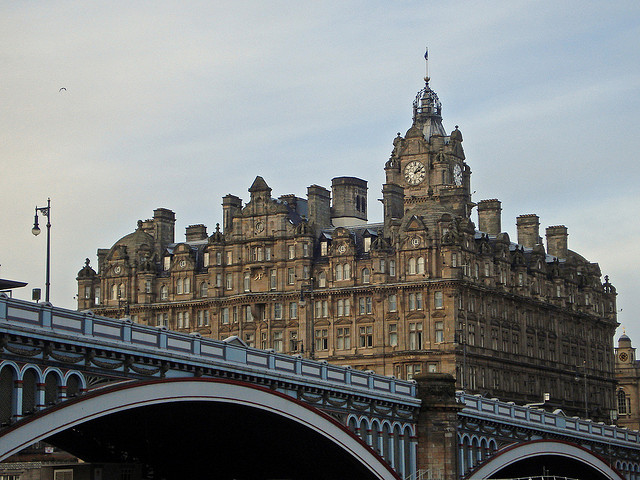} & \multicolumn{2}{|b{.65\linewidth}| }{
1. A very big building with many windows and a clock on it. \newline
2. A very old tall building with a large clock tower sticking out of it.\newline
3. The clock tower stands high above the city. \newline
4. A clock that is on the side of a large building. \newline
5. The bridge is in front of a huge building with a clock tower in the middle of it. }\\\hline
Precision: $53.33\%$ & Precision: $86.67\%$  & Precision; $100\%$ \\
\includegraphics[width=.32\linewidth]{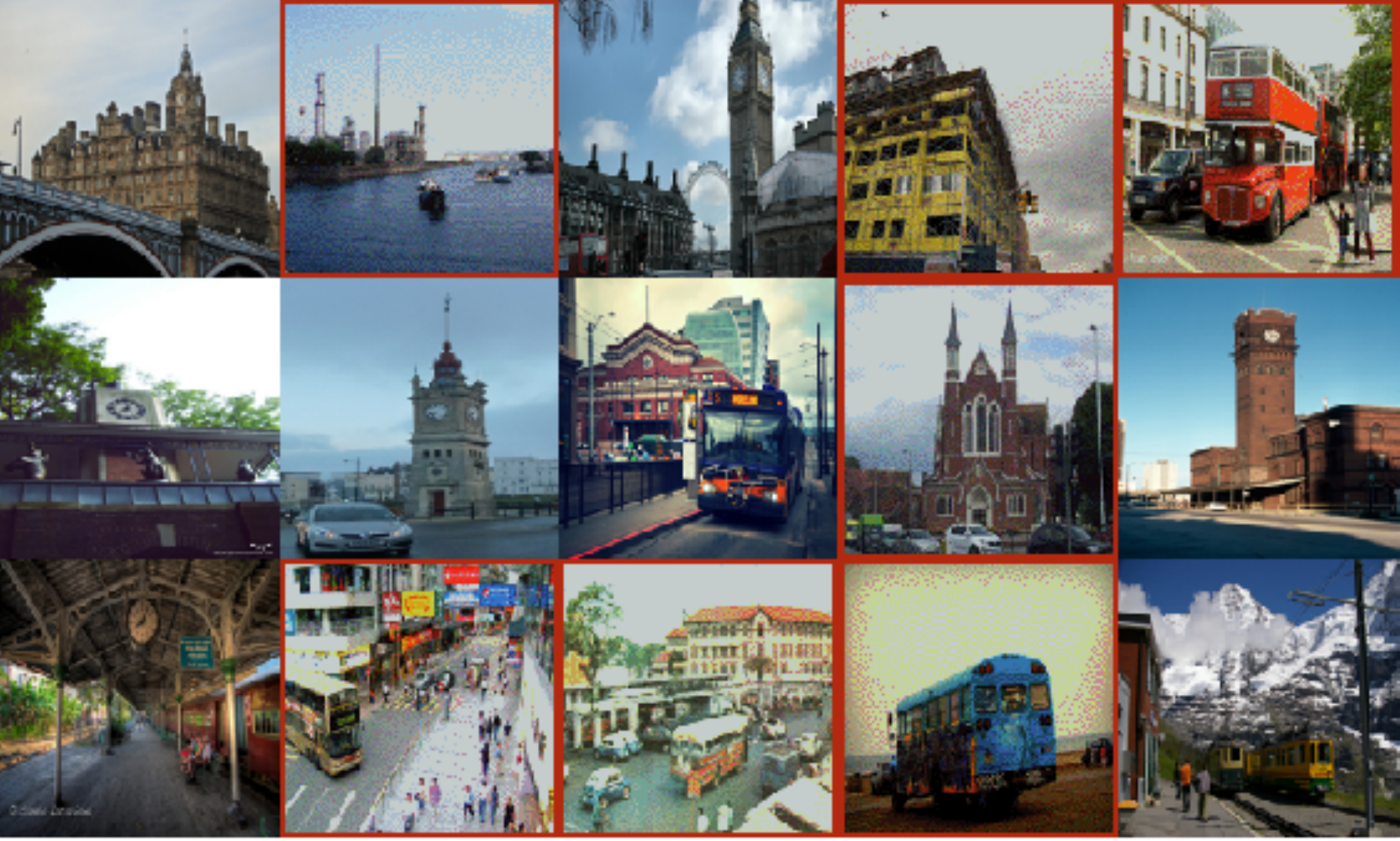}
&\includegraphics[width=.32\linewidth]{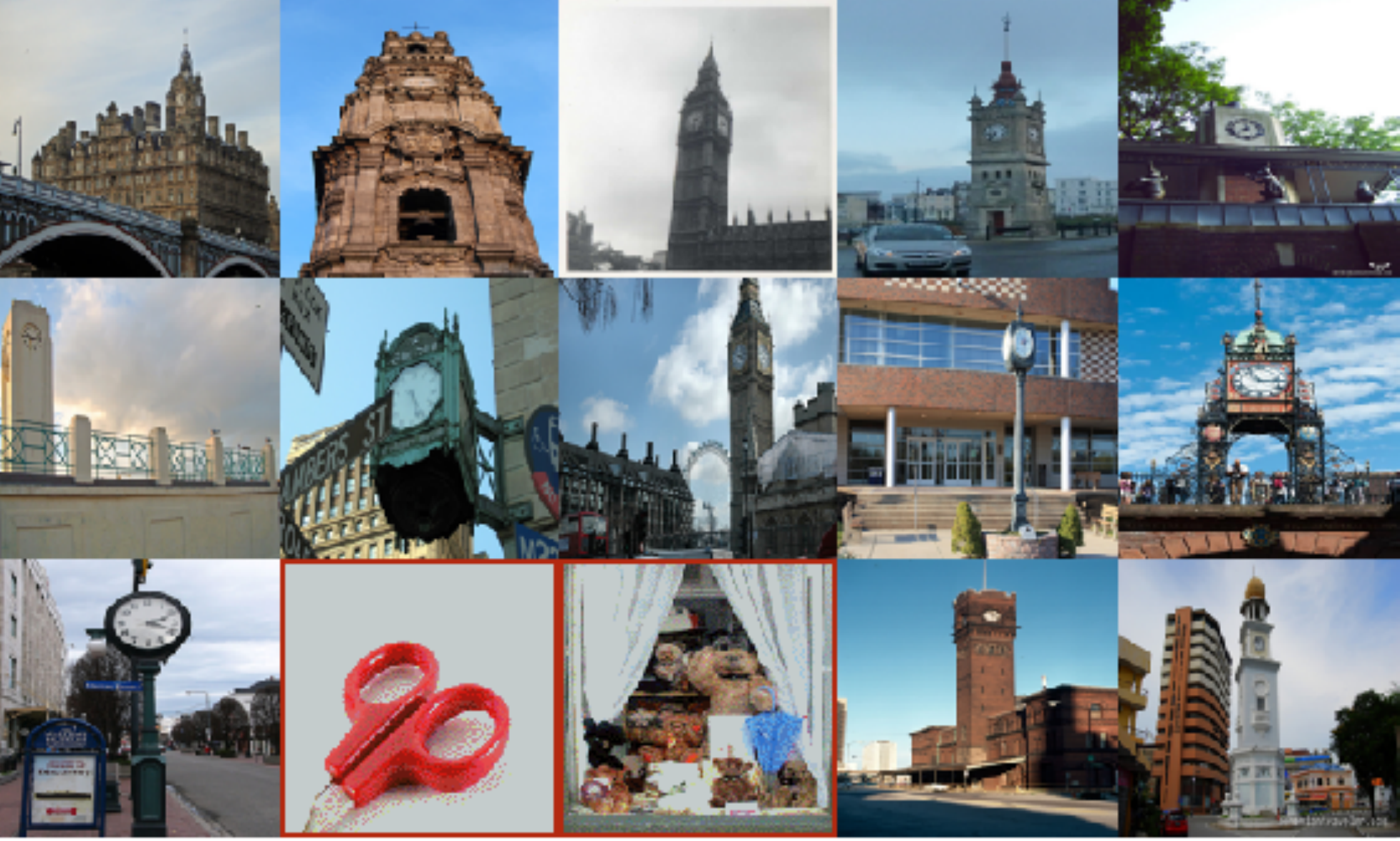}
&\includegraphics[width=.32\linewidth]{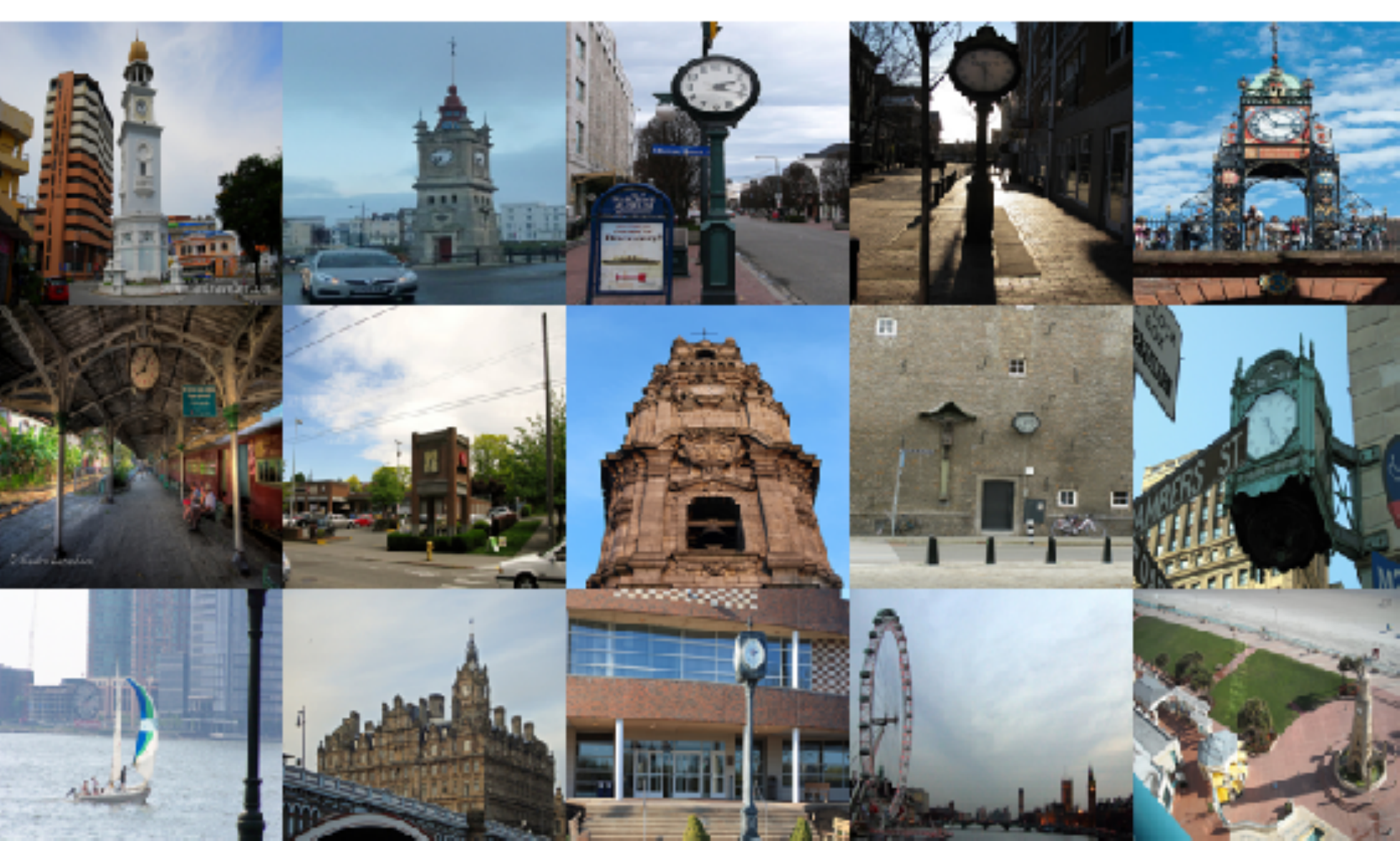} \\
(a) Query by original image feature & (b) Query by projected image feature & (c) Query by text \\ \hline \hline
Image Query & \multicolumn{2}{|c|}{Text Query} \\\hline
\includegraphics[width=.1\linewidth,height=.1\linewidth]{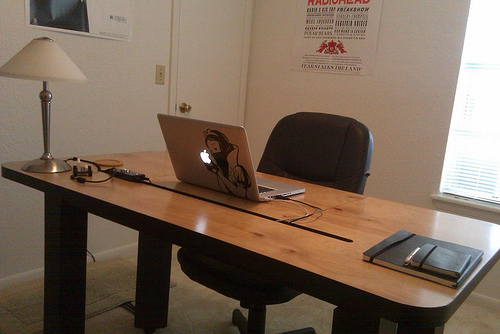} & \multicolumn{2}{|b{.65\linewidth}| }{
1. An open laptop sits on a desk in front of a window.  \newline
2. An Apple laptop sitting on a wooden desk. \newline
3. An Apple laptop sitting on a wooden desk in an office. \newline
4. An Apple laptop on a desk in an office. \newline
5. A desk with a laptop sitting on top of it.
} \\ \hline
Precision: $60.00\%$ & Precision: $86.67\%$ & Precision: $66.67\%$ \\
\includegraphics[width=.32\linewidth]{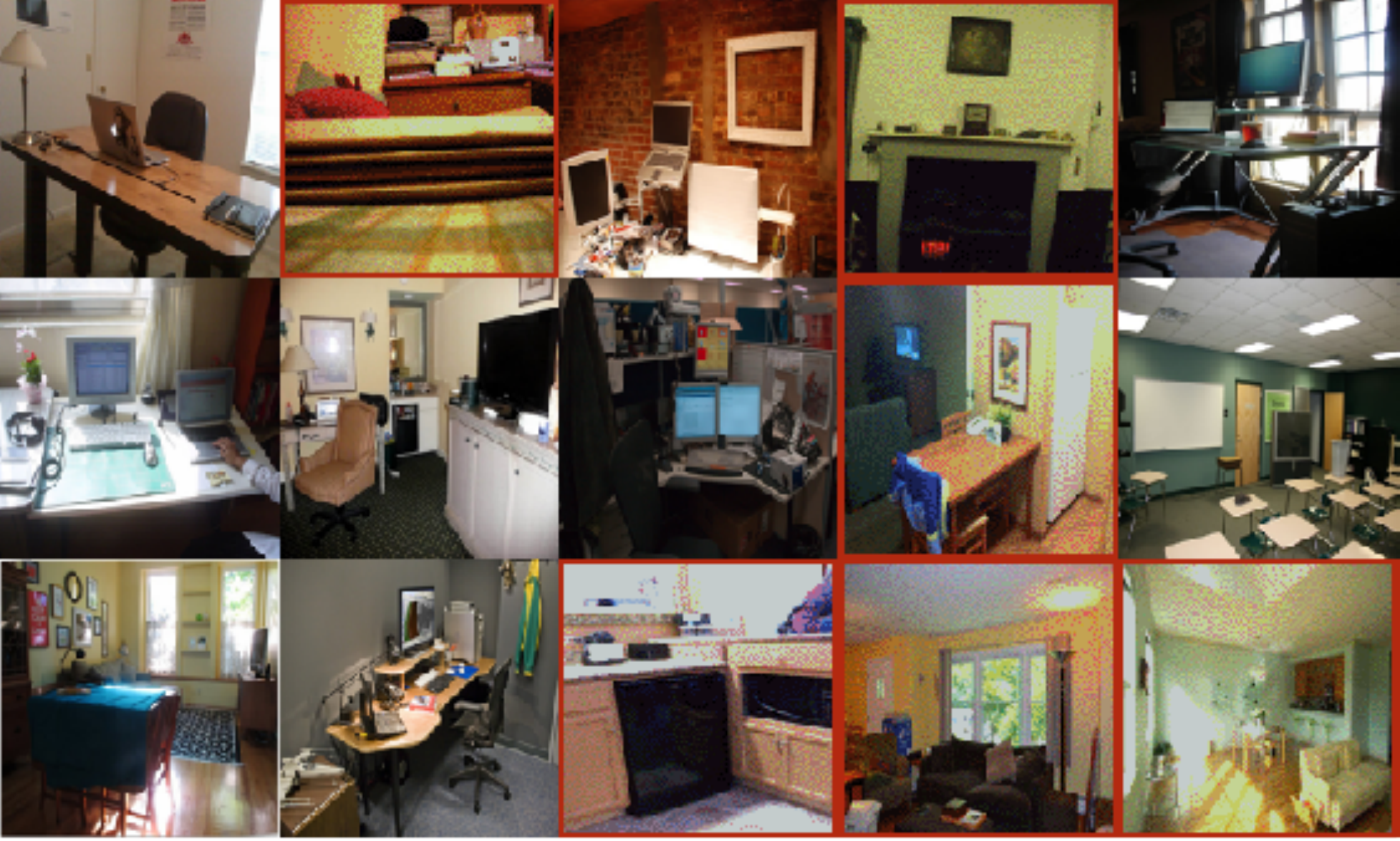}
&\includegraphics[width=.32\linewidth]{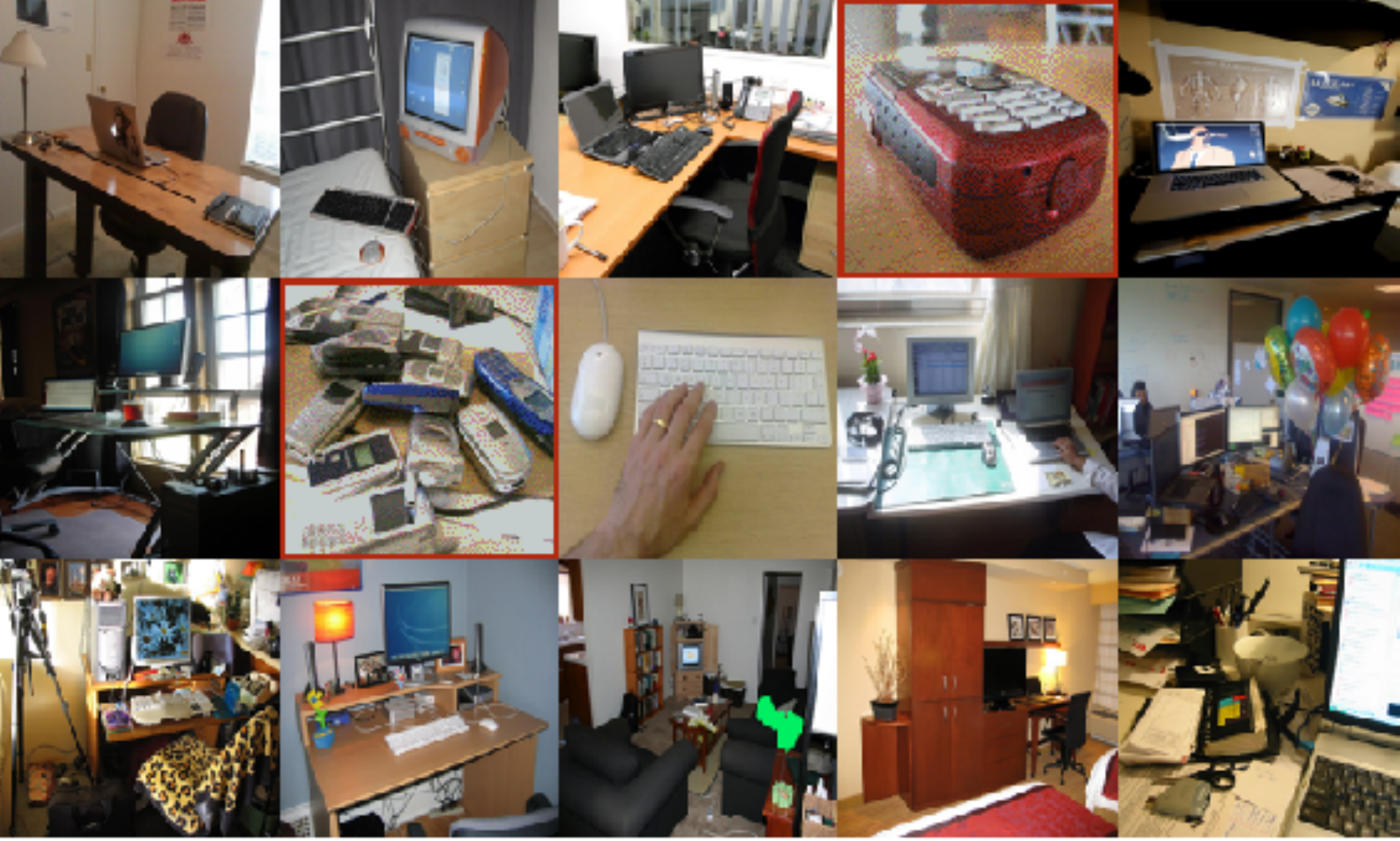}
&\includegraphics[width=.32\linewidth]{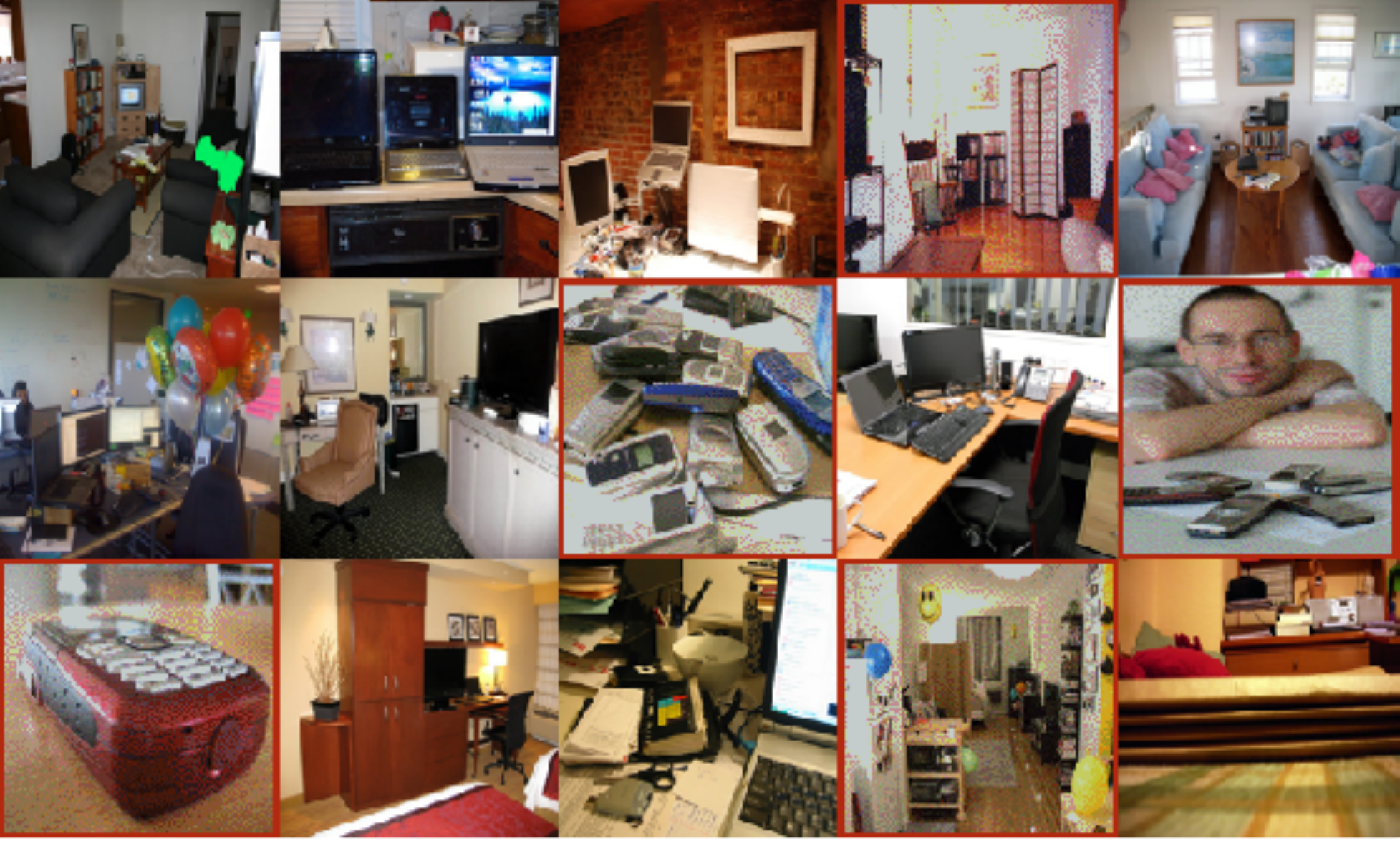} \\
(a) Query by original image feature & (b) Query by projected image feature & (c) Query by text \\ \hline
\end{tabular}}
  \end{center}
\caption{Sample retrieval results on the COCO dataset. The first row of each table presents the query image and text, and the second row shows the retrieved images by different query types. False positive results are bounded in red.}\label{fig:cbir}
\end{figure*}
\subsubsection{Content-based Image Retrieval (CBIR) Performance on the COCO dataset}\label{sec:cbir}
We also show the effectiveness of multi-view embedding method on the conventional CBIR task in Fig. \ref{fig:cbir}. We randomly pick two image-to-text pairs as queries, to perform image-to-image retrieval using both the \emph{VGG-16} visual feature and the projected visual feature by the 4-view DCCA. We also perform text-to-image retrieval by querying the corresponding captions of the query image used in CBIR in the last column. We observe the CBIR performance can be further improved by incorporating the semantic information. In Table \ref{tab:cbir}, we present the quantitative results of CBIR by the projected visual features. ``RAW'' in the Table shows the retrieval results by visual features directly, while the rest are the multi-view embedding results. It is shown that more modalities and non-linear projections yield a discriminant latent visual feature, which improves the retrieval performance.
\begin{table}
\begin{center}
\caption{MAP(\%) scores of CBIR on the COCO dataset}\label{tab:cbir}
\resizebox{0.8\linewidth}{!}{
\begin{tabular}{|l| c| c| c| }
\hline
Method & 2 views & 3 views & 4 views\\ \hline\hline
Raw & \multicolumn{3}{c|}{83.77} \\ \hline
Proposed LMvCCA &85.64&85.76&85.93\\
Proposed LMvPLS &84.30&84.30&84.32\\
Proposed KapMvCCA &85.43&85.47&85.49 \\
Proposed KapMvPLS &84.56 &84.57 &  84.58\\
Proposed DMvCCA &89.33&\textbf{89.62}&\textbf{89.84}\\
Proposed DMvPLS &89.50&89.34&89.79\\
DCCA2 \cite{Andrew2013}&\textbf{89.71}&-&-\\\hline
\end{tabular}}
\end{center}
\end{table}
\subsection{Parameter sensitivity analysis of dimension $d$ in linear and kernel cases} \label{sec:k}
\begin{figure}
\centering
\begin{subfigure}[b]{.49\linewidth}
\includegraphics[width=\textwidth]{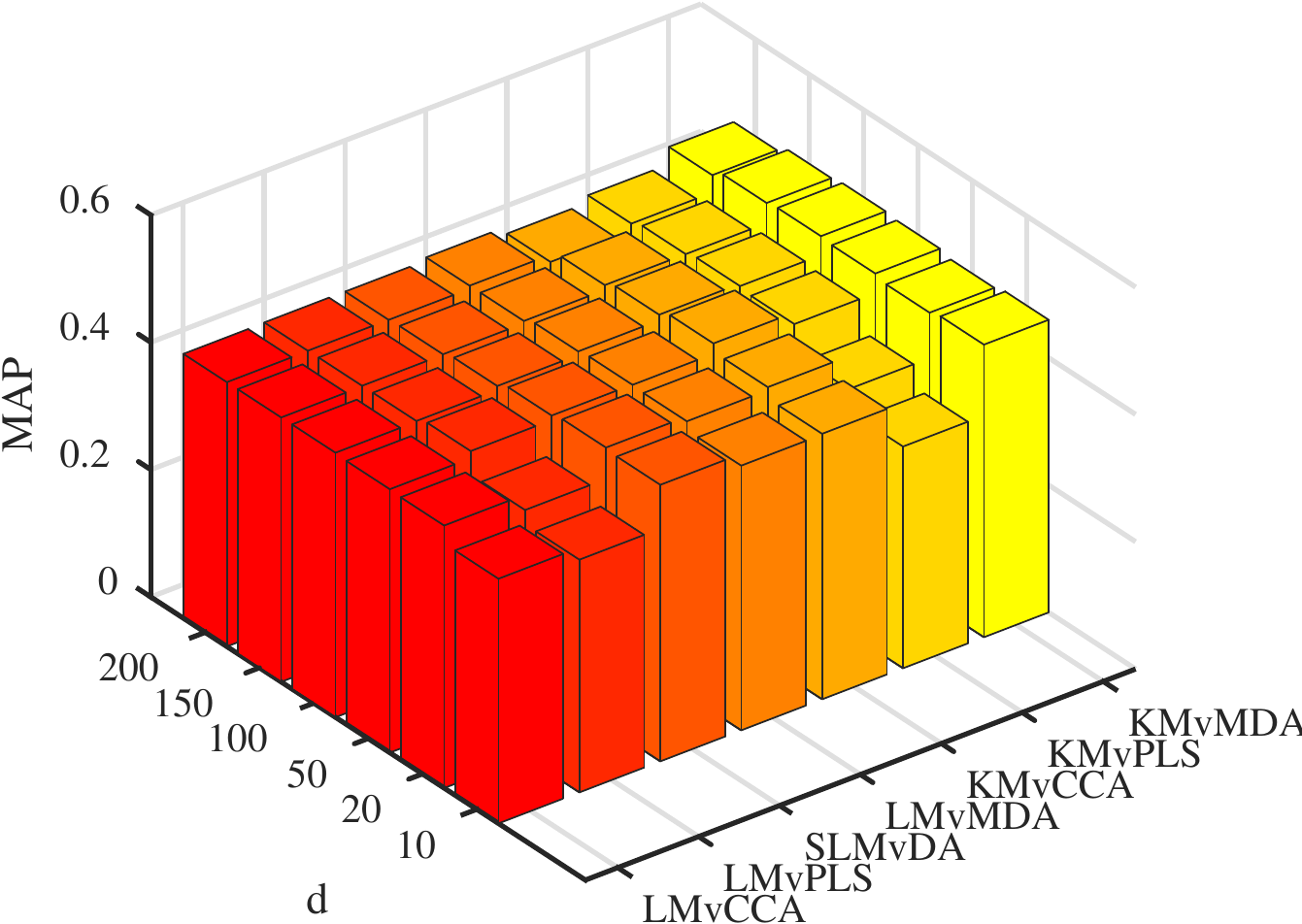}
\caption{Image queries with 2 views}
\end{subfigure}
\begin{subfigure}[b]{.49\linewidth}
\includegraphics[width=\textwidth]{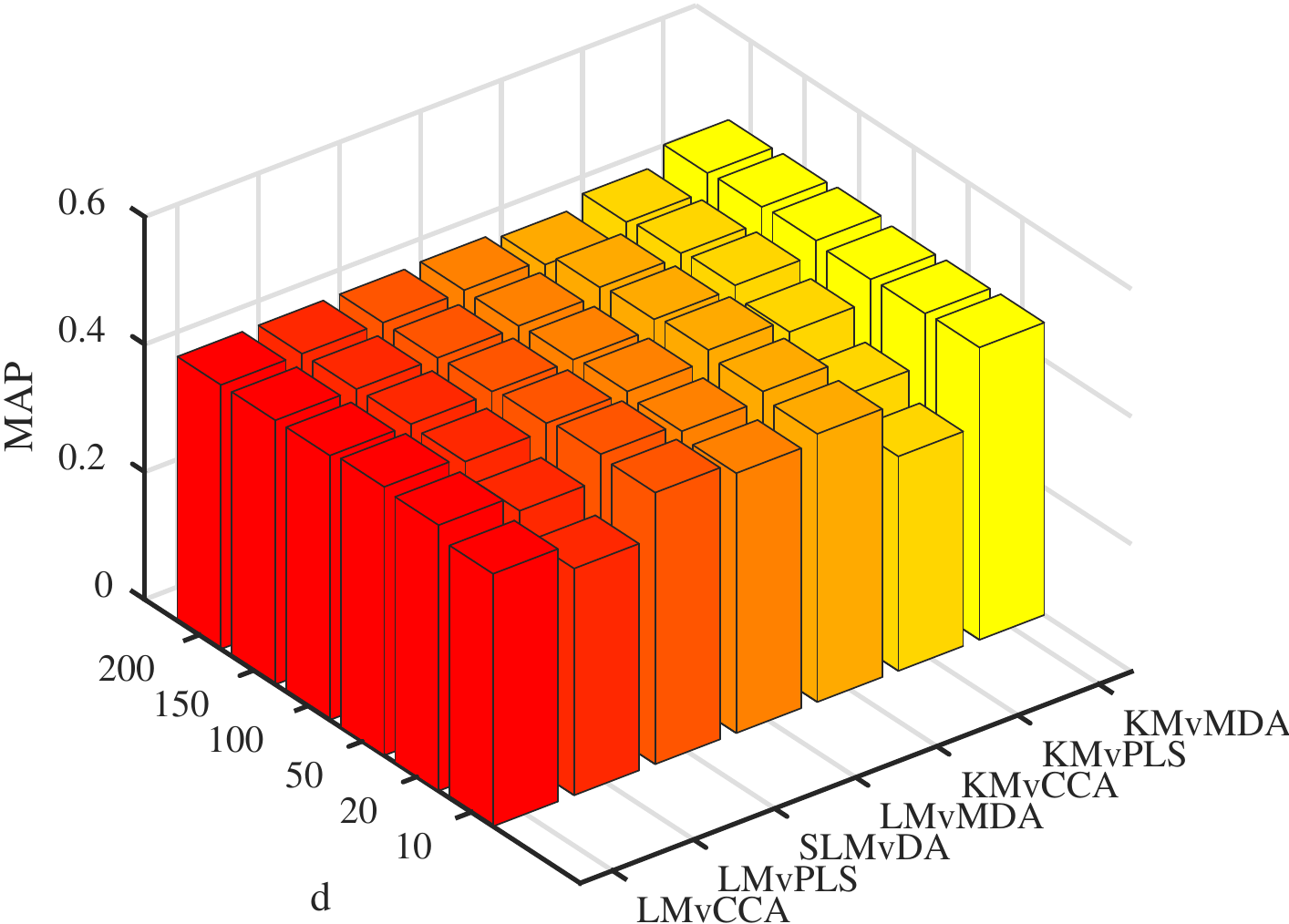}
\caption{Image queries with 3 views}
\end{subfigure}\\[2.5mm]
\begin{subfigure}[b]{.49\linewidth}
\includegraphics[width=\textwidth]{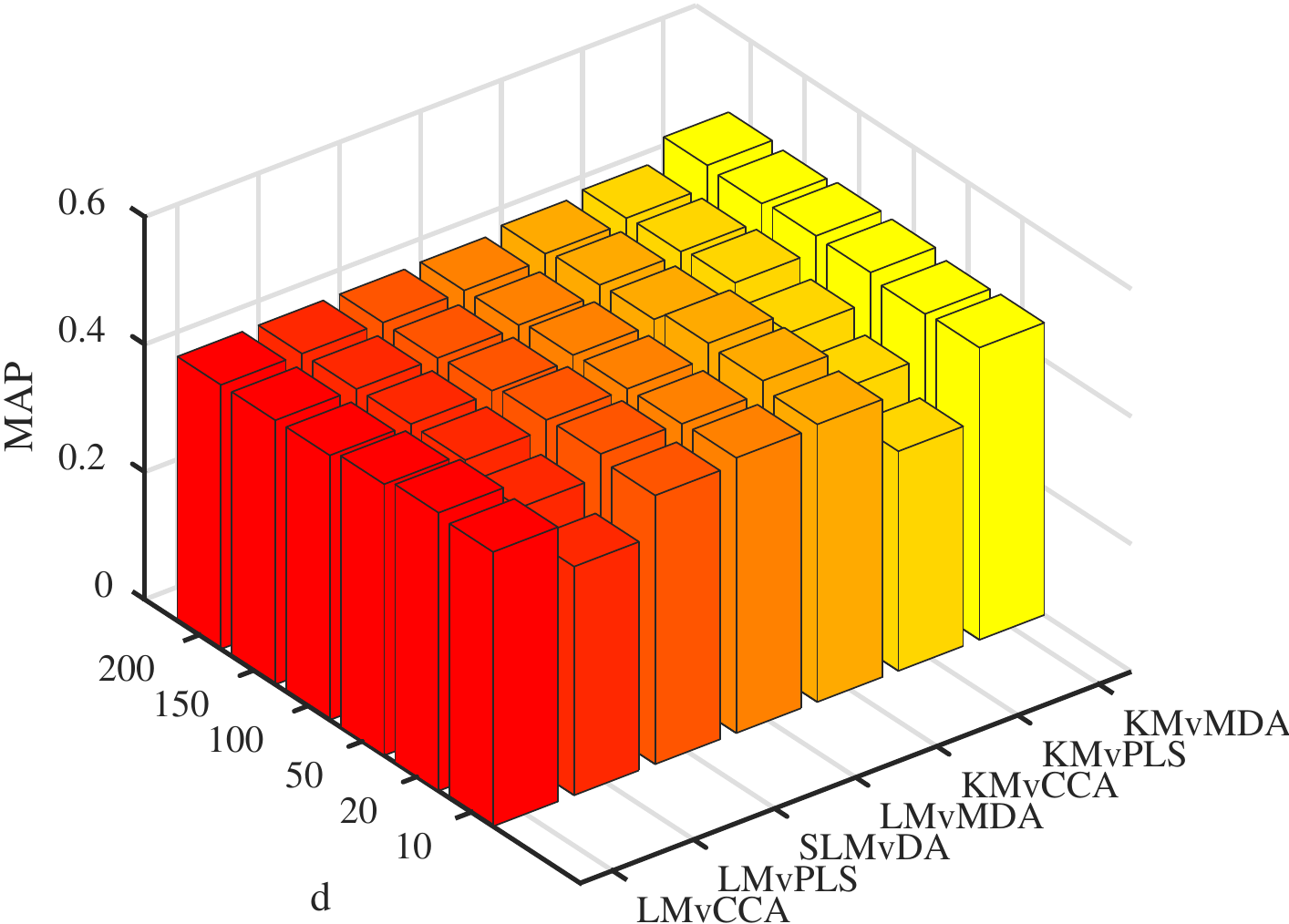}
\caption{Image queries with 4 views}
\end{subfigure}
\caption{Performance variation for image queries on texts of Wikipedia
dataset with respect to the different dimension $d$. }\label{fig:d1}
\end{figure}

\begin{figure}
\centering
\begin{subfigure}[b]{.49\linewidth}
\includegraphics[width=\textwidth]{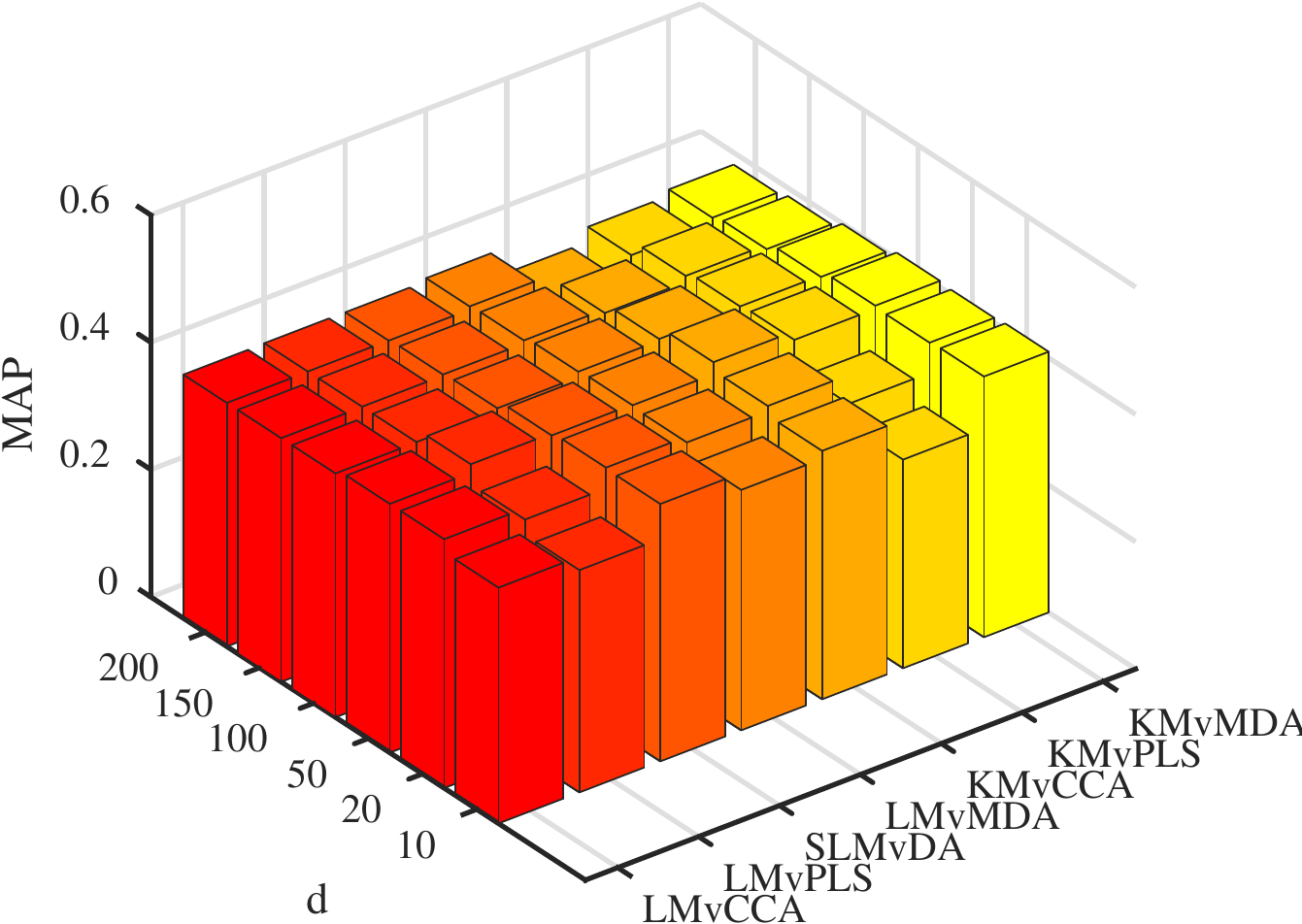}
\caption{Text queries with 2 views}
\end{subfigure}
\begin{subfigure}[b]{.49\linewidth}
\includegraphics[width=\textwidth]{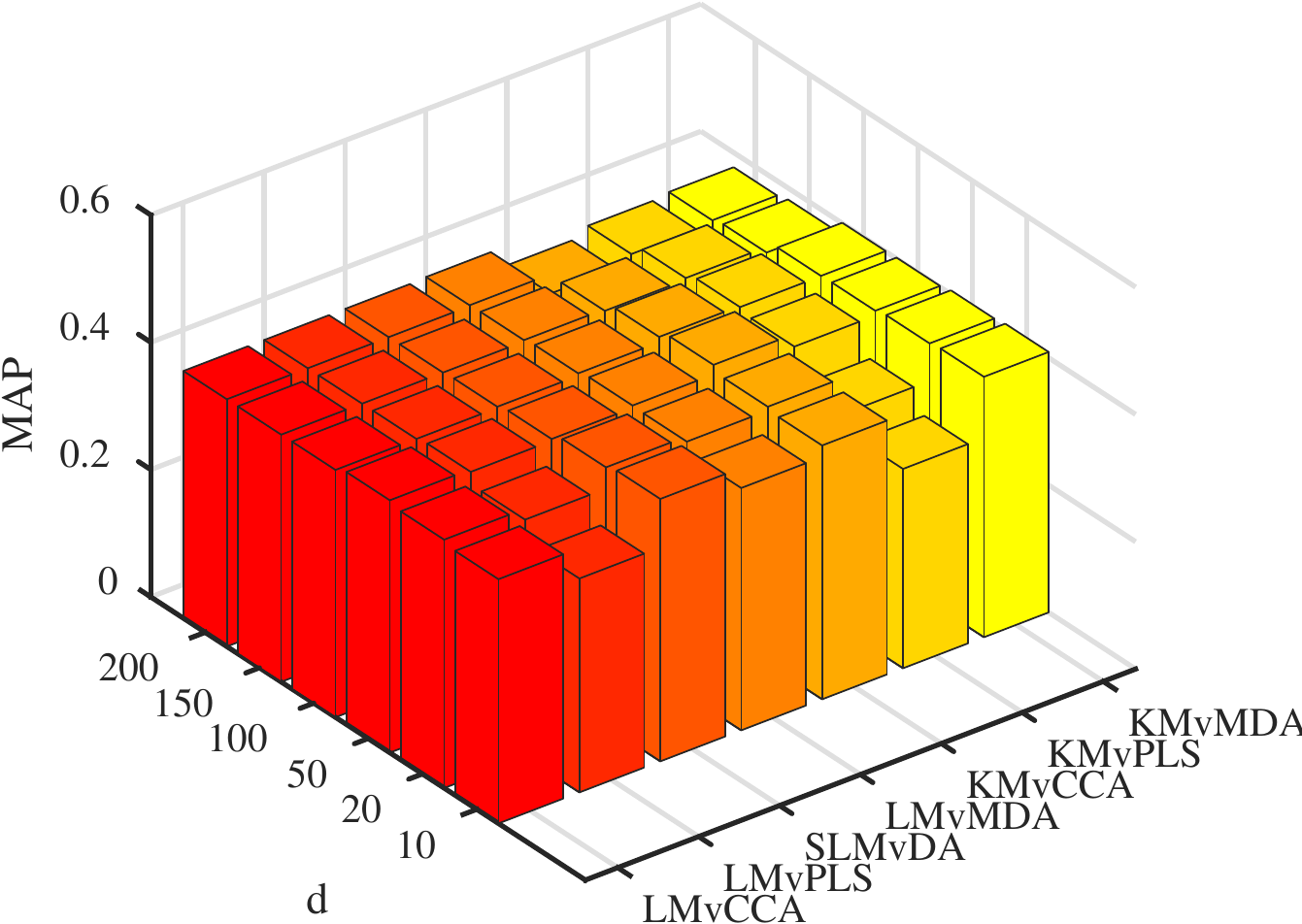}
\caption{Text queries with 3 views}
\end{subfigure}\\[2.5mm]
\begin{subfigure}[b]{.49\linewidth}
\includegraphics[width=\textwidth]{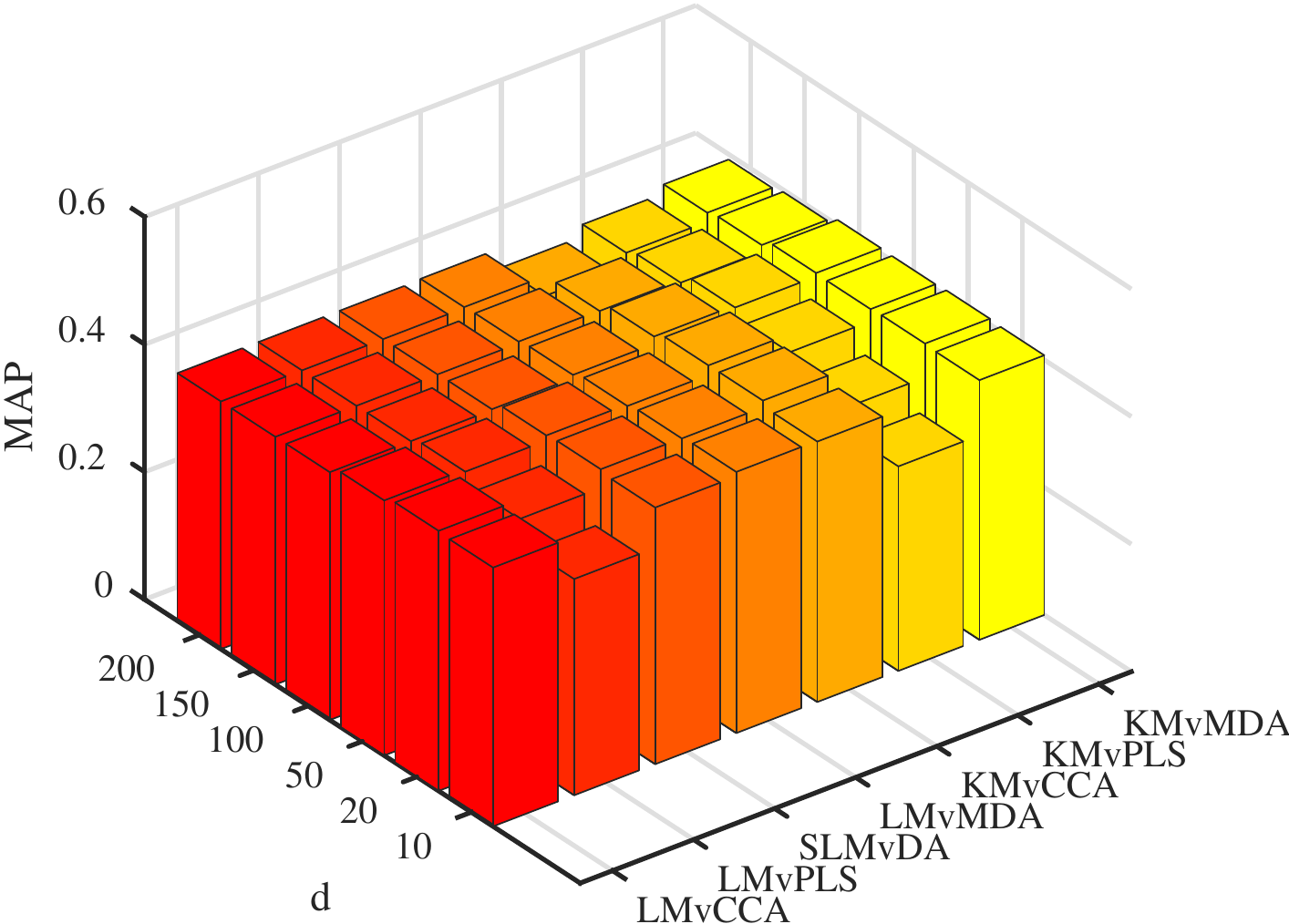}
\caption{Text queries with 4 views}
\end{subfigure}
\caption{Performance variation for text queries on images of Wikipedia
dataset with respect to the different dimension $d$. }\label{fig:d2}
\end{figure}

The number of dimension of the feature vectors in the latent space is determined by the top $d$ eigenvectors in the projection matrix, and it is pre-defined in the former experiments. Therefore in this section, we investigate the effect by the variation of $d$ shown in Fig. \ref{fig:d1} and \ref{fig:d2}, ranging from $\{10, 20, 50, 100, 150, 200\}$. The performance on the Wikipedia dataset is reported with both text queries on images and image queries on texts. The results on different number of views are also recorded. In general, we obtain a better retrieval performance when $d$ is between $50$ and $150$. It can be explained by the fact that the most informative eigenvectors are included within the range. Therefore, $d=50$ was chosen for the multi-view linear embeddings in the experiments. Except LMvPLS and KMvPLS, we find the majority of the methods are robust to the dimensionality changes in the subspace.
\section{Conclusion}\label{sec:con}
In this paper, we proposed a generalized multi-view embedding method using the graph embedding framework. We showed multi-view CCA, PLS and LDA can be characterized by their specific intrinsic and penalty graph matrices within the same framework. A novel discriminant analysis method named MvMDA was introduced by exploiting the distances between class centers of different views. Meanwhile, we also studied non-linear embeddings, and found implicit and explicit kernel mappings for multi-view learning. A unified scheme for learning by neural networks was developed which combined the learned representations with a linear embedding layer. We thereby formulated the expression of stochastic gradient descent for optimizing the proposed objective function. \par
We validated the formulation by conducting experiments in zero-shot visual object recognition and cross-modal image retrieval. It was shown that supervised and non-linear subspace learning outperformed the unsupervised and linear methods when large amount of images and texts were available. Moreover, the recognition or retrieval performance were consistently improved by embedding more views/modalities into the latent feature space. We also performed the traditional CBIR experiments where the multi-view embeddings can contribute to the performance gain.\par
Interesting future research directions include learning from the raw data to achieve an end-to-end solution for multi-view learning. We should further reduce the computational cost for kernel methods to cope with large scale of images. In addition, learning
from incomplete and unlabeled multi-view data should be studied for video analysis.
\normalsize
\bibliographystyle{IEEEtran}
\bibliography{ResearchBib.bib}
\clearpage
\onecolumn
\begin{appendices}
\include{supp}

\end{appendices}
\end{document}

%% file: supp.tex
\graphicspath{{./figures/}}
\title{Generalized Multi-view Embedding for Visual Recognition and Cross-modal Retrieval\\
Supplementary material}
\author{Guanqun Cao, Alexandros Iosifidis, \emph{Senior Member, IEEE}, Ke Chen and Moncef Gabbouj, \emph{Fellow, IEEE}        
}

\maketitle
\IEEEdisplaynontitleabstractindextext
{\centering \Huge Supplementary material \par}

\section{Derivation of the between-class scatter matrices $\mathbf{S}_B$ \eqref{eq:sbm} and $\mathbf{S}'_B$ \eqref{eq:sb2} and their difference}\label{app:lda}
\footnotesize
\begin{align}
&\mathbf{S}_B= \sum_{i=1}^V\sum_{j=1}^V\sum_{p=1}^C\sum_{\substack{q=1\\p\neq q}}^C(\bm{\mathbf{m}}_p^i-\bm{\mathbf{m}}_q^j)(\bm{\mathbf{m}}_p^i-\bm{\mathbf{m}}_q^j)^\top
\nonumber\\
\begin{split}
&  =  \sum_{i=1}^V\sum_{j=1}^V\sum_{p=1}^C\sum_{\substack{q=1\\p\neq q}}^C
\underbrace{(\W_i\X_i\e_p\e_p^\top \X_i^\top \W_i/N_p^2}_{\text{term 1}}
-\underbrace{\W_j\X_j\e_q\e_p^\top \X_i^\top \W_i/N_p/N_q}_{\text{term 2}}\\ 
&-\underbrace{ \W_i\X_i\e_p\e_q^\top \X_j^\top \W_j/N_p/N_q}_{\text{term 3}}
+ \underbrace{(\W_j\X_j\e_q\e_q^\top \X_j^\top \W_j/N_q^2)}_{\text{term 4}}
\end{split}
\end{align}
\begin{align}
    &\mathbf{S}'_B=  \sum_{i=1}^V\sum_{j=1}^V\sum_{p=1}^C\sum_{\substack{q=1\\p\neq q}}^C(\bm{\mathbf{m}}_p^i-\bm{\mathbf{m}}_q^i)(\bm{\mathbf{m}}_p^j-\bm{\mathbf{m}}_q^j)^\top
    \nonumber\\
&  =  \sum_{i=1}^V\sum_{j=1}^V\sum_{p=1}^C\sum_{\substack{q=1\\p\neq q}}^C \W_i{\X}_i ({\e}_p/N_p - {\e}_q/N_q)({\e}_p/N_p - {\e}_q/N_q)^\top \X_j^\top \W_j\nonumber\\
\begin{split}
&  =  \sum_{i=1}^V\sum_{j=1}^V\sum_{p=1}^C\sum_{\substack{q=1\\p\neq q}}^C
\underbrace{(\W_i\X_i\e_p\e_p^\top \X_j^\top \W_j/N_p^2}_{\text{term 5}}
-\underbrace{\W_i\X_i\e_q\e_p^\top \X_j^\top \W_j/N_p/N_q}_{\text{term 6}}\\
& -\underbrace{ \W_i\X_i\e_p\e_q^\top \X_j^\top \W_j/N_p/N_q}_{\text{term 7}}
+ \underbrace{(\W_i\X_i\e_q\e_q^\top \X_j^\top \W_j/N_q^2)}_{\text{term 8}}
\end{split}
\end{align}
\normalsize
\begin{singlespace}
We compare $\mathbf{S}_B$ and $\mathbf{S}'_B$, and find that term 2, 3, 6, 7 are interchangeable. Term 1 is equivalent to term 4, and term 5 is equivalent to term 8.
The difference between \eqref{eq:sbm} and \eqref{eq:sb2} is that term 1 is 
\end{singlespace}
\footnotesize
\begin{equation}
\begin{split}
\mathbf{S}_{B_1} =& (C-1)\bigg(\sum_{i=1}^V\sum_{c=1}^C \W_i^\top \X_i \e_c \e_c^\top \X_i^\top \W_i/N_c^2 \\
  &+ (V-1)\sum_{i=1}^V\sum_{c=1}^C \W_i^\top \X_i \e_c \e_c^\top \X_i^\top \W_i/N_c^2\bigg),
\end{split}
\end{equation}
\normalsize
while term 5 has
\footnotesize
\begin{equation}
\begin{split}
  \mathbf{S}'_{B_5} =& (C-1)\bigg(\sum_{i=1}^V\sum_{c=1}^C \W_i^\top \X_i \e_c \e_c^\top \X_i^\top \W_i/N_c^2 \\
  &+ \sum_{i=1}^V\sum_{\substack{j=1\\j\neq i}}^V\sum_{c=1}^C \W_i^\top \X_i \e_c \e_c^\top \X_j^\top \W_j/N_c^2\bigg),
\end{split}
\end{equation}
\normalsize
which explains the difference between $\mathbf{S}_B$ and $\mathbf{S}'_B$.
\vspace{-8pt}
\section{Proof of Gradient derivation in \eqref{eq:ccagrad} and \eqref{eq:ldagrad}}\label{app:grad}
\begin{singlespace}
Since we constrain to have a unit variance in the denominator, CCA and PLS then have the same gradient formulation as
\end{singlespace}
\footnotesize
\begin{align}
\frac{\partial \J}{\partial \Hb_i}
=\frac{\partial}{\partial \Hb_i} \operatorname{Tr}\Bigg(\,{\sum\limits_{i=1}^V \sum\limits_{\substack{j\neq i\\j=1}}^V} \W_i^\top \Hb_i\, \mathbf{L}\,\Hb_j^\top \W_j \Bigg) 
={\sum\limits_{i=1}^V \sum\limits_{\substack{j\neq i\\j=1}}^V}
\W_i \,\W_j^\top \Hb_j \,\mathbf{L}
\end{align}
\normalsize
\begin{singlespace}
If we replace the Laplacian matrix $\mathbf{L}$ in multi-view CCA and PLS with $\mathbf{L}_B^*$, then we get
\end{singlespace}
\footnotesize
\begin{equation}
\frac{\partial \J}{\partial \Hb_i}
 ={\sum\limits_{i=1}^V \sum\limits_{\substack{j\neq i\\j=1}}^V}
 \W_i \,\W_j^\top \Hb_j \,\mathbf{L}^*_B.
\end{equation}
\
\section{Quantitative results of all methods in Figures 8 and 9 in the paper}
\begin{table}[h!]
\begin{center}
\caption{MAP scores (\%) of Image queries with 2 views}
\begin{tabular}{|c | c| c| c| c | c | c | c|}
\hline
No. of Dim & LMvCCA & LMvPLS & SLMvDA & LMvMDA & KMvCCA & KMvPLS & KMvMDA \\ \hline
10 & 38.43 & 36.55 & 43.43 & 41.57 & 41.70 & 34.82 & 45.96 \\
20 &  41.22 & 38.79 & 43.73 & 42.98 & 43.46 & 38.28 & 45.42 \\
50 & 41.37 & 42.49 & 43.20 & 43.15 & 44.78 & 42.94 & 46.01 \\ 
100 & 41.54 & 41.65 & 42.28 & 42.84 & 43.73 & 43.35 & 46.32\\
150 & 41.53 & 41.63 & 41.70 &42.06 & 42.82 & 42.81 & 45.97\\
200 &  41.51 &41.57 & 41.55 & 42.05 & 40.90 & 42.05 & 44.79 \\ \hline
 \end{tabular}
  \end{center}
 \end{table}
\begin{table}[h!]
\begin{center}
\caption{MAP scores (\%) of Image queries with 3 views}
\begin{tabular}{|c | c| c| c| c | c | c | c|}
\hline
No. of Dim & LMvCCA & LMvPLS & SLMvDA & LMvMDA & KMvCCA & KMvPLS & KMvMDA \\ \hline
10 & 39.59 & 35.54 & 42.58 & 40.75 & 42.04 & 33.63 & 45.83 \\
20 & 41.66 & 39.04 & 43.05 & 41.77 & 43.12 & 37.61 & 45.61 \\ 
50 & 42.10 & 41.19 & 42.34 & 42.45 & 44.06 & 42.02 & 45.40 \\
100 & 41.46 & 41.51 & 41.76 & 41.86 & 43.32 & 43.78 & 45.92 \\
150 & 41.43 & 41.44 & 41.46 & 41.60 & 42.78 & 43.26 & 45.64 \\
200 & 41.41 & 41.44 & 41.43 &41.65 & 40.78 & 42.59 & 45.39 \\ \hline
 \end{tabular}
  \end{center}
 \end{table}
\begin{table}[h!]
\begin{center}
\caption{MAP scores (\%) of Image queries with 4 views}
\begin{tabular}{|c | c| c| c| c | c | c | c|}
\hline
No. of Dim & LMvCCA & LMvPLS & SLMvDA & LMvMDA & KMvCCA & KMvPLS & KMvMDA \\ \hline
10 & 43.01 & 35.85 & 42.13 & 43.15 &  43.53 & 34.44 & 45.81 \\
20 & 43.60 & 39.52 & 43.09 & 42.92 & 44.80 & 38.74 & 45.53 \\
50 & 42.53 & 41.44 & 42.86 & 42.86 & 45.13 & 41.94 & 46.48 \\
100 & 41.48 & 41.52 & 41.88 & 42.59 & 43.32 & 44.18 & 46.66 \\
150 & 41.43 & 41.45 & 41.43 & 41.73 & 43.15 & 43.51 & 46.18 \\
200 & 41.42 & 41.45 & 41.44 & 41.60 & 42.50 & 43.19 & 46.60 \\ \hline
 \end{tabular}
  \end{center}
 \end{table}
\begin{table}[h!]
\begin{center}
\caption{MAP scores (\%) of Text queries with 2 views}
\begin{tabular}{|c | c| c| c| c | c | c | c|}
\hline
No. of Dim & LMvCCA & LMvPLS & SLMvDA & LMvMDA & KMvCCA & KMvPLS & KMvMDA \\ \hline
10 & 37.04 & 34.92 & 40.43&  37.76&  39.02&  32.77&  40.95\\
20 & 39.09 & 37.30&   40.58 & 39.67&  40.52&  35.98 & 40.71\\
50 & 39.08 & 40.42&  40.07 & 39.94 & 41.83 & 40.47&  40.96 \\
100 & 38.30 & 38.35&  38.79&  39.66&  39.92&  40.16&  39.92\\
150 & 38.28 & 38.35&  38.53&  39.02&  38.44&  39.40&   38.72\\
200 & 38.26 & 38.32&  38.30 &   38.79 & 34.74&  37.10 &  38.10 \\ \hline
\end{tabular}
  \end{center}
 \end{table}
\begin{table}[h!]
\begin{center}
\caption{MAP scores (\%) of Text queries with 3 views}
\begin{tabular}{|c | c| c| c| c | c | c | c|}
\hline
No. of Dim & LMvCCA & LMvPLS & SLMvDA & LMvMDA & KMvCCA & KMvPLS & KMvMDA \\ \hline
10 & 38.36 & 33.59 & 41.22 & 38.06 & 39.87&  31.31 & 40.91\\
20 & 39.01 & 37.28 & 40.63 & 39.79 & 40.40 &   34.70 &   40.60 \\
50 & 39.64 & 39.23 & 39.52 & 39.91 & 41.41 & 39.41 & 40.17 \\
100 & 38.86&  38.85&  39.03 & 39.40 &   40.25 & 40.05&  40.09\\
150 & 38.84 & 38.83&  38.84&  39.06&  38.78&  39.06&  38.19 \\
200 &  38.83 & 38.83&  38.84 & 38.98 & 35.93&  37.27&  37.7 \\ \hline
\end{tabular}
  \end{center}
 \end{table}
\begin{table}[h!]
\begin{center}
\caption{MAP scores (\%) of Text queries with 4 views}
\begin{tabular}{|c | c| c| c| c | c | c | c|}
\hline
No. of Dim & LMvCCA & LMvPLS & SLMvDA & LMvMDA & KMvCCA & KMvPLS & KMvMDA \\ \hline
10 & 40.57 & 33.90 & 40.24&  40.95 & 40.84 & 32.07 & 40.72\\
20 & 40.78 & 37.77 & 40.68 &  40.64 & 41.71 & 35.59 & 40.90 \\
50 & 39.98 & 39.62 & 40.43 & 40.17 & 41.66 & 38.85&  40.73\\
100 & 38.85 & 38.85 & 38.98 & 39.49&  40.63 & 40.30 &   40.84\\
150 & 38.84 & 38.83 & 38.88 & 39.15&  39.08 & 38.64 & 39.67 \\
200 & 38.83 & 38.83 & 38.85 & 39.00 & 35.89 & 37.77 & 39.18 \\ \hline
\end{tabular}
  \end{center}
 \end{table}
\section{Contribution of adding views in retrieval performance on the Wikipedia dataset}
\begin{figure}[h!]
\centering
\includegraphics[width=.308\textwidth]{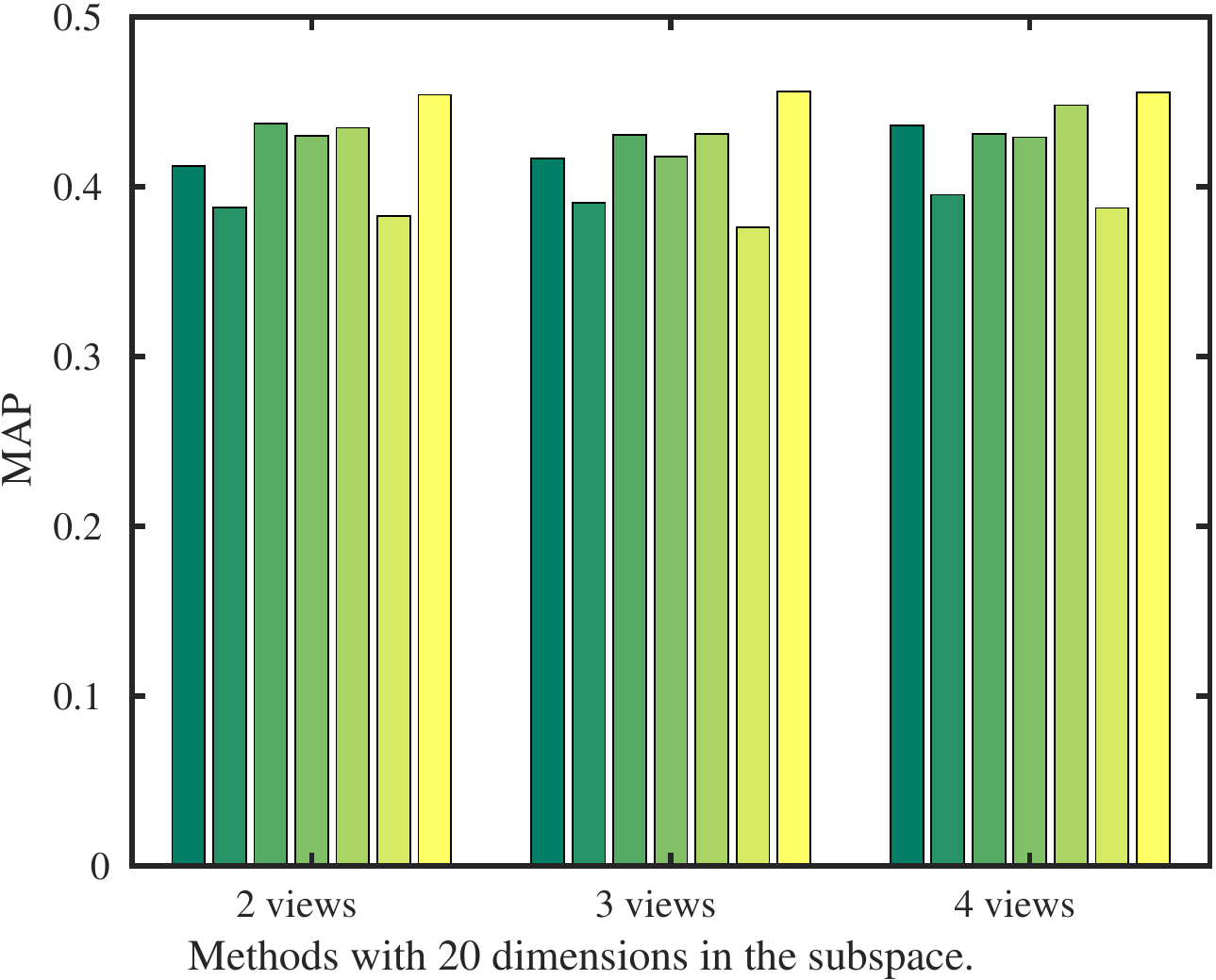}%
\includegraphics[width=.3\textwidth]{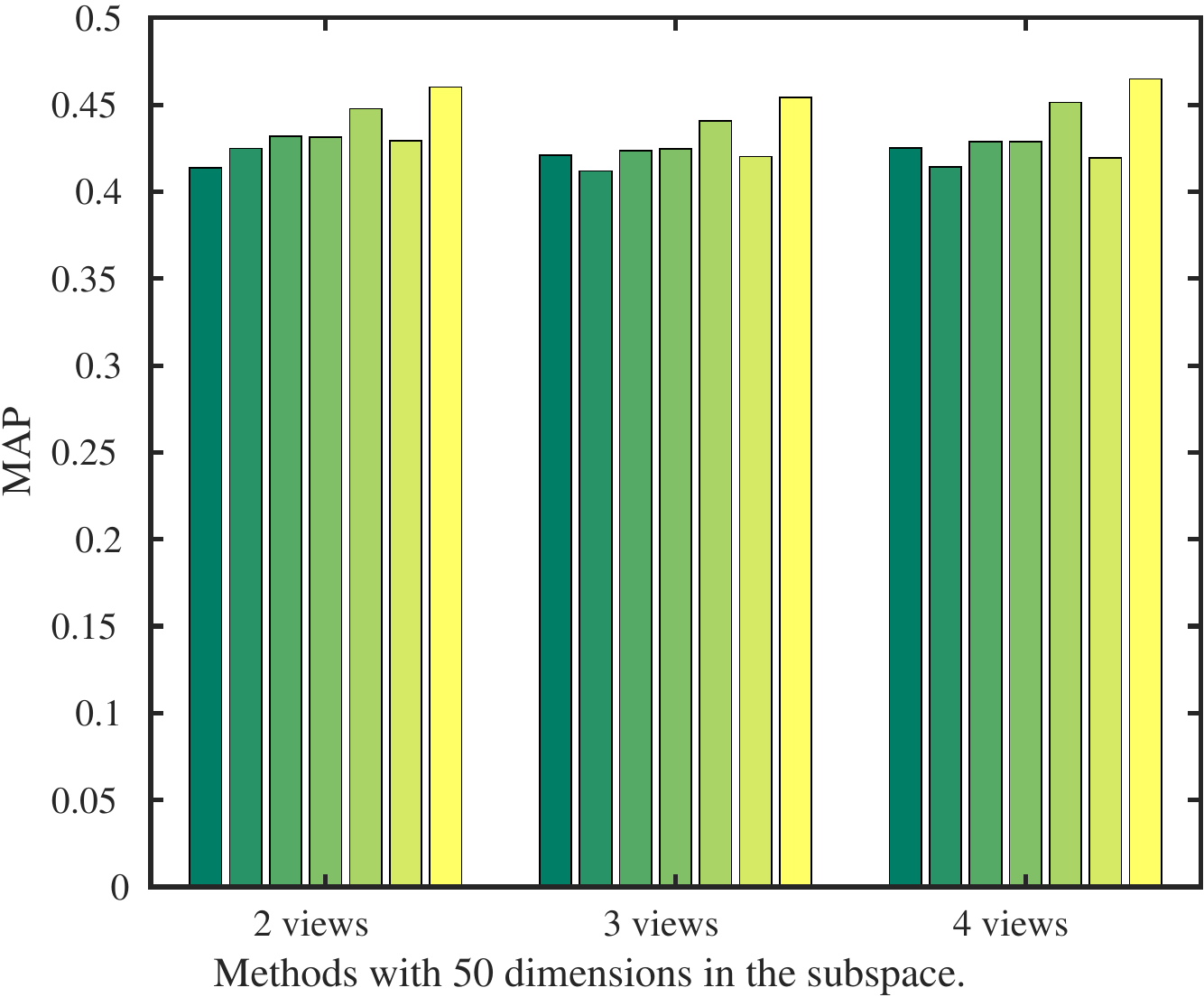}%
\includegraphics[width=.3\textwidth]{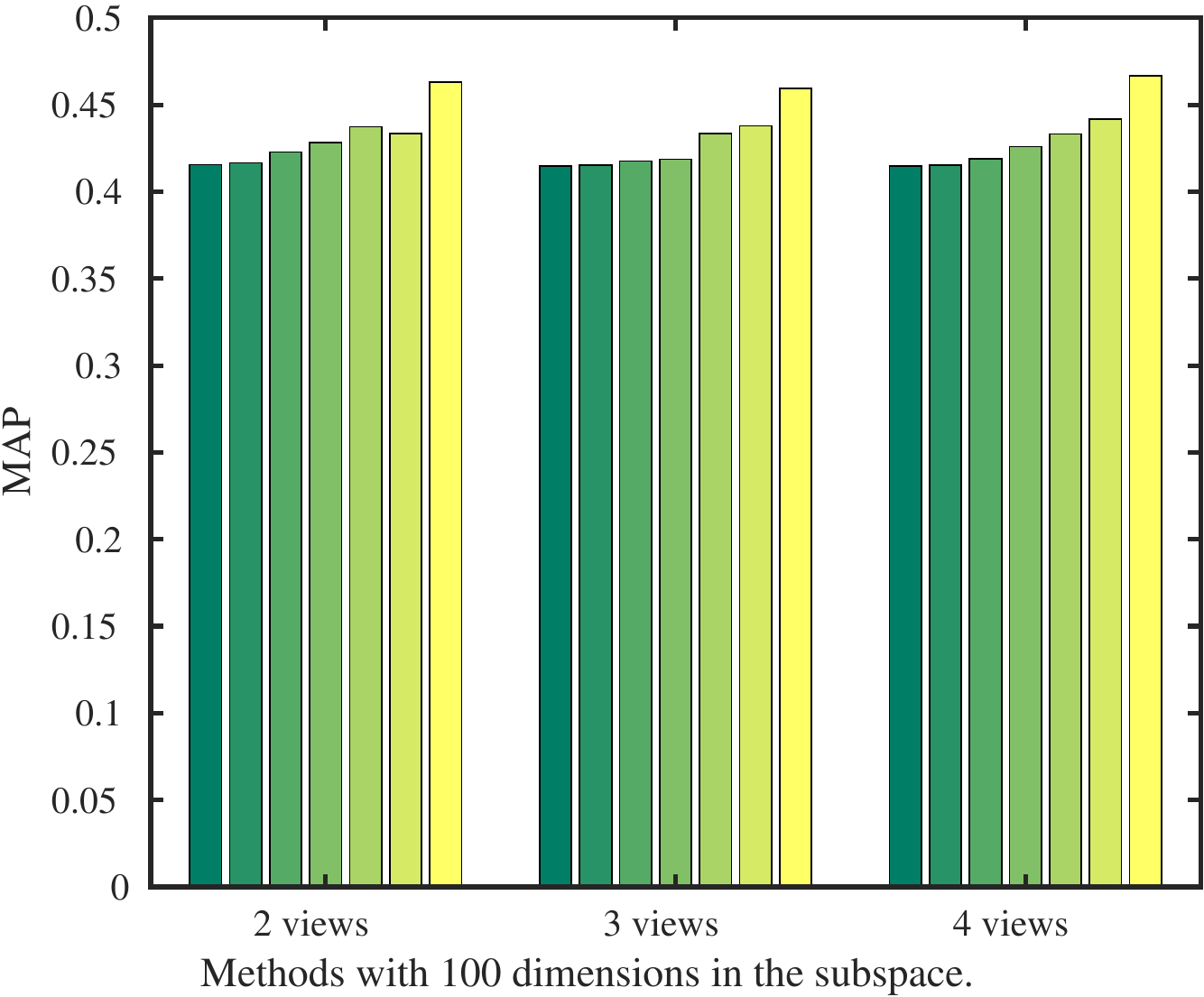}\\
\caption{Search texts using image queries. For each group of bars, from the left to the right, the methods are LMvCCA, LMvPLS, SLMvDA, LMvMDA, KMvCCA, KMvPLS, KMvMDA, respectively.}
\end{figure}
\begin{figure}[h!]
\centering
\includegraphics[width=.3\textwidth]{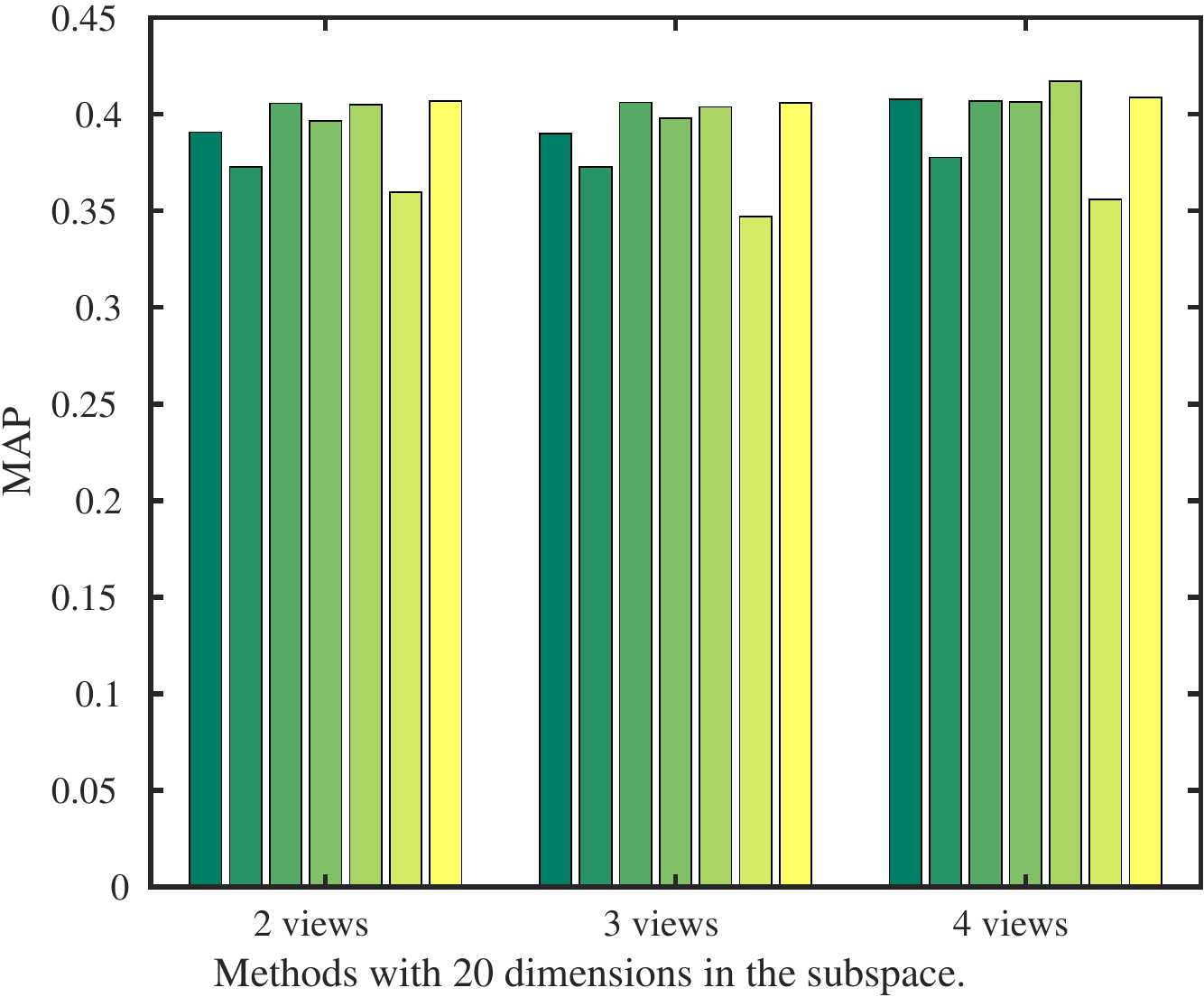}%
\includegraphics[width=.315\textwidth]{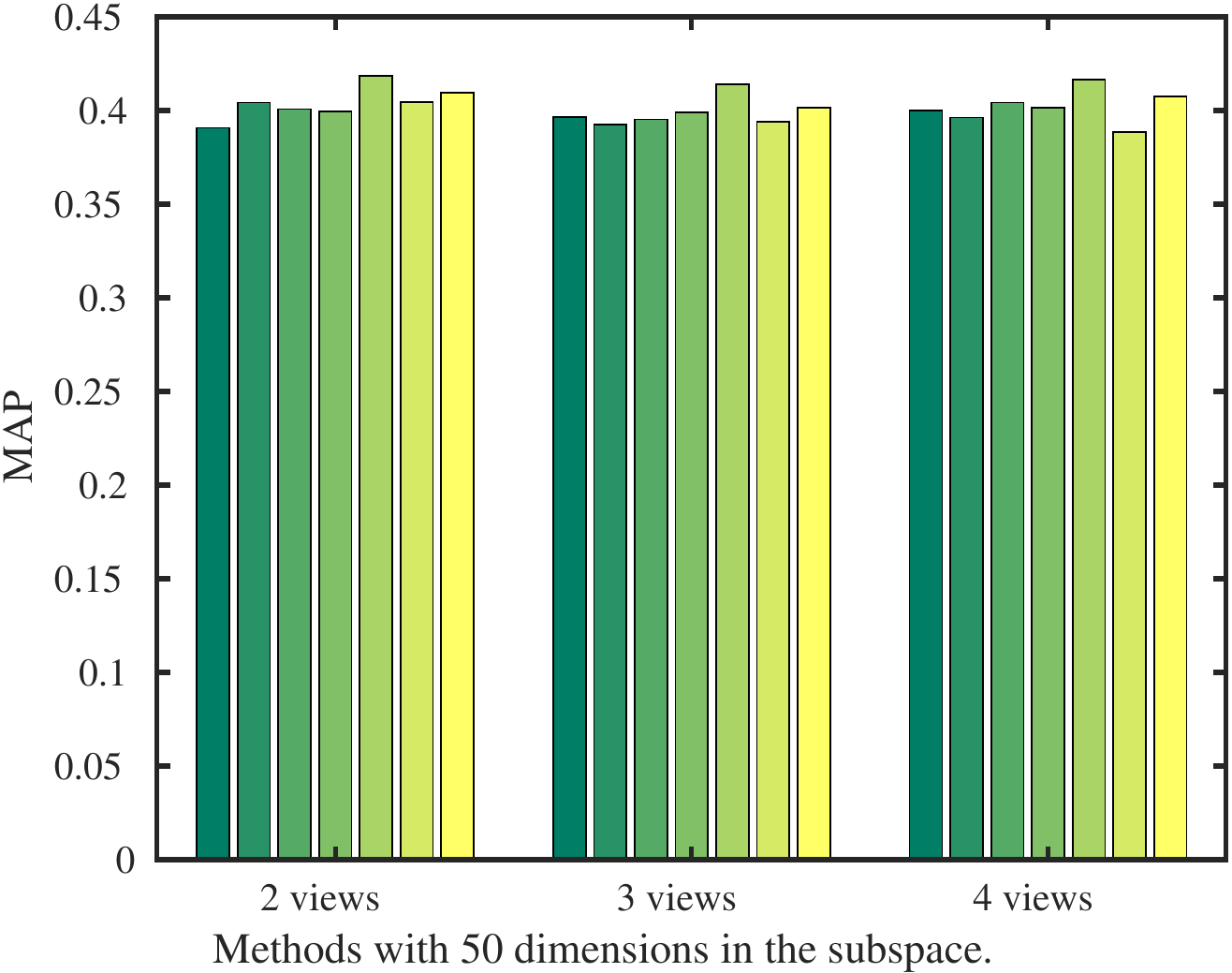}%
\includegraphics[width=.3\textwidth]{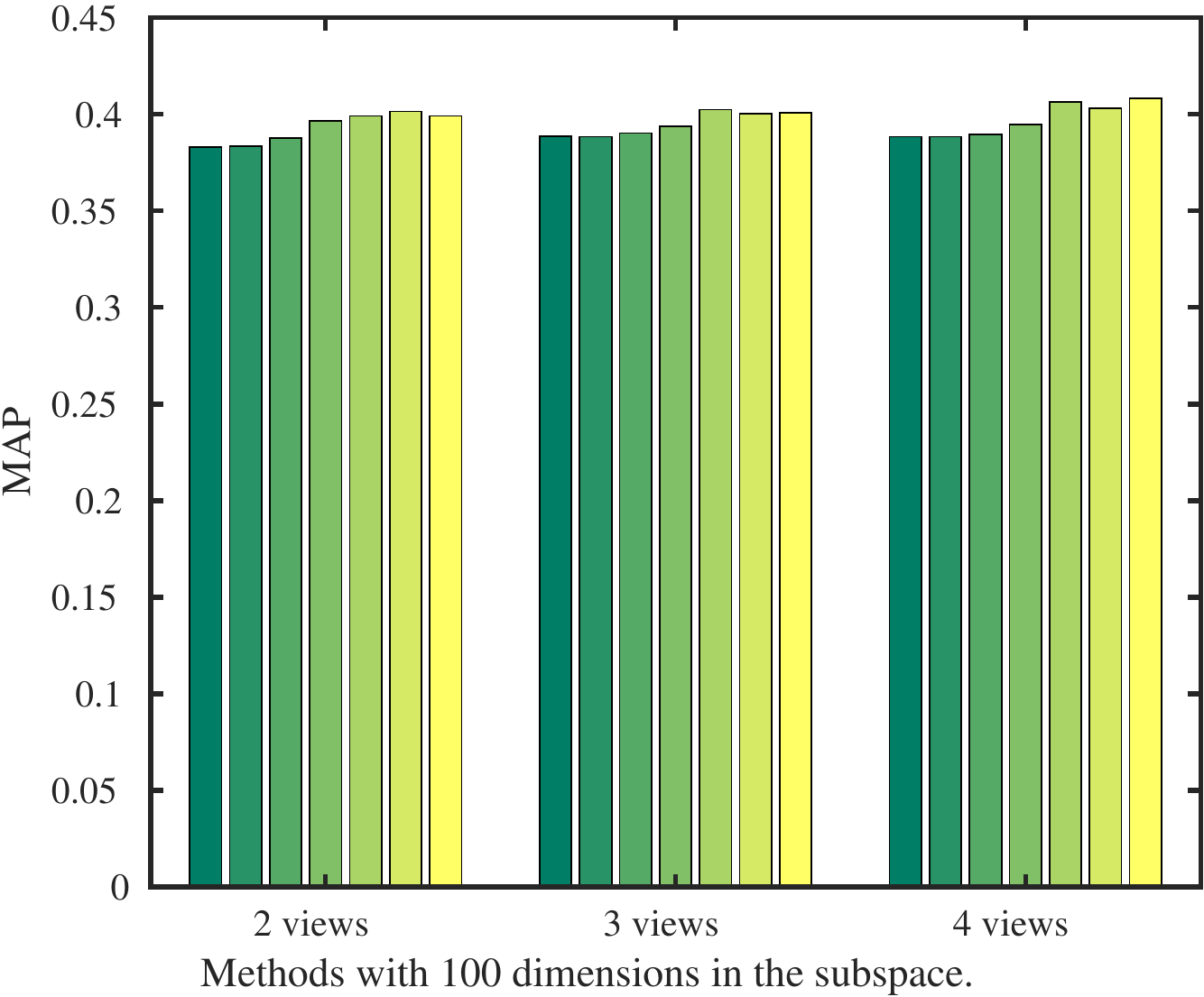}
\caption{Search images using text queries. For each group of bars, from the left to the right, the methods are LMvCCA, LMvPLS, SLMvDA, LMvMDA, KMvCCA, KMvPLS, KMvMDA, respectively.}
\end{figure}